\documentclass[journal]{IEEEtran}
\usepackage{graphicx}
\usepackage{amsmath}
\usepackage{amssymb}
\usepackage{amsthm}
\usepackage{multirow}
\usepackage{xcolor} 
\usepackage{booktabs}
\usepackage{algorithm}
\usepackage{algorithmic}
\usepackage{hyperref}

\newtheorem{proposition}{Proposition}

\newcommand{\x}{\boldsymbol{x}}
\newcommand{\ba}{\boldsymbol{a}}
\newcommand{\w}{\boldsymbol{w}}

\newcommand{\R}{\boldsymbol{R}}

\usepackage{pifont}% http://ctan.org/pkg/pifont
\newcommand{\yes}{{\text{\ding{51}}}}
\newcommand{\no}{{\text{\ding{55}}}}

\begin{document}
\title{Explaining Deep Neural Networks and Beyond:\\A Review of Methods and Applications}

\author{Wojciech~Samek$^{\dag*}$,~\IEEEmembership{Member,~IEEE,}        Gr{\'e}goire~Montavon$^{\dag*}$, Sebastian~Lapuschkin, Christopher~J.~Anders,
        and~Klaus-Robert~M{\"u}ller$^*$,~\IEEEmembership{Member,~IEEE}% <-this % stops a space
\thanks{$^\dag$ W.~Samek and G.~Montavon contributed equally to this work.}%
\thanks{$^*$ Corresponding authors: W.~Samek, G.~Montavon and K.-R.~M{\"u}ller.}%
\thanks{W.~Samek is with the Dept.\ of Artificial Intelligence, Fraunhofer Heinrich Hertz Institute, 10587 Berlin, Germany, and with BIFOLD -- Berlin Institute for the Foundations of Learning and Data, 10587 Berlin, Germany. (e-mail: wojciech.samek@hhi.fraunhofer.de).
}% <-this % stops a space
\thanks{G.~Montavon and C.~Anders are with the Machine Learning Group, Technische Universit{\"a}t Berlin, 10587 Berlin, Germany, and with BIFOLD -- Berlin Institute for the Foundations of Learning and Data, 10587 Berlin, Germany. (e-mail: gregoire.montavon@tu-berlin.de).}% <-this % stops a space
\thanks{S.~Lapuschkin is with the Dept.\ of Artificial Intelligence, Fraunhofer Heinrich Hertz Institute, 10587 Berlin, Germany.
}% <-this % stops a space
\thanks{K.-R.~M{\"u}ller is with the Machine Learning Group, Technische Universit{\"a}t Berlin, 10587 Berlin, Germany, and also with the Dept.\ of Artificial Intelligence, Korea University, Seoul 136-713, South Korea, the Max Planck  Institute for  Informatics,   66123  Saarbr{\"u}cken, Germany, and BIFOLD -- Berlin Institute for the Foundations of Learning and Data, 10587 Berlin, Germany. (e-mail: klaus-robert.mueller@tu-berlin.de).}% <-this % stops a space
\thanks{Manuscript accepted for publication at \textsc{Proceedings of the IEEE}. \\
\href{http://doi.org/10.1109/JPROC.2021.3060483}{http://doi.org/10.1109/JPROC.2021.3060483}\\
© 2021 IEEE.  Personal use of this material is permitted.  Permission from IEEE must be obtained for all other uses, in any current or future media, including reprinting/republishing this material for advertising or promotional purposes, creating new collective works, for resale or redistribution to servers or lists, or reuse of any copyrighted component of this work in other works.}}

\maketitle

\begin{abstract}
With the broader and highly successful usage of machine learning in industry and the sciences, there has been a growing  demand for Explainable AI. Interpretability and explanation methods for gaining a better understanding about the problem solving abilities and strategies of nonlinear Machine Learning, in particular, deep neural networks, are therefore receiving increased attention. In this work we aim to (1) provide a timely overview of this active  emerging field, with a focus on `post-hoc' explanations, and explain its theoretical foundations, (2) put interpretability algorithms to a test both from a theory and comparative evaluation perspective using extensive simulations, (3)  outline  best practice aspects i.e.\ how to best include interpretation methods into the standard usage of machine learning and (4)  demonstrate successful usage of explainable AI in a representative selection of application scenarios. Finally, we discuss challenges and possible future directions of this exciting foundational field of machine learning. 
\end{abstract}

\IEEEpeerreviewmaketitle

%%%%%%%%%%%%%%%%%%%%%%%%%%%%%%%%%%%%%%%%%%%%%%%%%%%%%%%%%%%%%%%%%%%%%%%%%%%%%%%%%%%%
\section{Introduction}
\label{sec:intro}
A main goal of machine learning is to learn accurate decision systems respectively predictors that can help automatizing tasks, that would otherwise have to be done by humans.
Machine Learning (ML) has supplied a wealth of algorithms
that have demonstrated important successes in the sciences and industry; most popular ML work horses are considered to be  kernel methods (e.g.\ \cite{vapnik95,scholkopf1998nonlinear,muller2001introduction,scholkopf2002learning,williams2006gaussian}) and particularly during the last decade deep learning methods (e.g.\ \cite{bishop1996neuralnets,goodfellow2016deep,lecun2012efficient,lecun2015deep,schmidhuber2015deep,hochreiter1997long}) have gained highest popularity. 

As ML is increasingly  used in real-world applications, a general consensus has emerged that high prediction accuracy alone may not be sufficient in practice \cite{lapuschkin2019unmasking,caruana2015intelligible,DBLP:series/lncs/11700}. Instead, in practical engineering of systems, critical features that are typically considered beyond excellent prediction itself are (a) robustness of the system to measurement artefacts or adversarial perturbations \cite{DBLP:conf/eccv/SuZCYCG18}, (b) resilience to drifting data distributions \cite{DBLP:journals/csur/GamaZBPB14}, (c) ability to accurately assess the confidence of its own predictions \cite{DBLP:journals/tnn/PapadopoulosEM01, DBLP:conf/cvpr/NguyenYC15}, (d) safety and security aspects ~\cite{berkenkamp2017safe, DBLP:conf/cav/KatzBDJK17,carlini2017towards,warnecke2019don}, (e) legal requirements or adherence to social norms \cite{goodman2016european,DBLP:conf/cvpr/GuptaJFSA18}, (f) ability to complement human expertise in decision making \cite{Jarrahi2018}, or (g) ability to reveal to the user the interesting correlations it has found in the data \cite{Khan2001,schutt2017quantum}.

Orthogonal to the quest for better and more holistic machine learning models, Explainable AI (XAI) \cite{DBLP:series/lncs/11700,holzinger2018machine,lipton2018mythos,montavon2018methods,Bau2020} has developed as a subfield of machine learning that seeks to augment the training process, the learned representations and the decisions with human-interpretable explanations. An example is {\em medical diagnosis}, where the input examples (e.g.\ histopathological images) come with various artifacts (e.g.\ stemming from image quality or suboptimal annotations) that have in principle nothing to do with the diagnostic task, yet, due to the limited amount of available data, the ML model may harvest specifically these spurious correlations with the prediction target  (e.g.\ \cite{hagele2019resolving,Soneson2014}). Here interpretability could point at anomalous or awkward decision strategies before harm is caused in a later usage as a diagnostic tool.

Similarly essential when using ML in the sciences is again interpretabilty, since ideally, the transparent ML model --- having learned from data --- may  have embodied scientific knowledge that would subsequently   provide insight to the scientist, occasionally this can even be novel scientific insight (see e.g.\ \cite{schutt2017quantum}). --- Note that in numerous scientific applications it has been most common so far to use linear models \cite{Rosenblatt1958}, favoring interpretabilty often at the expense of predictivity (see e.g.\ \cite{DBLP:journals/neuroimage/HaufeMGDHBB14, DBLP:journals/bmcbi/MaSH07}).

To summarize, there is a strong push toward better understanding  ML systems that are being used and in consequence blackbox algorithms are more and more abandoned for many applications. This growing consensus has led to a strong growth of a subfield of ML, namely  explainable AI (XAI) that strives to produce transparent nonlinear learning methods, and supplies novel theoretical perspectives on machine learning models, along with powerful practical tools for a better understanding and interpretation of AI systems.

In this review paper, we will summarize the recent exciting developments, present different classes of XAI methods that have been proposed in the context of deep neural networks, provide theoretical insights, and highlight the current best practices when applying these methods.
Note finally, that we do not attempt an encyclopedic treatment of all available XAI literature, rather, we present a slightly biased point of view (and in doing so we often draw from the work of the authors). In particular, we focus on \textit{post-hoc explanation methods}, which take any model, typically the best performing one, and analyze it in a second step in order to uncover its decision strategy. We also provide --- to the best of our knowledge --- reference to other related works for further reading.

%%%%%%%%%%%%%%%%%%%%%%%%%%%%%%%%%%%%%%%%%%%%%%%%%%%%%%%%%%%%%%%%%%%%%%%%%%%%%%%%%%%%
\section{Towards Explaining Deep Neural Networks}
\label{section:towards}

\begin{figure}[ht]
    \centering
    \includegraphics[width=0.9\linewidth]{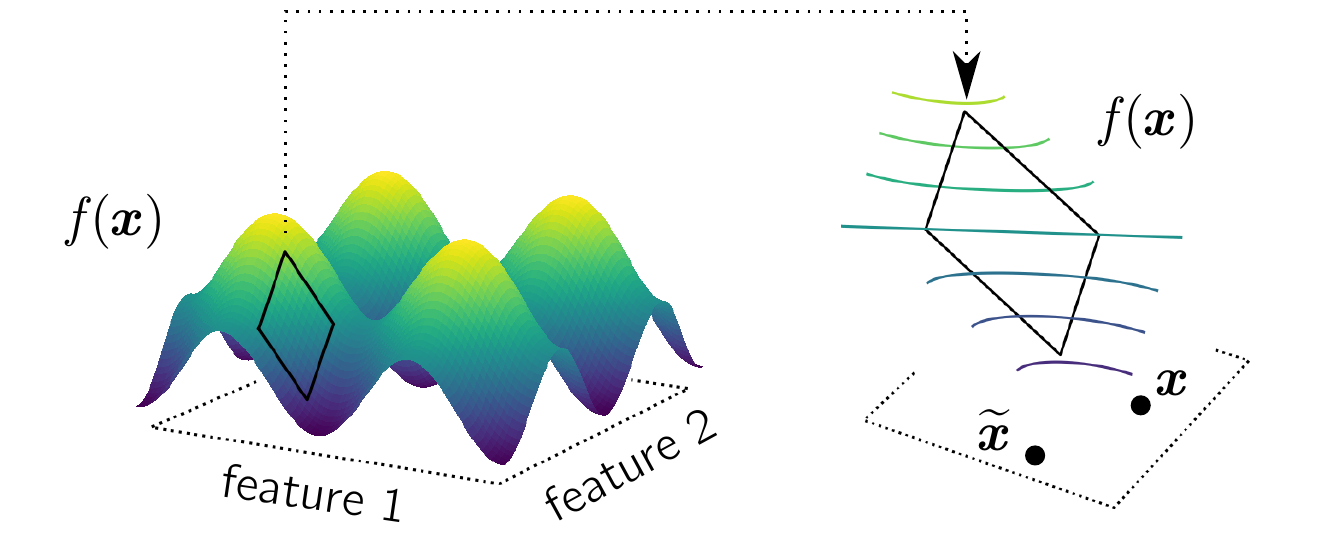}
    \caption{Example of a nonlinear function of the input features, which produces some prediction. The function can be approximated locally as a linear model.}
    \label{fig:function}
\end{figure}

Before discussing aspects of the problem of explanation that are specific to deep neural networks, we first introduce some basics of Explainable AI, which apply to a fairly general class of machine learning models.  The ML model will be assumed to be \textit{already trained} and the input-output relation it implements will be abstracted by some \textit{function}:
$$
f\colon\mathbb{R}^d \to \mathbb{R}.
$$
This function receives as input a vector of real-valued features $\x = (x_1,\dots,x_d)$ typically corresponding to various sensor measurements. The function produces as an output a real-valued score on which the decision is based. A sketch of such function receiving two features $x_1$ and $x_2$ as input is given in \mbox{Fig.\ \ref{fig:function}}.

In the context of ML classification, the function output can be interpreted as the amount of evidence for\,/\,against deciding in favor of a certain class. A classification decision is then obtained from the output score by testing whether the latter is above a certain threshold, or for multiclass problems, larger than the output score of other functions representing the remaining classes.

\smallskip

In a medical scenario, the function may receive as input an array of clinical variables, and the output of the function may represent the evidence for a certain medical condition \cite{Kourou2015}. In an engineering setting, the input could be the composition of some compound material, and the output could be a prediction of its strength \cite{Yeh1998} or stability.

\smallskip

Suppose a given instance is predicted by the machine learning model to be healthy, or a compound material is predicted to have high strength. We may choose to trust the prediction and go ahead with next step within an application scenario. However, we may benefit from taking a closer look at that prediction, e.g.\ to verify that the prediction `healthy' is associated to relevant clinical information, and not some spurious features that accidentally correlate with the predicted quantity in the dataset \cite{LapCVPR16,lapuschkin2019unmasking}. Such problem can often be identified by building an {\em explanation} of the ML prediction \cite{lapuschkin2019unmasking}.

Conversely, suppose that another instance is predicted by the machine learning model to be of low health or low strength. Here, an explanation could prove equally useful as it could hint at actions to be taken on the sample to improve its predicted score \cite{DBLP:conf/fat/UstunSL19}, e.g.\ possible therapies in a medical setting, or small adjustments of the compound design that lead to higher strength.

%%%%%%%%%%%
\subsection{How to Explain: Global vs.\ Local}

Numerous approaches have emerged to shed light onto machine learning predictions. Certain approaches such as activation-maximization \cite{DBLP:journals/corr/SimonyanVZ13,DBLP:journals/corr/NguyenYC16,DBLP:conf/nips/NguyenDYBC16} aim at a {\em global} interpretation of the model, by identifying prototypical cases for the output quantity
$$
\x^\star = \arg\max_{\x} f(\x)
$$
and allowing in principle to verify that the function has a high value only for the valid cases.  While these prototypical cases may be interesting per se, both for model validation or knowledge discovery, such prototypes will be of little use to understand for a given  example $\x$ (say, near the decision boundary) what features play in favor or against the model output $f(\x)$. 

Specifically, we would like to know for that very example what input features contribute positively or negatively to the given prediction. These {\em local} analyses of the decision function have received growing attention and many approaches have been proposed \cite{baehrens2010explain,zeiler2014visualizing, BachPLOS15, ribeiro2016should, SundararajanTY17}. For simple models with limited nonlinearity, the decision function can be approximated locally as the linear function \cite{BachPLOS15}:
\begin{align}
f(\x) \approx \sum_{i=1}^d \,\underbrace{[\nabla f(\widetilde{\x})]_i \cdot (x_i - \widetilde{x}_i)}_{R_i}
\label{eq:simple}
\end{align}
where $\widetilde{\x}$ is some nearby root point
(cf.\ Fig.\ \ref{fig:function}). This expansion takes the form of a weighted sum over the input features, where the summand $R_i$ can be interpreted as the contribution of feature $i$ to the prediction. Specifically, an inspection of the summands reveals that a feature $x_i$ will be attributed strong relevance if the following two conditions are met: (1) the feature must be expressed in the data, i.e.\ it differs from the reference value $\widetilde{x}_i$, and (2) the model output should be sensitive to the presence of that feature, i.e.\ $[\nabla f(\widetilde{\x})]_i \neq 0$. An explanation for the prediction can then be formed by the vector of relevance scores $(R_i)_i$. It can be given to the user as a histogram over the input features or as a heatmap.

\begin{figure}[ht]
    \centering
    \includegraphics[scale=0.65]{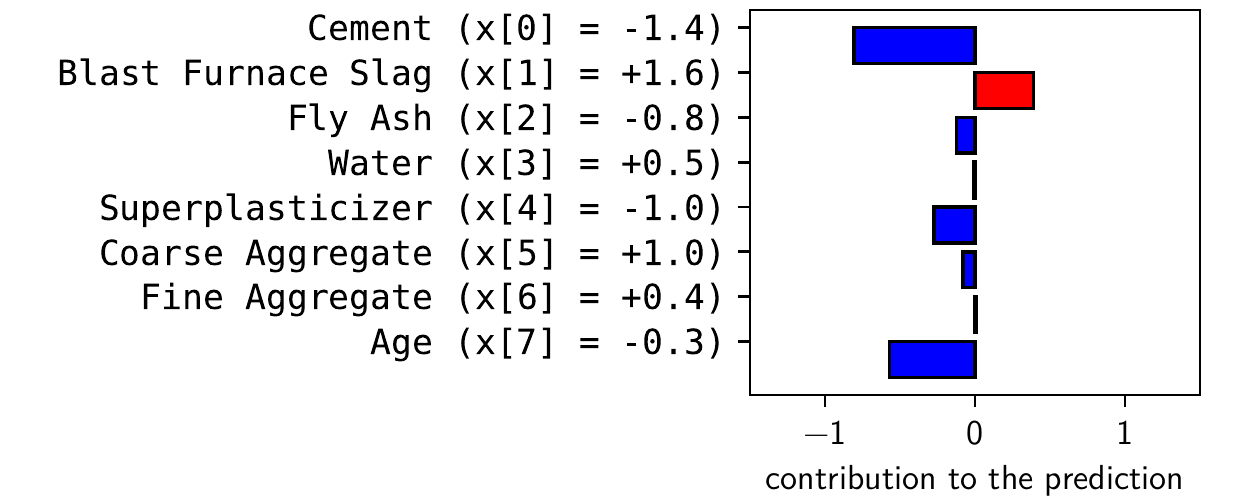}
    \caption{Input example predicted to have low compressive strength, and a feature-wise explanation of the prediction. Red and blue color indicate positive and negative contributions.}
    \label{fig:explanation}
\end{figure}

For illustration, consider the problem of explaining a prediction for a data point from the Concrete Compressive Strength dataset \cite{Yeh1998}. For this data point, a simple two-layer neural network model predicts a low compressive strength. Applying the analysis above gives an explanation for this prediction, which we show in Fig.\ \ref{fig:explanation}.

For this example low cement concentration and below average age are factors of low compressive strength, although this is partly compensated by a high quantity of blast furnace slag.

Furthermore, for an explanation to be interpretable by its receiver, the latter must be able to make sense of the input features. Some features such as `cement', `water', and `age', are understandable to everyone, however, more technical terms such as `blast furnace slag' or `superplaticizer' may only be accessible to a domain expert. Therefore, when using these explanation techniques, we make the implicit assumption that those input features are interpretable to the receiver.

%%%%%%%%%%%
\subsection{Deep Networks and the Difficulty of Explaining Them}
\label{section:difficult}

In practice, linear models or shallow neural networks may not be sufficiently expressive to predict the task optimally. Deep neural networks (DNNs) have been proposed as a way of producing more predictive models. They can be abstracted as a sequence of layers
$$
f(\x) = f_L \circ \dots \circ f_1(\x),
$$
where each layer applies a linear transformation followed by an element-wise nonlinearity. Combining a large number of these layers endows the model with high prediction power. DNNs have proven especially successful on computer vision tasks \cite{DBLP:conf/nips/KrizhevskySH12, DBLP:journals/corr/SimonyanZ14a, DBLP:conf/cvpr/HeZRS16}. However, DNN models are also much more complex and nonlinear, and quantities entering into the simple explanation model of Eq.\ \eqref{eq:simple} become considerably harder to compute and to estimate reliably.

A first difficulty comes from the multiscale and distributed nature of neural network representations. Some neurons are activated for only a few data points, whereas others apply more globally. The prediction is thus a sum of local and global effects, which makes it difficult (or impossible) to find a root point $\widetilde{\x}$ that linearly expands to the prediction for the data point of interest. The transition from the global to local effect indeed introduces a nonlinearity, which Eq.\ \eqref{eq:simple} cannot capture.

\begin{figure}[ht]
    \centering
    \includegraphics[width=1.0\linewidth]{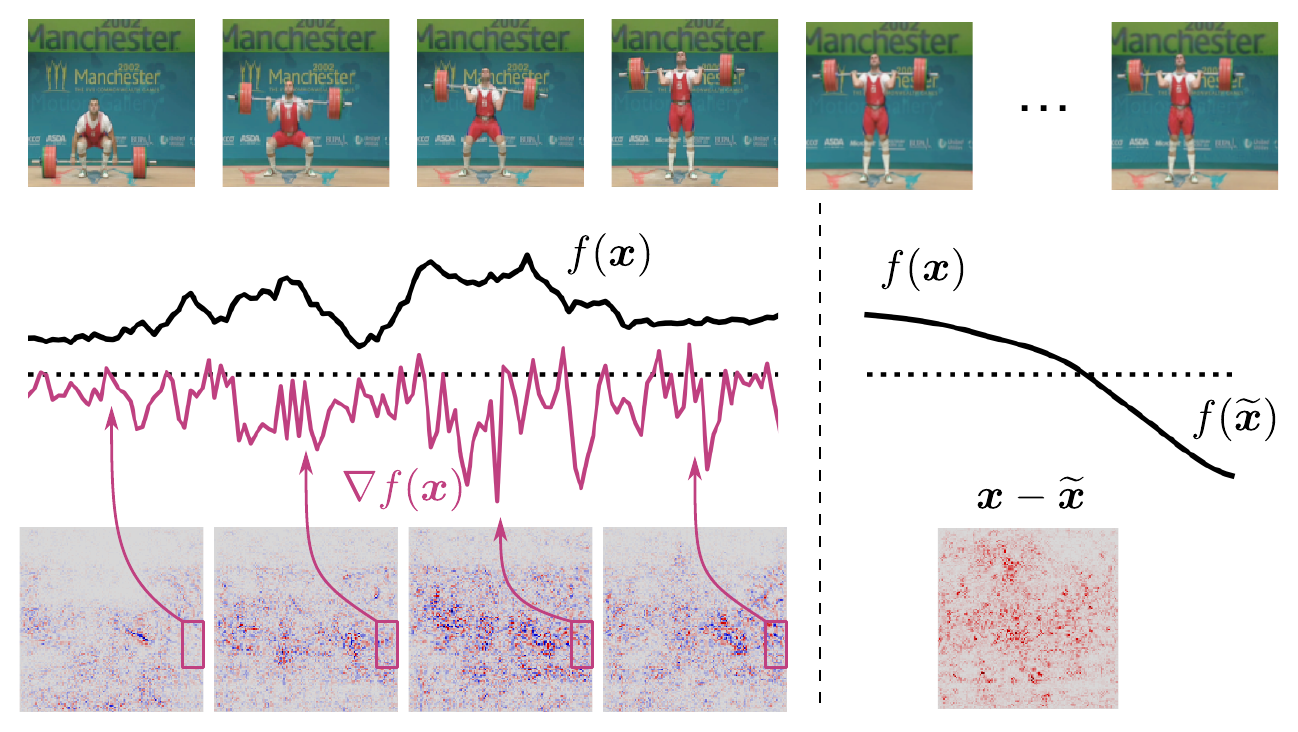}
    \caption{Two difficulties encountered when explaining DNNs. {\em Left:} Shattered gradient effect causing gradients to be highly varying and too noisy to be used for explanation. {\em Right:} Pathological minima in the function, making it difficult to search for meaningful reference points.}
    \label{fig:difficulties}
\end{figure}

A second source of instability arises from the high depth of recent neural networks, where a `shattered gradient' effect was observed \cite{DBLP:conf/icml/BalduzziFLLMM17}, noting that the gradient locally resembles white noise. In particular, it can be shown that for deep rectifier networks, the number of discontinuities of the gradient can grow in the worst case exponentially with depth \cite{DBLP:conf/nips/MontufarPCB14}. The shattered gradient effect is illustrated in \mbox{Fig.\ \ref{fig:difficulties}} (left) for the well-established VGG-16 network \cite{DBLP:journals/corr/SimonyanZ14a}: The network is fed multiple consecutive video frames of an athlete lifting a barbell, and we observe the prediction for the output neuron `barbell'. The gradient of the prediction is changing its value much more quickly than the prediction itself. An explanation based on such gradient would therefore inherit this noise.

A last difficulty comes from the challenge of searching for a root point $\widetilde{\x}$ on which to base the explanation, that is both close to the data and not an adversarial example \cite{DBLP:journals/corr/GoodfellowSS14,DBLP:conf/cvpr/NguyenYC15}. The problem is illustrated in \mbox{Fig.\ \ref{fig:difficulties}} (right), where we showcase a reference point $\widetilde{\x}$ that does not carry any meaningful visual difference to the original data $\x$, but for which the function output has changed dramatically. The problem of adversarial examples can be explained by the gradient noise, that causes the model to `overreact' to certain pixel-wise perturbations, and also by the high dimensionality of the data ($224 \times 224 = 50176$ pixels for VGG-16 and the ImageNet data set) where many small pixel-wise effects cumulate into a large effect on the model output.

\section{Practical Methods for Explaining DNNs}
\label{section:methods}

In view of the multiple challenges posed by analyzing deep neural network functions, building robust and practical methods to explain their decisions has developed into an own research area \cite{montavon2018methods, DBLP:journals/csur/GuidottiMRTGP19, DBLP:series/lncs/11700} and an abundance of  methods have been proposed. In parallel, efficient software (cf.\ Appendix \ref{appendix:software} for a list) makes these newly developed methods readily usable in practice, and allows researchers to perform systematic comparisons between them on reference models and datasets.

In this section, we focus on four families of explanation techniques: Interpretable Local Surrogates, Occlusion Analysis, Gradient-based techniques, and Layer-Wise Relevance Propagation. In our view, these techniques exemplify the current diversity of possible approaches to explaining predictions in terms of input features, and taken together provide a broad coverage of the types of models to explain and the practical use cases. We give references to further related methods in the corresponding subsections. Table \ref{tab:xai-glossary} in Appendix \ref{appendix:software} provides a glossary of all referenced methods.

%%%%%%%%%%%
\subsection{Interpretable Local Surrogates \cite{ribeiro2016should}}
\label{section:surrogate}

This category of methods aims to replace the decision function by a local surrogate model that is structured in a way that it is self-explanatory (an example of a self-explanatory model is the linear model). This approach is embodied in the LIME algorithm~\cite{ribeiro2016should}, which was successfully applied to DNN classifiers for images and text. Explanation can be achieved by first defining some local distribution $p_{\x}(\boldsymbol{\xi})$ around our data point $\x$, learning the parameter $\boldsymbol{v}$ of the linear model that best matches the function locally:
\begin{align*}
\min_{\boldsymbol{v}} ~  \int \big[f(\boldsymbol{\xi}) - \boldsymbol{v}^\top \boldsymbol{\xi} \big]^2 \cdot d p_{\x}(\boldsymbol{\xi})
\end{align*}
and then extracting local feature contributions, e.g.\ $R_i = v_i x_i$. Because the method does not rely on the gradient of the original DNN model, it avoids some of the difficulties discussed in Section \ref{section:difficult}. The LIME method also covers the incorporation of sparsity or simple decision trees to the surrogate model to further enhance interpretability. Additionally, the learned surrogate model may be based on its own set of \emph{interpretable features}, allowing to produce explanations in terms of features that are maximally interpretable to the human.
Other methods that explain by building a local surrogate include LORE \cite{guidotti2018local} and Anchors \cite{ribeiro2018anchors}. Furthermore, a broader set of methods do not consider a specific location in the input space and builds instead a global surrogate model of the decision function, where the surrogate model readily incorporates interpretability structures. We discuss these global `self-explainable' models in Section \ref{section:others}.

%%%%%%%%%%%
\subsection{Occlusion Analysis \cite{zeiler2014visualizing}}
\label{section:occlusion}

Occlusion Analysis is a particular type of perturbation analysis where we repeatedly test the effect on the neural network output, of occluding patches or individual features in the input image \cite{zeiler2014visualizing,zintgraf2017visualizing}:
$$
R_i = f(\x) - f(\x \odot (1-\boldsymbol{m}_i))
$$
where $\boldsymbol{m}_i$ is an indicator vector for the patch or feature to remove, and `$\odot$' denotes the element-wise product.
A heatmap $(R_i)_i$ can be built from these scores highlighting locations where the occlusion has caused the strongest decrease of the function.
Because occlusion may produce visual artefacts, inpainting occluded patterns (e.g.\ using a generative model \cite{Agarwal_2020_ACCV}) rather than setting them to gray was proposed as an enhancement.

Attribution based on Shapley values \cite{DBLP:conf/nips/LundbergL17,DBLP:journals/jmlr/StrumbeljK10} (see Section \ref{sec:shapley} for a definition), can also be seen as an occlusion analysis. Here, instead of occluding features one at a time, a much broader set of occlusion patterns are considered, and this has the effect of also integrating global effects in the explanation. SHAP and Kernel SHAP \cite{DBLP:conf/nips/LundbergL17} are practical algorithms to approximate Shapley values, that sample a few occlusions according to the probability distribution used to compute Shapley values, and then fit a linear surrogate model that correctly predicts the effect of these occlusions on the output. An explanation can then be easily extracted, and this explanation retains some similarity with the original Shapley values.

A further extension of occlusion analysis is Meaningful Perturbation \cite{fong2017interpretable}, where an occluding pattern is synthesized, subject to a sparsity constraint, in order to engender the maximum drop of the function value $f$. The explanation is then readily given by the synthesized pattern. The perturbation-based approach was latter embedded in a rate distortion theoretical framework \cite{macdonald2019rate}.

%%%%%%%%%%%
\subsection{Integrated Gradients / SmoothGrad \cite{SundararajanTY17,DBLP:journals/corr/SmilkovTKVW17}} 
\label{section:gradients}

\emph{Integrated Gradients}~\cite{SundararajanTY17} explains by integrating the gradient $\nabla f(\x)$ along some trajectory in input space connecting some root point $\widetilde{\x}$ to the data point $\x$. The integration process addresses the problem of locality of the gradient information (cf.\ Section \ref{section:difficult}), making it well-suited for explaining functions that have multiple scales. In the simplest form, the trajectory is chosen to be the segment $[\widetilde{\x},\x]$ connecting some root point to the data. Integrated gradients defines feature-wise scores as:
$$
R_i(\x) = (x_i-\widetilde{x}_i) \cdot \int_0^1 [\nabla f(\widetilde{\x} + t \cdot (\x-\widetilde{\x}))]_i \,dt
$$
It can be shown that these scores satisfy $\sum_i R_i(\x) = f(\x)$ and thus constitute a {\em complete} explanation. If necessary, the method can be easily extended to any trajectories in input space. For implementation purposes, integrated gradients must be discretized. Specifically, the continuous trajectory is approximated by a sequence of data points $\x^{(1)}, \dots, \x^{(N)}$. Integrated gradients is then implemented as shown in Algorithm~\ref{alg:integrated-gradients}.

\begin{algorithm}
\caption{Integrated Gradients}
\begin{algorithmic}
\STATE $\R = \boldsymbol{0}$
\FOR{$n=1$ \textbf{to} $N-1$}
\STATE $\R = \R + \nabla f(\x^{(n)}) \odot (\x^{(n+1)} - \x^{(n)})$
\ENDFOR
\RETURN $\R$
\end{algorithmic}
\label{alg:integrated-gradients}
\end{algorithm}

\noindent The gradient can easily be computed using automatic differentiation.
The larger the number of discretization steps, the closer the output gets to the integral form, but the more computationally expensive the procedure gets.

\medskip

Another popular gradient-based explanation method is \emph{SmoothGrad}~\cite{DBLP:journals/corr/SmilkovTKVW17}. The function's gradient is averaged over a large number of locations corresponding to small random perturbations of the original data point $\x$:
\begin{align*}
\nabla_\text{smooth} f(\x) &= \mathbb{E}_{\boldsymbol{\varepsilon} \sim \mathcal{N}(\boldsymbol{0},\sigma^2 I)} [\nabla f(\x + \boldsymbol{\varepsilon})]
\end{align*}
Like the method's name suggests, the averaging process `smoothes' the explanation, and in turn also addresses the shattered gradient problem described in Section \ref{section:difficult}.  (See also \cite{morch1995visualization,baehrens2010explain,DBLP:journals/corr/SimonyanVZ13} for earlier gradient-based explanation techniques).

\smallskip

In Section \ref{section:comparison}, we experiment with a combination of Integrated Gradients and SmoothGrad \cite{DBLP:journals/corr/SmilkovTKVW17}, similar to Expected Gradients (cf.\ \cite{Sturmfels2020}), where relevance scores obtained from Integrated Gradients are averaged over several integration paths that are drawn from some random distribution. The resulting method preserves the advantages of Integrated Gradients and further reduces the gradient noise.

%%%%%%%%%%%
\subsection{Layer-Wise Relevance Propagation (LRP) \cite{BachPLOS15}}
\label{section:lrp}

The \emph{Layer-wise Relevance Propagation} (LRP) method~\cite{BachPLOS15} makes explicit use of the layered structure of the neural network and operates in an iterative manner to produce the explanation. Consider the neural network
$$
f(\x) = f_L \circ \dots \circ f_1(\x)
$$
First, activations at each layer of the neural network are computed until we reach the output layer. The activation score in the output layer forms the prediction. Then, a reverse propagation pass is applied, where the output score is progressively redistributed, layer after layer, until the input variables are reached. The redistribution process follows a conservation principle analogous to Kirchoff's laws in electrical circuits. Specifically, all `relevance' that flows into a neuron at a given layer flows out towards the neurons of the layer below. At a high level, the LRP procedure can be implemented as a forward-backward loop, as shown in Algorithm \ref{algorithm:lrp}.

\begin{figure}[ht]
    \centering
    \includegraphics[width=0.95\linewidth]{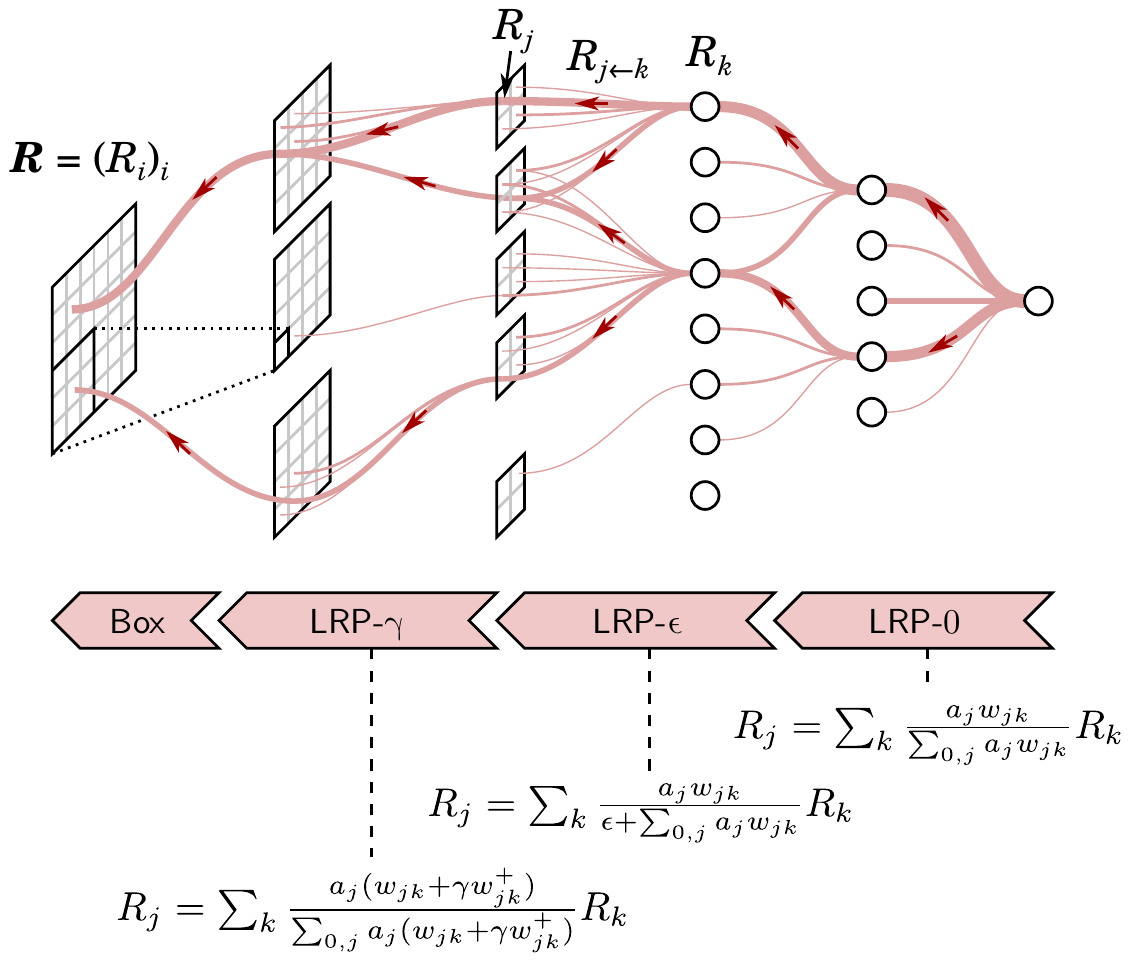}
        \caption{Illustration of the LRP propagation procedure applied to a neural network. The prediction at the output is propagated backward in the network, using various propagation rules, until the input features are reached. The propagation flow is shown in red.}
    \label{fig:lrp}
\end{figure}

\begin{algorithm}
\caption{Layer-wise Relevance Propagation}
\label{algorithm:lrp}
\begin{algorithmic}
\STATE $\ba^{(0)} = \x$
\FOR{$l=1 \dots L$}
\STATE $\ba^{(l)} = f_l(\ba^{(l-1)})$
\ENDFOR
\STATE $\R^{(L)} = \ba^{(L)}$
\FOR{$l=L \dots 1$}
\STATE $\R^{(l-1)} = \texttt{relprop}(\ba^{(l-1)},\R^{(l)}, f_l)$
\ENDFOR
\RETURN $\R^{(0)}$
\end{algorithmic}
\end{algorithm}
The function \texttt{relprop} performs redistribution from one layer to the layer below and is based on `propagation rules' defining the exact redistribution policy. Examples of propagation rules are given later in this section, and their implementation is provided in Appendix \ref{appendix:lrp}. The LRP procedure is shown graphically in Fig.\ \ref{fig:lrp}.

While LRP can in principle be performed in any forward computational graph, a class of neural networks which is often encountered in practice, and for which LRP comes with efficient propagation rules that can be theoretically justified (cf.\ Section \ref{section:theory}) is deep rectifier networks \cite{DBLP:journals/jmlr/GlorotBB11}. The latter can be in large part abstracted as an interconnection of neurons of the type:
$$
\textstyle a_k = \max\big(0\,,\,\sum_{0,j} a_j w_{jk}\big),
$$
where $a_j$ denotes some input activation, and $w_{jk}$ is the weight connecting neuron $j$ to neuron $k$ in the layer above. The notation $\sum_{0,j}$ indicates that we sum over all neurons $j$ in the lower layer plus a bias term $w_{0k}$ with $a_0=1$. For this class of networks, various propagation rules have been proposed (cf.\ Fig.\ \ref{fig:lrp}). For example, the LRP-$\gamma$ rule \cite{DBLP:series/lncs/MontavonBLSM19} defined as
\begin{align}
R_j = \sum_k \frac{a_j (w_{jk} + \gamma w_{jk}^+)}{\sum_{0,j} a_j (w_{jk} + \gamma w_{jk}^+)} R_k
\label{eq:lrpgamma}
\end{align}
redistributes based on the contribution of lower-layer neurons to the given neuron activation, with a preference for positive contributions over negative contributions. This makes it particularly robust and suitable for the lower-layer convolutions. Other propagation rules such as LRP-$\epsilon$ or LRP-0 are suitable for other layers \cite{DBLP:series/lncs/MontavonBLSM19}. Additional propagation rules have been proposed for special layers such as min/max pooling \cite{BachPLOS15,DBLP:journals/pr/MontavonLBSM17,Kauffmann20} and LSTM blocks \cite{ArrWASSA17, DBLP:series/lncs/ArrasAWMGMHS19}. Furthermore, a number of other propagation techniques have been proposed \cite{shrikumar2016not,shrikumar2017learning,LandeckerTBMKB13,zhang2018neural} with some of the rules overlapping with LRP for certain choices of parameters. For a technical overview of LRP including a discussion of the various propagation rules and further recent heuristics, see~\cite{DBLP:series/lncs/MontavonBLSM19}.

An inspection of Eq.\ \eqref{eq:lrpgamma} shows an important property of LRP, that of conserving relevance from layer to layer, in particular, we can show that in absence of bias terms, $\sum_j R_j = \sum_k R_k$. A further interesting property of this propagation rule is `smoothing': Consider the relevance can be written as $R_j = a_j c_j$ and $R_k = a_k c_k$ a product of activations and factors. Those factors can be directly related by the equation
\begin{align}
c_j = \sum_k (w_{jk} + \gamma w_{jk}^+) \frac{\max(0,\sum_{0,j} a_j w_{jk})}{\sum_{0,j} a_j (w_{jk} + \gamma w_{jk}^+)} c_k.
\label{eq:modgradient}
\end{align}
This equation can be interpreted as a smooth variant of the chain rule for derivatives used for computing the neural network gradient \cite{DBLP:series/lncs/Montavon19}. Thus, analogous to SmoothGrad \cite{DBLP:journals/corr/SmilkovTKVW17}, LRP also performs some gradient smoothing, however, it embeds it tightly into the deep architecture, so that only a single backward pass is required. In addition to smoothing, Eq.\ \eqref{eq:modgradient} can also be interpreted as a gradient that has been biased to positive values, an idea also found in methods such as
\emph{DeconvNet}~\cite{zeiler2014visualizing} or \emph{Guided Backprop}~\cite{springenberg2015striving}. This modified gradient view on LRP can also be leveraged to achieve a simpler and more general implementation of LRP based on `forward hooks', which we describe in the second part of Appendix \ref{appendix:lrp}, and which we use to apply LRP on \mbox{VGG-16}~\cite{DBLP:journals/corr/SimonyanZ14a} and \mbox{ResNet-50}~\cite{DBLP:conf/cvpr/HeZRS16} in Section~\ref{section:comparison}.

\subsection{Other Methods}
\label{section:others}

We discuss in this section several other popular Explainable AI approaches, that either do not fall in the category of post-hoc explanation approaches (and therefore are not covered in the sections above), that are specialized for a particular neural network architecture, or that make use of different units of interpretability than the input features.

\smallskip

In contrast to the discussed post-hoc methods that apply to any DNN model, \textit{self-explainable models} are designed from scratch with interpretability in mind. A self-explainable model can either be trained to solve a machine learning task directly from a supervised dataset, or it can be used to approximate a black-box model on some representative input distribution. Examples of self-explainable models include simple linear models, or specific nonlinear models, e.g.\ neural networks with an explicit top-level sum-pooling structure \cite{PoulinESLGWFPMA06, lin2014network, caruana2015intelligible, zhou2016learning, DBLP:conf/iclr/BrendelB19}. In all of these models, each summand is linked only to one of a few input variables, which makes attribution of their prediction on the input variables straightforward. More complex architectures involving \textit{attention mechanisms} were also proposed \cite{DBLP:conf/nips/LarochelleH10, DBLP:journals/corr/BahdanauCB14, DBLP:conf/icml/XuBKCCSZB15}, and inspection of the attention mechanism itself can also deliver useful insights into the model prediction. While self-explainable models can be useful for many real-world tasks (a list of arguments in favor of these models can be found e.g.\ in \cite{Rudin2019}), their applicability becomes more limited when the goal is to explain the strategy of some existing black-box model. In such scenario, one would have to achieve the difficult task of closely replicating the black-box model for every possible input and perturbation of it, while at the same time being constrained by the predefined interpretable structure.

Other methods are \textit{specialized} for a particular deep neural network model for which generic explanation methods do not provide a direct solution. One such model is the graph neural network \cite{DBLP:journals/tnn/ScarselliGTHM09,DBLP:conf/iclr/KipfW17}, where the graph adjacency matrix given as input does not appear as it is usually the case in the first layer, but instead at every layer. Methods that have been proposed to explain these particular neural networks include the GNNExplainer \cite{DBLP:conf/nips/YingBYZL19}, or GNN-LRP \cite{Schnake2020Graphs}. Other neural network architectures have a more conventional structure but still require a non-trivial adaptation of existing explanation methods, for example, extensions of LRP have been proposed to deal with the special LSTM blocks in recurrent neural networks \cite{ArrWASSA17,DBLP:series/lncs/ArrasAWMGMHS19} or to handle attention units in the context of neural machine translation \cite{ding2017visualizing}.

Further methods do not seek to explain in terms of input features, but in terms of the latent space, where the directions in the latent space code for higher-level concepts, such as color, material, object part, or object \cite{zhou2018interpreting, bau2019visualizing,Bau2020}. In particular, the TCAV method \cite{DBLP:conf/icml/KimWGCWVS18} produces a latent-space explanation for every individual prediction. Some techniques integrate multiple levels of abstraction (e.g.\ different layers of the neural networks), to arrive at more informative explanation of the prediction process \cite{Zhang2020ExGraph, DBLP:journals/pami/SimonRDD20,Schnake2020Graphs}. Finally, generative approaches have been proposed to build structured textual explanations of a machine learning model \cite{DBLP:conf/eccv/HendricksARDSD16, DBLP:conf/acl/LiuYW19}.

%%%%%%%%%%%%%%%%%%%%%%%%%%%%%%%%%%%%%%%%%%%%%%%%%%%%%%%%%%%%%%%%%%%%%%%%%%%%%%%%%%%%
\section{Comparing Explanation Methods}
\label{section:comparison}
\begin{figure}[ht]
    \centering
    \includegraphics[width=1.0\linewidth]{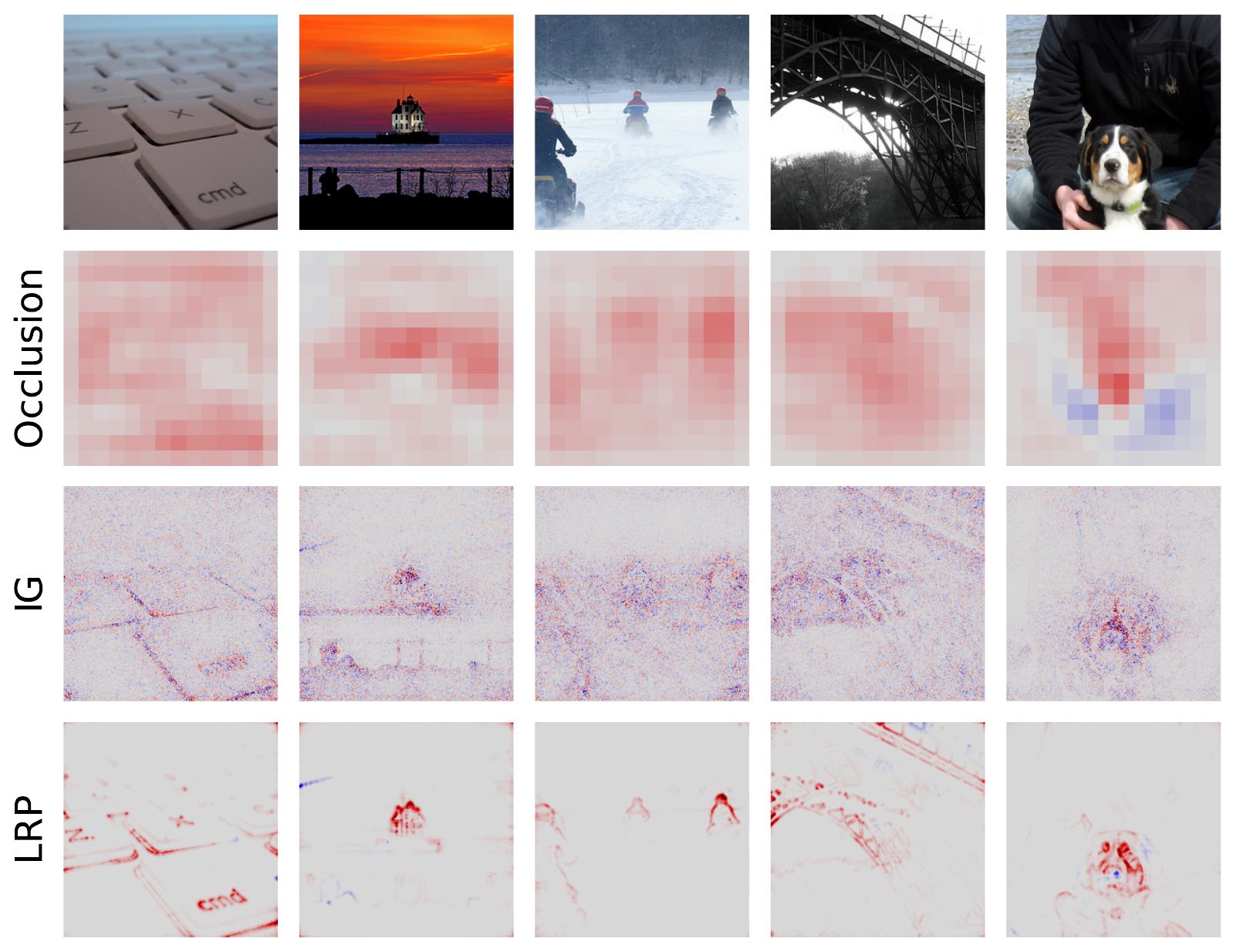}
    \caption{Examples of images from ImageNet \cite{DBLP:journals/ijcv/RussakovskyDSKS15} with classes `space bar`, `beacon/lighthouse`, `snow mobile`, `viaduct`, `greater swiss mountain dog`. Images are correctly predicted by the VGG-16 \cite{DBLP:journals/corr/SimonyanZ14a} neural network, and shown along with an explanation of the predictions. Different explanation methods lead to different qualities of explanation.}
    \label{fig:heatmaps}
\end{figure}

The methods presented in Section \ref{section:methods} highlight the variety of approaches available for attributing the prediction of a deep neural network to its input features. This variety of techniques also translates into a variety of qualities of explanations. Illustrative examples of images and the explanation of predicted evidence for the ground truth class as produced by the different explanation methods are shown in Fig.\ \ref{fig:heatmaps}. Occlusion Analysis is performed by occluding patches of size $32 \times 32$ pixels with stride $16$. Integrated Gradients performs $5$ integration steps starting from $5$ random points near the origin in order to add smoothing (cf.\ Appendix \ref{appendix:igs}), resulting in $25$ function evaluations. LRP explanations are obtained by applying the same LRP rules as in \cite{DBLP:series/lncs/MontavonBLSM19}. We observe the following qualitative properties of the explanations: Occlusion-based explanations are coarse and are indicating relevant regions rather than the relevant pixel features. Integrated Gradients produces very fine pixel-wise explanations containing both substantial amounts of evidence in favor and against the prediction (red and blue pixels). LRP preserves the fine explanation structure but tends to produce less negative scores and attributes relevance to whole features rather than individual pixels.

In practice, it is important to reach an objective assessment of how good an explanation is. Unfortunately, evaluating explanations is made difficult by the fact that it is generally impossible to collect `ground truth' explanations. Building such ground truth explanations would indeed require the expert to understand how the deep neural network decides.

Standard machine learning models are usually evaluated by the utility (expected risk) of their decision behavior (e.g.\ \cite{vapnik95}). Transposing this concept of maximizing utility to the domain of explanation, quantifying the utility of the explanation would first require to define what is the ultimate target task (the explanation being only an intermediate step), and then assessing by how much the use of explanation by the human increases its performance on the target task, compared to not using it (see e.g.\ \cite{baehrens2010explain,hansen2011visual,doshi2017towards,DBLP:series/lncs/11700}). Because such end-to-end evaluation schemes are hard to set up in practice, general desiderata for ML explanations have been proposed \cite{Swartout1993,DBLP:journals/ai/Miller19}. Common ones include (1) faithfulness\,/\,sufficiency (2) human-interpretability, and (3) possibility to practically apply it to an ML model or an ML task (e.g.\ algorithmic efficiency of the explanation algorithm).

%%%%%%%%%%%
\subsection{Faithfulness\,/\,Sufficiency}
A first desideratum of an explanation is to reliably and comprehensively represent the local decision structure of the analyzed ML model. A practical technique to assess such property of the model is `pixel-flipping' \cite{samek2016evaluating}. The pixel-flipping procedure tests whether removing the features highlighted by the explanation (as most relevant) leads to a strong decay of the network prediction abilities. The procedure is summarized in Algorithm \ref{algorithm:pixelflipping}.

\begin{algorithm}
\caption{Pixel-Flipping}
\label{algorithm:pixelflipping}
\begin{algorithmic}
\STATE \texttt{pfcurve} = [\,]
\FOR{$p$ \textbf{in} $\mathrm{argsort}(-\boldsymbol{R})$}
    \STATE $\x \leftarrow \x - \{x_p\}$ (remove pixel $p$ from the image).
    \STATE \texttt{pfcurve.append}$(f(\x))$.
\ENDFOR
\RETURN \texttt{pfcurve}
\end{algorithmic}
\end{algorithm}

Pixel-flipping runs from the most to the least relevant input features, iteratively removing them and monitoring the evolution of the neural network output. The series of recorded decaying prediction scores can be plotted as a curve. The faster the curve decreases, the more faithful the explanation method is w.r.t.\ the decision of the neural network. The pixel-flipping curve can be computed for a single example, or averaged over a whole dataset in order to get a global estimate of the faithfulness of an explanation algorithm under study.

\begin{figure}[ht]
    \centering
    \includegraphics[width=1.0\linewidth]{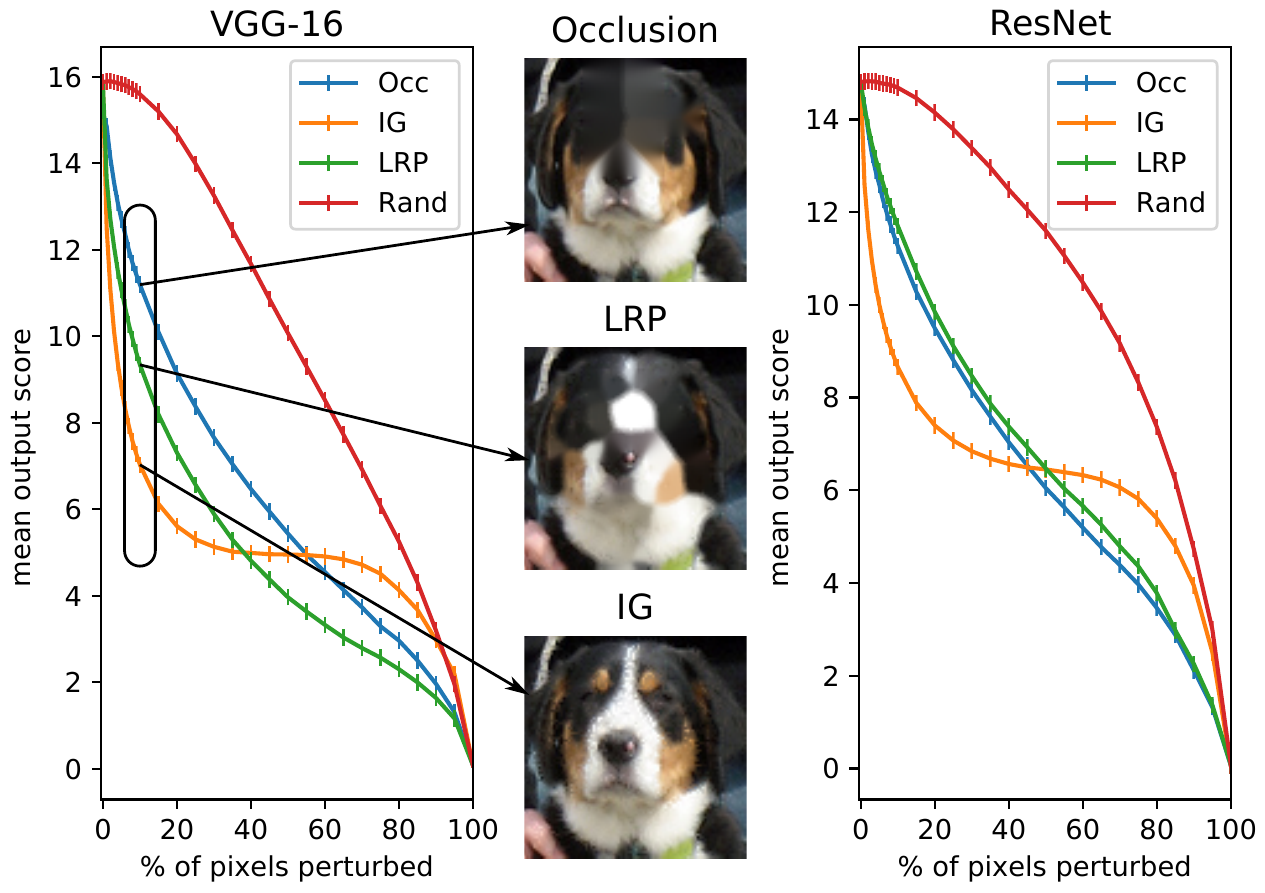}
    \caption{Pixel-flipping experiment for testing faithfulness of the explanation. We remove pixels found to be the most relevant by each explanation method and verify how quickly the output of the network decreases.}
    \label{fig:pixelflipping}
\end{figure}

Fig.\ \ref{fig:pixelflipping} applies pixel-flipping to the three considered explanation methods and on two models: VGG-16 \cite{DBLP:journals/corr/SimonyanZ14a} and \mbox{ResNet-50} \cite{DBLP:conf/cvpr/HeZRS16}. At each step of pixel-flipping, removed pixels are imputed using a simple inpainting algorithm, which avoids introducing visual artefacts in the image.

We observe that for all explanation methods, removing relevant features quickly destroys class evidence. In particular, they perform much better than a random explanation baseline. Fine differences can however be observed between the methods: For example, LRP performs better on VGG-16 than on ResNet-50. This can be explained by VGG-16 having a more explicit structure (standard pooling operations for VGG-16 vs.\ strided convolution for ResNet-50), which better supports the process of relevance propagation (see also \cite{rieger18} for a discussion of the effect of structure on the performance of explanation methods).

A second observation in Fig.\ \ref{fig:pixelflipping} is that Integrated Gradients has by far the highest decay rate initially but stagnates in the later phase of the pixel-flipping procedure. The reason for this effect is that IG focuses on pixels to which the network is the most sensitive, without however being able to identify fully comprehensively the relevant pattern in the image. This effect is illustrated in Fig.\ \ref{fig:pixelflipping} (middle) on a zoomed-in exemplary image of class `greater swiss mountain dog', where the image after 10\% flipping has lost most of its prediction score, but visually appears almost intact. Effectively, IG has built an adversarial example \cite{DBLP:journals/corr/SzegedyZSBEGF13,DBLP:conf/cvpr/NguyenYC15}, i.e.\ an example whose visual content clearly disagrees with the prediction at the output of the network. We note that Occlusion and LRP do not run into such adversarial examples. For these methods, pixel-flipping steadily and comprehensively removes features until class evidence has totally disappeared.

Overall, the pixel-flipping algorithm characterizes various aspects of the faithfulness of an explanation method. We note however that faithfulness of an explanation does not tell us how easy it will be for a human to make sense of that explanation. We address this other key requirement of an explanation in the following section.

%%%%%%%%%%%
\subsection{Human Interpretability}
Here, we discuss whether the presented explanation techniques deliver results that are meaningful to the human, i.e.\ whether the human can gain understanding into the classifier's decision strategy from the explanation. Human interpretability is hard to define in general \cite{DBLP:journals/ai/Miller19}. Different users may have different capabilities at reading explanations and at making sense of the features that support them \cite{ribeiro2016should,Naranyan2018}. For example, the layman may wish for a visual interpretation, even approximate, whereas the expert may prefer an explanation supported by a larger vocabulary, including precise scientific or technical terms \cite{baehrens2010explain}.

For the image classification setting, interpretability can be quantified in terms of the amount of information contained in the heatmap (e.g.\ as measured by the file size). An explanation with a small associated file size is more likely to be interpretable by a human. The table below shows average file sizes (in bytes\footnote{JPEG compression using the Pillow image processing library for python with a quality setting of 75/100 (standard settings).}) associated to the various explanation techniques and for two neural networks.

\begin{center}
\small
\begin{tabular}{l|ccc}\toprule
 & Occ & IG & LRP\\\midrule
 VGG-16 & {\bf 698.4} & 5795.0 & 1828.3\\
 ResNet-50 & {\bf 693.6} & 5978.0 & 2928.2\\\bottomrule
 \end{tabular}
\end{center}

We observe that occlusion produces the lowest file size and is therefore the most `interpretable'. It indeed only presents to the user rough localization information without going into the details of which exact feature has supported the decision as done e.g.\ by LRP. On the other side of the interpretability spectrum we find Integrated Gradients. In the explanations this last method produces, every single pixel contains information, and this makes it clearly overwhelming to the human.

\smallskip

In practice, neural networks do not need to be explained in terms of input features. For example, the TCAV method~\cite{DBLP:conf/icml/KimWGCWVS18} considers directional derivatives in the space of activations (where the directions correspond to higher-level human-interpretable concepts) in place of the input gradient. Similar higher-level interpretations are also possible using the Occlusion and LRP methods, respectively by perturbing groups of activations corresponding at a given layer to a certain concept, or by stopping the LRP procedure at the same layer and pooling scores on some group of neurons representing the desired concept.

%%%%%%%%%%%
\subsection{Applicability and Runtime}
Faithfulness and interpretability do not fully characterize the overall usefulness of an explanation method. To characterize usefulness, we also need to determine whether the explanation method is applicable to a range of models that is sufficient large to include the neural network model of interest, and whether explanations can be obtained quickly with finite compute resources.

\emph{Occlusion}-based explanations are the easiest to implement. These explanations can be obtained for any neural network, even those that are not differentiable. This also includes networks for which we do not have the source code and where we can only access their prediction through some online server. Technically, occlusion can therefore be used to understand the predictions of third-party models such as \texttt{https://cloud.google.com/vision/} and \texttt{https://www.clarifai.com/models}. \emph{Integrated gradients} requires instead for each prediction an access to the neural network gradient. Given that most machine learning models are differentiable, this method is widely applicable also for neural networks with complex structures, such as ResNets \cite{DBLP:conf/cvpr/HeZRS16} or SqueezeNets \cite{DBLP:journals/corr/IandolaMAHDK16}.
Integrated Gradients is also easily implemented in state-of-the-art ML frameworks such as PyTorch or TensorFlow, where we can make use of automatic differentiation.
LRP assumes that the model is structured as (or can be converted to \cite{Kauffmann19,Kauffmann20}) a neural network with a canonical sequence of layers, for example, an alternation of linear/convolution layers, ReLU layers, and pooling layers.
This stronger requirement and the implementation overhead caused by explicitly accessing the different layers (cf.\ Appendix \ref{appendix:lrp}) will however be offset by a last characteristic we consider in this section, which is the computational cost associated producing the explanation. A runtime comparison\footnote{Explanations are computed in batches of (up to) 16 samples on a GPU and with explanation techniques implemented in PyTorch. Results are averaged over 10 repetitions.} of the three explanation methods studied here is given in the table below (measured in {\em explanations per second}).

%\begin{center}
%\small
%\begin{tabular}{lccc}\toprule
% & Occlusion & IG & LRP\\\midrule
% VGG-16 & 42.4 & 17.1 & {\bf 0.49}\\
% ResNet-50 & 24.9 & 11.5 & {\bf 0.53}\\\bottomrule
% \end{tabular}
%\end{center}
%%%%%%%%%%%%
%% applied 100/x
%%%%%%%%%%%%

\begin{center}
\small
\begin{tabular}{l|ccc}\toprule
 & Occ & IG & LRP\\\midrule
 VGG-16 & 2.4 & 5.8 & {\bf 204.1}\\
 ResNet-50 & 4.0 & 8.7 & {\bf 188.7}\\\bottomrule
 \end{tabular}
\end{center}

Occlusion is the slowest method as it requires to reevaluate the function for each occluded patch. For image data, the runtime of Occlusion increases quadratically with the step size, making the obtainment of high-resolution explanations with this method computationally prohibitive. Integrated Gradients inherits pixel-wise resolution from the gradient computation which is $O(1)$ but requires multiple iterations for the integration. The runtime is further increased if performing an additional loop of smoothing. LRP is the fastest method in our benchmark by an order of magnitude. The LRP runtime is only approximately three times higher than that of computing a single forward pass. This makes LRP particularly convenient for the large-scale analyses we introduce in Section \ref{section:datasetwide} where an explanation needs to be produced for every single example in the dataset.

%%%%%%%%%%%%%%%%%%%%%%%%%%%%%%%%%%%%%%%%%%%%%%%%%%%%%%%%%%%%%%%%%%%%%%%%%%%%%%%%%%%%
\section{Theoretical Foundations of Explanation Methods}
\label{section:theory}
In parallel to developing explanation methods that address application requirements such as faithfulness, interpretability, usability and runtime, some works have focused on building theoretical foundations for the problem of explanation \cite{DBLP:journals/pr/MontavonLBSM17, DBLP:conf/nips/LundbergL17} and establishing theoretical connections between the different methods \cite{shrikumar2016not, ancona2018towards, montavon2018methods}.

Here, we present three frameworks: the Shapley Values \cite{Shapley,DBLP:journals/jmlr/StrumbeljK10,DBLP:conf/nips/LundbergL17} which comes from game theory, the Taylor expansions \cite{BachPLOS15,Bazen2013}, and the Deep Taylor Decomposition~\cite{DBLP:journals/pr/MontavonLBSM17}, which applies Taylor expansions repeatedly at each layer of a DNN. We then show how Occlusion, Integrated Gradients, or LRP intersect for certain choices of parameters with these mathematical approaches.

\subsection{Shapley Values}
\label{sec:shapley}

Shapley values \cite{Shapley} is a framework originally proposed in the context of game theory to determine individual contributions of a set of cooperating players $\mathcal{P}$. The method considers every subset of cooperating players $\mathcal{S} \subseteq \mathcal{P}$ and tests the effect of removing/adding the player $i$ to $\mathcal{S}$ on the total payoff $v(\mathcal{S})$ obtained by $\mathcal{S}$ if they cooperate. Specifically, Shapley values identify the contribution of player $i$ to the overall coalition $\mathcal{P}$ to be:
$$
\phi_i =\sum_{\mathcal{S} \subseteq \mathcal{P} \setminus \{i\}} \alpha_\mathcal{S} \cdot (v(\mathcal{S} \cup \{i\}) - v(\mathcal{S}))
$$
where each subset $\mathcal{S}$ is weighted by the factor $\alpha_\mathcal{S} =  |\mathcal{S}|! \cdot (|\mathcal{P}| - 1 - |\mathcal{S}|)! / |\mathcal{P}|!$. Shapley values satisfy a number of axioms, in particular, efficiency ($\sum_i \phi_i = v(\mathcal{P}$)), symmetry, linearity, and zero added value of a dummy player. Shapley values are in fact the unique assignment strategy that jointly satisfies these axioms \cite{Shapley}.

\medskip

When transposing the method to the task of explaining a machine learning model \cite{DBLP:journals/jmlr/StrumbeljK10,DBLP:conf/nips/LundbergL17}, the players of the cooperating game become the input features, and the payoff function becomes related to the DNN output. In $\cite{DBLP:conf/nips/LundbergL17}$, the payoff function is chosen to be the conditional expectation: $v(\mathcal{S}) = \mathbb{E}[f(\x)\,|\,\x_\mathcal{S}]$. Alternately, to make the score depend only on the model without assuming a specific input distribution, the payoff function can be set to 
\begin{align}
v(\mathcal{S}) = f(\x_\mathcal{S}),
\label{eq:shapley-value}
\end{align}
i.e.\ input features not in $\mathcal{S}$ are set to zero (see e.g.\ \cite{pmlr-v97-ancona19a}), and we will use this formulation to make connections to practical explanation methods in Section \ref{section:embedding}. Note that Shapley values make almost no assumptions about the structure of the function $f$ and can therefore serve as a general theoretical framework to analyze explanation methods. For specific functions, e.g.\ additive models of the type $f(\x) = \sum_{i} f_i(x_i)$, Shapley values (using Eq.\ \eqref{eq:shapley-value}) take the simple form $\phi_i = f_i(x_i)$.

%%%%%%%%%%%
\subsection{Taylor Decomposition}
\label{section:taylor}
Taylor expansions are a well-known mathematical framework to decompose a function into a series of terms associated to different degrees and combinations of input variables. Unlike Shapley values which evaluates the function $f(\x)$ multiple times, the Taylor expansion framework for explaining a ML model \cite{BachPLOS15,Bazen2013,Eberle2020Similarity} evaluates the function once at some reference point $\widetilde{\x}$ and assigns feature contributions by locally extracting the gradient (and higher-order derivatives). Specifically, the Taylor expansion of some smooth and differentiable function $f:\mathbb{R}^d \to \mathbb{R}$ at some reference point $\widetilde{\x}$ is given by:
\begin{align*}
\textstyle f(\x) &= \textstyle f(\widetilde{\x}) \\
&\textstyle \quad + \sum_i [\nabla f(\widetilde{\x})]_i \cdot (x_i - \widetilde{x}_i)\\
&\textstyle \quad\quad +  \frac12 \sum_{ii'} [\nabla^2f(\widetilde{\x})]_{ii'} (x_i - \widetilde{x}_i) (x_{i'} - \widetilde{x}_{i'})\\
&\textstyle \quad\quad\quad+ \dots
\end{align*}
where $\nabla f$ and $\nabla^2f$ denote the gradient and the Hessian respectively, and $\dots$ denote the (non-expanded) higher-order terms. The zero-order term is the function value at the reference point and is zero if choosing a root point. There are as many first-order terms as there are dimensions and each of them is bound to a particular input variable. Thus, they offer a natural way of attributing a function value $f(\x)$ onto individual linear components. There are as many second-order terms as there are pairs of ordered variables, and even more third-order and higher-order terms. When the function is approximately locally linear, second and higher-order terms can be ignored, and we get the following simple attribution scheme:
$$
R_i  = [\nabla f(\widetilde{\x})]_i \cdot (x_i - \widetilde{x}_i)~,
$$
a product of the gradient and the input relative to our root point. In the general case, there are no closed-form approach to find the root point and it is instead obtained using an optimization technique.

%%%%%%%%%%%
\subsection{Deep Taylor Decomposition}
\begin{figure}[ht]
    \centering
    \includegraphics[width=1.0\linewidth]{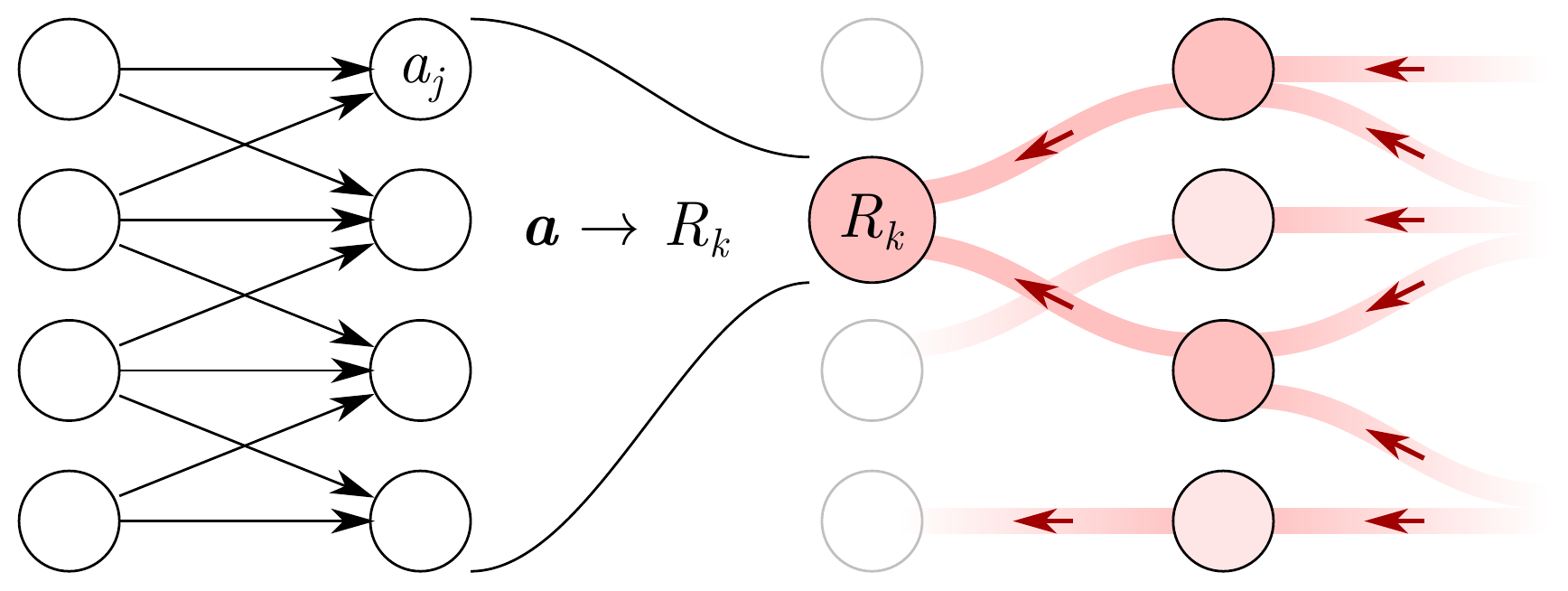}
    \caption{Graphical illustration of the function $R_k(\ba)$ that DTD seeks to decompose on the input dimensions. Because $R_k$ is complex, it is often replaced by an analytically more tractable model $\widehat{R}_k(\ba)$ that only depends on local activations.}
    \label{fig:dtd}
\end{figure}

An alternate way of formalizing the problem of attribution of a function onto input features is offered by the recent framework of Deep Taylor Decomposition (DTD)~\cite{DBLP:journals/pr/MontavonLBSM17}. Deep Taylor Decomposition assumes the function is structured as a deep neural network and seeks to attribute the prediction onto input features by performing a Taylor decomposition at every neuron of each layer instead of directly on the whole neural network function. Deep Taylor decomposition assumes the output score has already been attributed onto some layer of activations $(a_k)_k$ and attribution scores are denoted by $R_k$. Deep Taylor Decomposition then considers the function $R_k(\ba)$ where $\ba = (a_j)_j$ is the collection of neuron activations in the layer below. These quantities are illustrated in Fig.\ \ref{fig:dtd}.

The function $R_k(\ba)$ is typically very complex as it corresponds to a composition of multiple forward and backward computations. This function can however be approximated locally by some `relevance model' $\widehat{R}_k(\ba)$, the choice of which will depend on the method we have used for computing $R_k$. We then compute a Taylor expansion of this function:
\begin{align*}
\textstyle \widehat{R}_k(\ba) &= \widehat{R}_k(\widetilde{\ba})\\
&\textstyle \quad + \sum_j [\nabla \widehat{R}_k(\widetilde{\ba})]_j \cdot (a_j - \widetilde{a}_j)\\
&\textstyle \quad\quad+ \dots
\end{align*}
The linear terms define `messages' $R_{j \leftarrow k}$ that can be redistributed to neurons in the lower layer, and messages received by a given neuron at a certain layer are summed to form a total relevance score:
\begin{align}
\textstyle R_j = \sum_k [\nabla \widehat{R}_k(\widetilde{\ba}^{(k)})]_j \cdot (a_j - \widetilde{a}_j^{(k)})
\label{eq:dtd}
\end{align}
here, we have added an index $\{\}^{(k)}$ to the root point to make explicit that different root points can be used for expanding different neurons. The redistribution procedure is iterated from the top layer towards the lower layers, until the input features are reached.

%%%%%%%%%%%
\subsection{Connections between Explanation Methods}
\label{section:embedding}

Having described the Shapley values and the simple and deep Taylor decomposition frameworks, we now present some results from the literature showing how some explanation methods reduce for certain choices of parameters to these frameworks. The different connections we outline here are summarized in Table~\ref{tab:theory}.

% \begin{figure}[ht]
%     \centering
%     \includegraphics[width=1.0\linewidth]{figures/tree.pdf}
%     \caption{Relation between explanation methods and Taylor decomposition / Deep Taylor Decomposition (DTD), for certain choices of hyperparameters and models.}
%     \label{fig:theory}
% \end{figure}

\begin{table}[ht]
    \caption{Examples of explanation methods applied with different parameters on different models, and whether they can be embedded in each of the three presented theoretical frameworks.}
    \centering 
    \begin{tabular}{lccc}\toprule
        & Shapley & Taylor & DTD\\\midrule
        \textit{Linear models}\\
        ~~Occlude-1 & $\checkmark$ & $\checkmark$ & \color{gray} ($\checkmark$)\\
        ~~IG on $\text{Segment}(\boldsymbol{0},\x)$ & $\checkmark$ & $\checkmark$ &\color{gray}  ($\checkmark$) \\
        ~~LRP-0 & $\checkmark$ & $\checkmark$ & \color{gray} ($\checkmark$) \\[1mm]
        \textit{Nonlinear additive models}\\
        ~~Occlude-1 & $\checkmark$ & \\[1mm]
        \textit{Deep rectifier networks}\\
        ~~Occlude-1 & & & \\
        ~~IG on $\text{Segment}(\boldsymbol{0},\x)$ & & $\checkmark$ & \\
        ~~LRP-0 & & $\checkmark$ & $\checkmark$\\
        ~~LRP-$\epsilon$/$\gamma$/$\dots$ & & & $\checkmark$\\
        \bottomrule
    \end{tabular}
    \label{tab:theory}
\end{table}

We start by connecting occlusion-based explanations of a linear model to Shapley values and Taylor decomposition.

\begin{proposition} When applied to homogeneous linear models (of the type $f(\x) = \w^\top \x$), occlusion with patch size $1$ and replacement value $0$ is equivalent to a Taylor decomposition with root point $\widetilde{\x} = \boldsymbol{0}$, as well as Shapley values with value function given by Eq.\ \eqref{eq:shapley-value}.
\label{proposition:occlusion}
\end{proposition}

The first connection is shown by the chain of equations $f(\x) - f(\x - \{x_i\}) = w_i x_i = [\nabla f(\boldsymbol{0})]_i \cdot (x_i - 0)$. For the Shapley values, we simply observe $\phi_i = \sum_{\mathcal{S} \subseteq \mathcal{P} \setminus \{i\}} \alpha_\mathcal{S} \cdot (w_i x_i) = 1 \cdot (w_i x_i)$, which again gives the same result. Integrated gradients and LRP-0 also yield the same result. Hence, for this simple linear model, all explanation methods behave consistently and in agreement with the existing theoretical frameworks. The connection of explanation methods to Shapley values for linear models was also made in \cite{pmlr-v97-ancona19a}.

\medskip

The connection between Integrated Gradients and Taylor decomposition holds for a broader class of neural network functions, specifically deep rectifier networks (without biases):

\begin{proposition} When applied to deep rectifier networks of the type $f(\x) = \rho(W_L\, \rho( \dots \rho ( W_2\, \rho ( W_1\, \x))))$, Integrated Gradients with integration path $\{t\x ; 0 < t \leq 1\}$ is equivalent to Taylor decomposition at $\widetilde{\x} = \varepsilon \x$ in the limit $\varepsilon \to 0$.
\label{proposition:intgrad}
\end{proposition}

This can be shown by making the preliminary observation that a deep rectifier network is linear with constant gradient on the segment $(\boldsymbol{0},\x]$ and then applying the chain of equations $\int_{\varepsilon}^1 x_i [\nabla f(t\x)]_i dt = (1-\varepsilon) x_i [\nabla f(\varepsilon\x)]_i = \displaystyle [\nabla f(\varepsilon \x)]_i (x_i - \varepsilon x_i)$. This connection, along with the observation that a single gradient evaluation of a deep network can be noisy (cf.\ Section~\ref{section:difficult}) speaks against integrating on the segment $(\boldsymbol{0},\x]$. For this reason, we have opted in the experiments of Section \ref{section:comparison} to use a smoothed version of IG. A further result shows an equivalence between a `naive' version of LRP (using LRP-0 at every layer) and Taylor decomposition.

\begin{proposition} For deep rectifier nets of the type $f(\x) = \rho(W_L\, \rho( \dots \rho ( W_2\, \rho ( W_1\, \x))))$, applying LRP-0 at each layer is equivalent to a Taylor decomposition at $\widetilde{\x} = \varepsilon \x$ in the limit $\varepsilon \to 0$.
\label{proposition:lrp0}
\end{proposition}

This result can be derived by taking the LRP formulation of Eq.\ \eqref{eq:modgradient} and setting $\gamma=0$. This equation then reduces to:
$$
\textstyle c_j = \sum_k w_{jk} \mathrm{step}\big(\sum_{0,j} a_j w_{jk}\big) c_k
$$
where $\mathrm{step}(t) = 1_{t>0}$. This equation is exactly the same as the one that propagates gradients in a deep rectifier network. Hence, the input relevance computed by LRP becomes \mbox{$R_i = x_i c_i = x_i [\nabla f(\x)]_i$} for which we have already shown the equivalence to simple Taylor decomposition in the proposition above. The connection has been originally made in \cite{shrikumar2016not}.

\begin{proposition} For deep rectifier networks of the type $f(\x) = \rho(W_L\, \rho( \dots \rho ( W_2\, \rho ( W_1\, \x))))$, applying LRP-$\gamma$ is equivalent to performing one step of deep Taylor decomposition and choosing the nearest root point on the line $\{\ba - t \ba \odot (\boldsymbol{1} + \gamma \cdot \boldsymbol{1}_{\w_k \succeq \boldsymbol{0}}); t \in \mathbb{R} \}$.
\label{proposition:lrpgen}
\end{proposition}

We choose the relevance model $\widehat{R}_k(\ba) = a_k(\ba) \cdot c_k$ with $c_k$ constant (cf.\ \cite{DBLP:series/lncs/MontavonBLSM19} for a justification).
Injecting the root point in the first-order terms of DTD (summands of Eq.\ \eqref{eq:dtd}) gives:
\begin{align*}
R_{j \leftarrow k} &=\textstyle w_{jk} \cdot c_k \cdot (a_j - (a_j - t a_j \cdot (1 + \gamma \cdot 1_{w_{jk} \geq 0})))\\
&= \textstyle a_j \cdot (w_{jk} + \gamma w_{jk}^+) \cdot t \cdot c_k
\end{align*}
where $t$ is resolved using the conservation equation $\sum_j R_{j \leftarrow k} = R_k$. LRP-$0$ is a special case of LRP-$\gamma$ with $\gamma=0$. A similar procedure with another choice of reference point gives LRP-$\epsilon$~(cf.\ \cite{DBLP:series/lncs/MontavonBLSM19}).

%%%%%%%%%%%%%%%%%%%%%%%%%%%%%%%%%%%%%%%%%%%%%%%%%%%%%%%%%%%%%%%%%%%%%%%%%%%%%%%%%%%%
\section{Extending Explanations}
%\section{Explanations for Unsupervised Learning and Beyond}
\label{section:beyond}
    
The explanation methods we have presented in the previous sections were applied to a particular class of models (deep neural networks) and produced particular types of explanations (attribution on the input features). We present in the following various extensions that broaden the applicability of these methods and diversify the type of explanation that can be produced. In particular, we will discuss (1) higher-order methods to produce richer explanations involving combination of features, (2) a systematic way of extending explanation methods to non-neural network models e.g.\ in unsupervised learning where explanations are also needed, (3) a principled way to ensure that explanations of DNN classifiers are class-discriminative, and (4) strategies to go beyond individual explanations to arrive at a general understanding of the ML model.

\subsection{Explaining Beyond Heatmaps}
\label{section:higherorder}

The locally linear structure of deep neural networks lends itself well to the heatmap-based methods we have reviewed in this paper, which we call first-order methods. However, special types of neural networks, e.g.\ that incorporate products between input or latent variables lose that property. Neural networks with product structures commonly occur for relational tasks such as comparing images \cite{DBLP:journals/pami/Memisevic13} or collaborative filtering \cite{DBLP:conf/www/HeLZNHC17}. Graph neural networks \cite{DBLP:journals/tnn/ScarselliGTHM09,DBLP:conf/iclr/KipfW17} multiply the input connectivity matrix multiple times, and consequently also exhibit product structures. In that case, the neural network is no longer piecewise linear and typically becomes piecewise polynomial with its input.

\smallskip

\begin{figure}[ht]
    \centering
    \includegraphics[width=.8\linewidth,clip=True,trim=20 80 20 80]{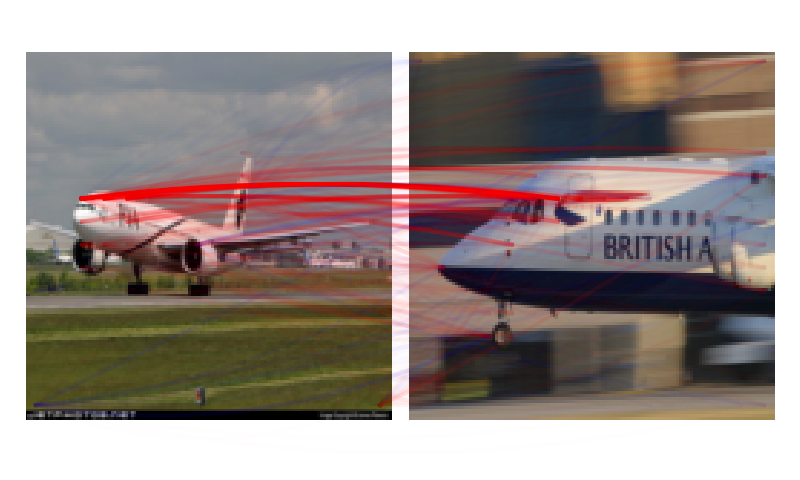} %\includegraphics[width=.48\linewidth,clip=True,trim=30 20 15 25]{figures/gnnlrp.pdf}
    \caption{Example of two images predicted to be similar, along with a BiLRP second-order attribution of their similarity score rendered as a bipartite graph. (Figure from \cite{Eberle2020Similarity}). The explanation shows that the front part of the two planes jointly contributes to the predicted similarity.}
    \label{fig:bilrp}
\end{figure}

For illustration, we present the BiLRP method \cite{Eberle2020Similarity}, which assumes we have a similarity model built as a dot product on some hidden representation $\phi = \phi_L \circ \dots \circ \phi_1$ of a deep network:
$$
y(\x,\x') = \langle \phi_L \circ \dots \circ \phi_1(\x),\phi_L \circ \dots \circ \phi_1(\x')\rangle
$$
If all functions $\phi_1,\dots,\phi_L$ are piecewise homogeneous linear, then $y$ can be rewritten as a composition:
$$
y(\x,\x') = \psi_L \circ \dots \circ \psi_1(\x,\x')
$$
with $\psi_1,\dots,\psi_L$ piecewise bilinear. Using deep Taylor decomposition, but at each step applying a \textit{second-order} Taylor expansion, we arrive conceptually at an attribution of the similarity score $y(\x,\x')$ to \textit{pairs} of input features. (Practically, the attribution can be expressed as a product of two branches of LRP computation, hence the name BiLRP.) An example of BiLRP explanation for the similarity of two planes in VGG-16 feature space is shown in Fig.\ \ref{fig:bilrp}.

We observe that the explanation does not highlight individual features but instead pairs of features from the two input images, reflecting the fact that they are jointly relevant to explain the high similarity score.

Other higher-order methods include GNN-LRP which explains graph neural networks in terms of collections of edges \cite{Schnake2020Graphs}, or Integrated Hessians \cite{Janizek2020IntegratedHessians}, which can be seen as a second-order extension of Integrated Gradients.

\smallskip

Finally, we note that the Shapley framework also offers the possibility to quantify the joint contribution of two interacting players through the Shapley interaction index \cite{grabisch1999axiomatic}. Lundberg et al. \cite{lundberg2020local} recently applied this concept to compute
higher-order explanations of a machine learning model. Here the interaction value between feature $i$ and feature $j$ can be computed as
$$
\phi_{ij} =\!\!\!\!\sum_{\mathcal{S} \subseteq \mathcal{P} \setminus \{i,j\}}\!\!\!\!\alpha_\mathcal{S} \cdot (v(\mathcal{S} \cup \{i,j\}) - v(\mathcal{S} \cup \{i\}) - v(\mathcal{S} \cup \{j\}) + v(\mathcal{S}))
$$
with the weight factor $\alpha_\mathcal{S} =  |\mathcal{S}|! \cdot (|\mathcal{P}| - 2 - |\mathcal{S}|)! / 2|\mathcal{P} - 1|!$. Shapley interaction values allow for a separate consideration of interaction effects, which can be crucial for understanding the model prediction strategy (see Section \ref{sec:shapexample} for a worked-through example in a medical application). However, although theoretically valid, these Shapley interaction terms are in most cases infeasible to compute for complex machine learning models such as deep neural networks, and approximations are therefore required \cite{matsui2001np}.

%%%%%%%%%%%
\subsection{Explaining Beyond Deep Networks}

Deep neural networks have been shown to perform extremely well on classification or regression tasks. However for unsupervised problems such as anomaly detection or clustering, although deep models have also reached successes \cite{DBLP:conf/eccv/CaronBJD18,ruff2018deep,ruff2020unifying}, shallow models (e.g.\ centroid-based \cite{macqueen1967}, PCA-based \cite{hawkins1974}, kernel-based \cite{parzen1962}, or combinations of them \cite{scholkopf1998nonlinear,DBLP:conf/kdd/DhillonGK04,hoffmann2007}) remain popular workhorses. As these models are not given in the form of a neural network,
%, and still composed of strongly nonlinear functions such as the exponential,
a direct application of methods designed in the context of linear models and DNNs is not feasible.

While a possible approach to explaining would be to train a surrogate neural network to match the output of the unsupervised model, it was found that some unsupervised models such as k-means clustering \cite{macqueen1967} or kernel density estimation \cite{parzen1962} can be directly rewritten as a neural network (or `neuralized'), without requiring any retraining or architecture search~\cite{Kauffmann19,Kauffmann20}.
%These functional `copies' are furthermore only composed of `canonical' neural network functions, e.g.\ linear or pooling. This general concept of neuralization was first introduced in the context of explanation methods for unsupervised learning, namely, one-class SVMs \cite{Kauffmann20} and k-means clustering models \cite{Kauffmann19}, where combinations of kernel RBF functions can be rewritten as pooling operations over linear or distance functions.

\begin{figure}[ht]
\centering
\includegraphics[width=1.0\linewidth]{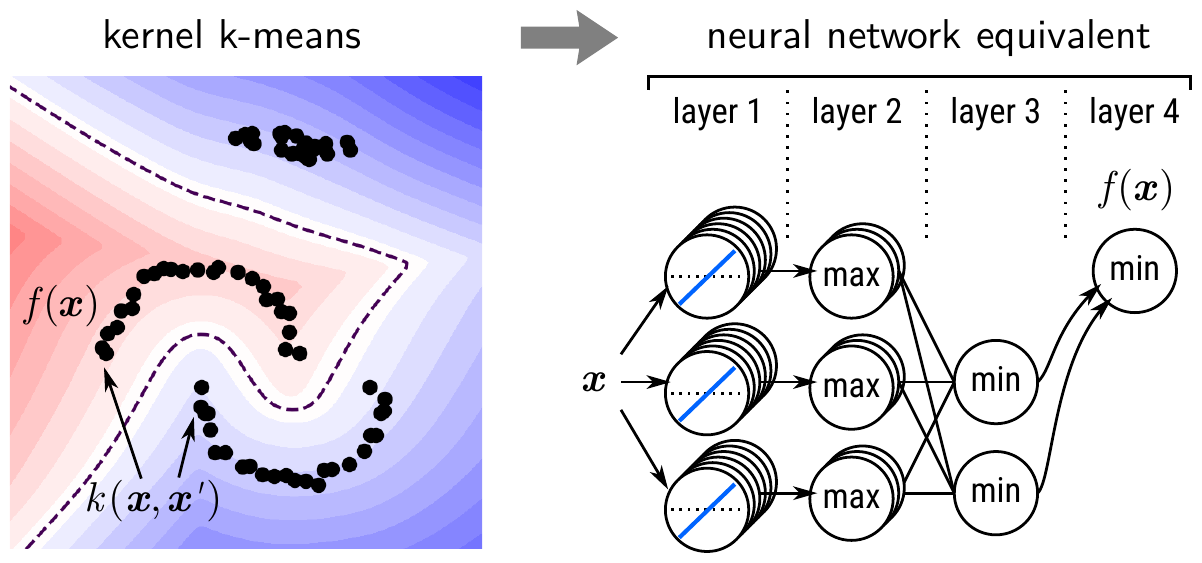}
\caption{{\em Left:} Kernel k-means applied to a toy two-dimensionnal problem with three clusters. Red and blue color in the background represent the positive and negative values of the logit function for a given cluster. {\em Right:} 4-layer neural network equivalent of the kernel k-means logit score \cite{Kauffmann19}.}
\label{fig:kernelkmeans}
\end{figure}

Consider for illustration a kernel k-means model of the type studied in \cite{DBLP:conf/kdd/DhillonGK04}. For this type of model, and assuming a Gaussian kernel $\mathbb{K}(\x,\x') = \exp(-\gamma \|\x-\x' \|^2)$, the probability ratio in favor a given cluster $\omega_c$ can be expressed as:
\begin{align}
\frac{P(\omega_c|\x)}{1 - P(\omega_c|\x)} = \frac{\big(Z_c^{-1} \sum_{i \in \mathcal{C}_c} \mathbb{K}(\x,\x_i)\big)^{\beta/\gamma}}{\sum_{k \neq c} \big( Z_k^{-1}\sum_{j \in \mathcal{C}_k} \mathbb{K}(\x,\x_j)\big)^{\beta/\gamma}}
\label{eq:kernelkmeans}
\end{align}
This is a power-assignment model applied to the kernel density functions of each cluster. The sets $\mathcal{C}_c$ and $\mathcal{C}_k$ are the representatives for clusters $c$ and $k$, and $Z_c, Z_k$ are respective normalization factors. An example of decision function produced by this model for a three-cluster problem is shown in Fig.\ \ref{fig:kernelkmeans} (left). Clearly, Eq.\ \eqref{eq:kernelkmeans} is a priori not composed of neurons. However, it can be reorganized into the following sequence of detection and pooling functions \cite{Kauffmann19}:
\begin{align*}
\log \Big[ \frac{P(\omega_c|\x)}{1 - P(\omega_c|\x)} \Big]
=
\beta \, {\min_{k \neq c}}^\beta \big\{
{\min_{j \in \mathcal{C}_k}}^\gamma \big\{
{\max_{i \in \mathcal{C}_c}}^{\gamma} \big\{
\w_{ij}^\top \x + b_{ijk}
\big\} 
\big\} 
\big\}
\end{align*}
with $\w_{ij} = 2 (\x_i - \x_j)$ and $b_{ijk} = \|\x_j\|^2 - \|\x_i\|^2 + \gamma^{-1} (\log Z_k - \log Z_c)$ are parameters of the first linear layer. This layer is followed by a hierarchy of log-sum-exp computations interpretable as canonical max- and min-pooling operations. The neuralized version of kernel k-means is depicted in Fig.\ \ref{fig:kernelkmeans} (right).

%With the resulting equivalent neural network architecture, we have a more structured representation of the clustering solution. The first layer consists of directional filters between individual data points, the second layer extract representatives for each cluster, the third layer find the nearest neighbors from other clusters, and the top layer finds the nearest cluster among competing clusters.
With this structure, explanation techniques such as LRP can be used to produce explanations of cluster membership \cite{Kauffmann19}. In particular, the LRP backward pass first identifies the most relevant class competitors, then the most relevant representatives (i.e.\ data points) of these class competitors and of the cluster of interest, and finally the most relevant directions in input space, thereby producing a heatmap-based explanation of the cluster assignment \cite{Kauffmann19}.

\subsection{Explaining Beyond Output Neurons}

\begin{figure}[ht]
\centering
\includegraphics[width=1.0\linewidth]{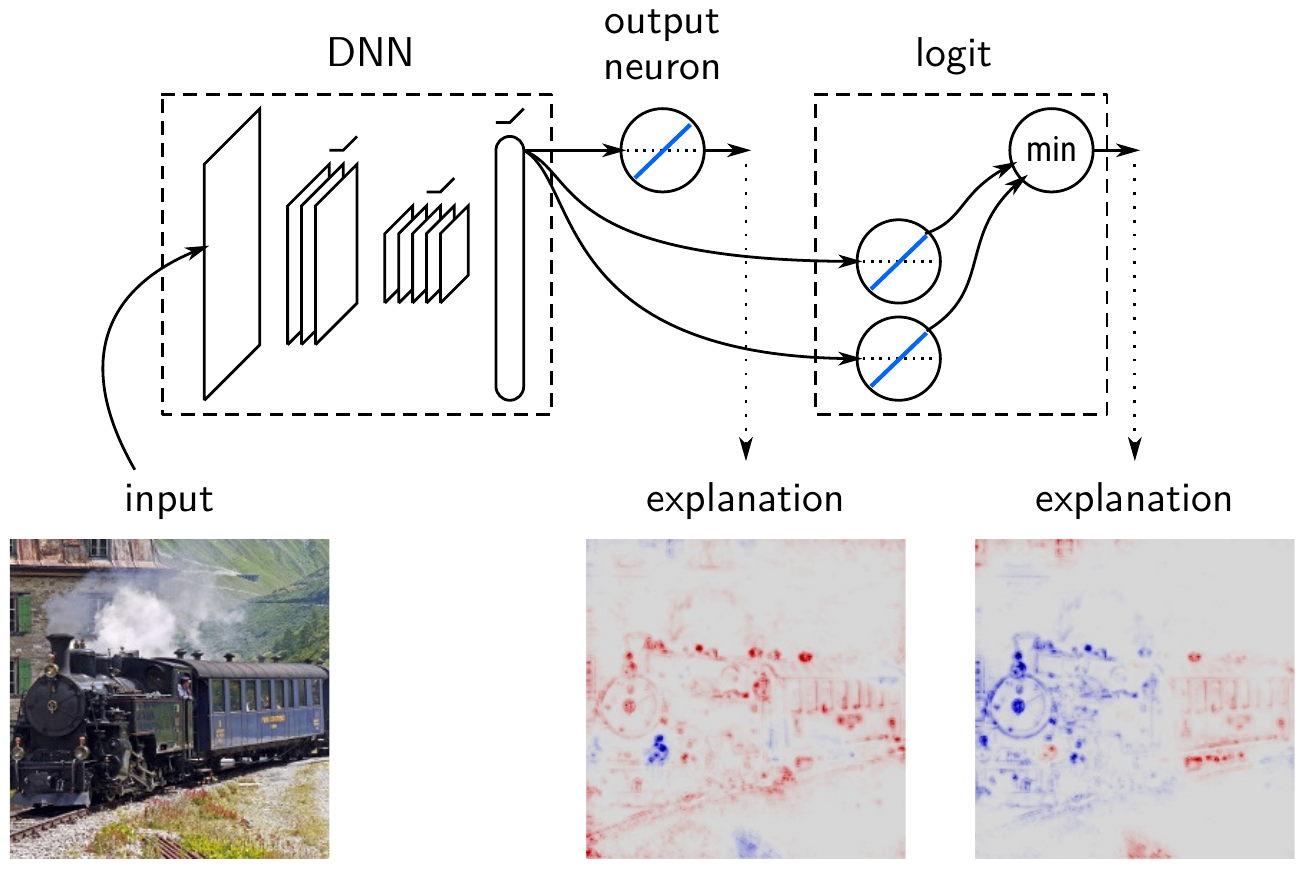}
\caption{Deep neural network to which we append a neuralized version of the log-likelihood ratio \cite{DBLP:series/lncs/MontavonBLSM19}. Considering the latter quantity instead of the DNN output leads to a different explanation.}
\label{fig:softmax}
\end{figure}

The concept of neuralization can also be applied in a supervised learning setting, for improving the explanation of a deep neural network classifier. So far, we have explained quantities at the output of the last linear layer of the network. Because these output quantities are unnormalized they may respond positively to several classes, thereby lacking selectivity. The problem of class selectivity was highlighted e.g.\ in \cite{DBLP:conf/accv/GuYT18, DBLP:conf/iccvw/IwanaKU19, DBLP:series/lncs/MontavonBLSM19} along with practical solutions to overcome this effect. Here, we present the `neuralization' approach of \cite{DBLP:series/lncs/MontavonBLSM19}, which first makes the observation that ratios of probabilities as given by the top-layer soft-assignment model can be expressed as:
\begin{align*}
\frac{P(\omega_c|\x)}{1 - P(\omega_c|\x)} &= \frac{\exp(\w_c^\top \ba)}{\sum_{k \neq c} \exp(\w_k^\top \ba)}
\end{align*}
This computation can then be reorganized in the two-layer neural network
\begin{align*}
\log\Big[\frac{P(\omega_c|\x)}{1 - P(\omega_c|\x)}\Big] = {\min_{k \neq c}}\big\{
(\w_{c} - \w_{k})^\top \ba
\big\}
\end{align*}
where $\min$ is a soft minimum implemented by a log-sum-exp computation. The DNN processing up to the output neuron or up to the output of the neuralized logit model is illustrated in Fig.\ \ref{fig:softmax} along with LRP explanations for these two quantities associated to the class `passenger\_car'.

In the first explanation, both the passenger car and the locomotive can be seen to contribute. In the second explanation, the locomotive turns blue. The latter is indeed speaking for the class locomotive, which mechanistically lowers the probability for the class `passenger\_car' \cite{DBLP:series/lncs/MontavonBLSM19}. This example shows that it is important in presence of correlated features to precisely define what quantity (unnormalized score or logit) we would like to explain.

We note that while neuralization has served here to support LRP-type explanations, the concept could potentially be used for other purposes. The identified neural network structure may help to gain further understanding of the model or provide intermediate representations that are potentially useful to solve related tasks.

%%%%%%%%%%%
\subsection{Explaining Beyond Individual Predictions}
\label{section:datasetwide}

In practice, we may not only be interested in explaining how the DNN predicts a single data point, but also in the statistics of them for a whole dataset. This may be useful to validate the model in a more complete manner. Let $f:\mathbb{R}^d \to \mathbb{R}$ be a function that takes a data point as input and predicts evidence for a certain class for each data point. Consider a dataset $\x_1,\dots,\x_N$ of such data points. The total class evidence can be represented as a function $g\colon\mathbb{R}^{N \times d} \to \mathbb{R}$
where:
$$
g(\x_1,\dots,\x_N) = \textstyle \sum_{n=1}^N f(\x_n)
$$
This composition of the neural network output and a sum-pooling remains explainable by all methods surveyed here, however, the explanation is now high-dimensional ($N \times d$).

\subsubsection{Relevance Pooling}
\label{section:pooling}

\begin{figure}[ht]
    \centering
    \hfill
    \includegraphics[width=0.84\linewidth]{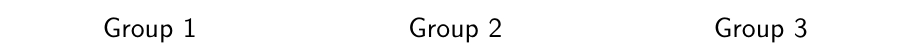}\\[2mm]
    \hfill
    \includegraphics[width=0.28\linewidth]{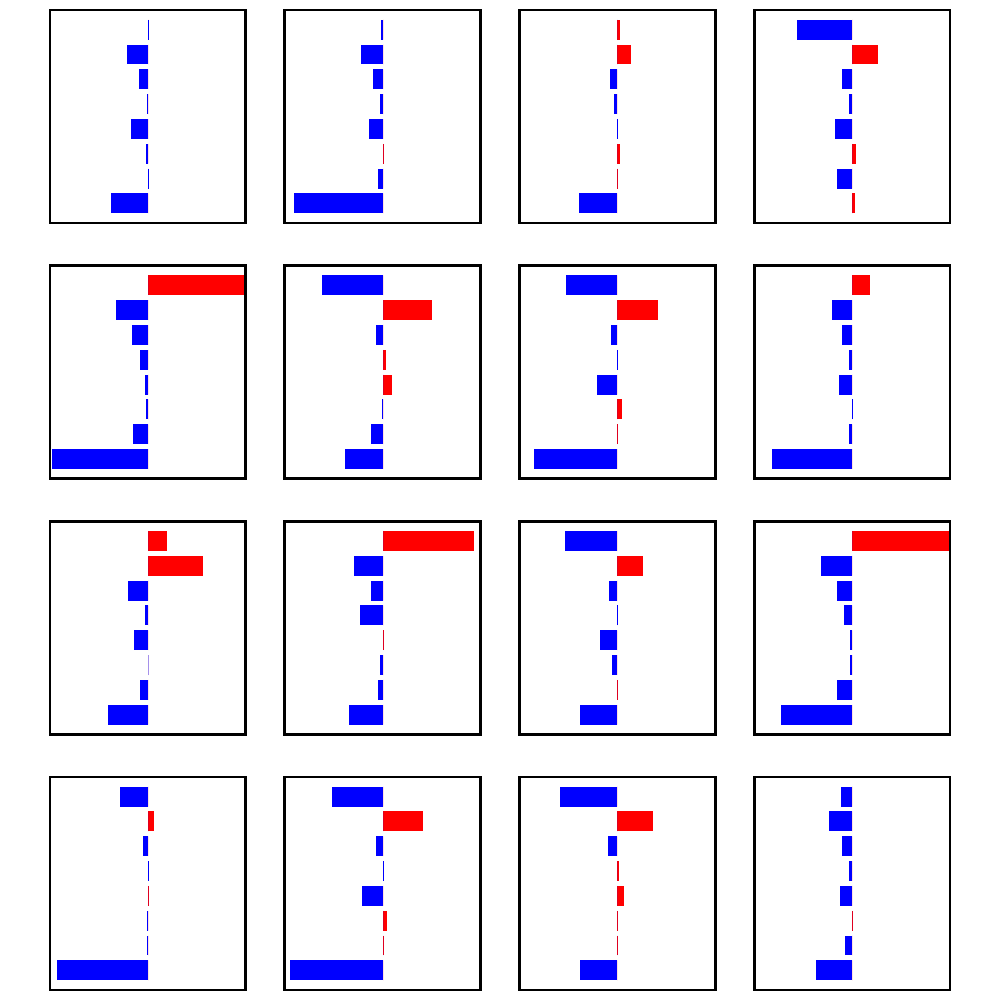}%
    \includegraphics[width=0.28\linewidth]{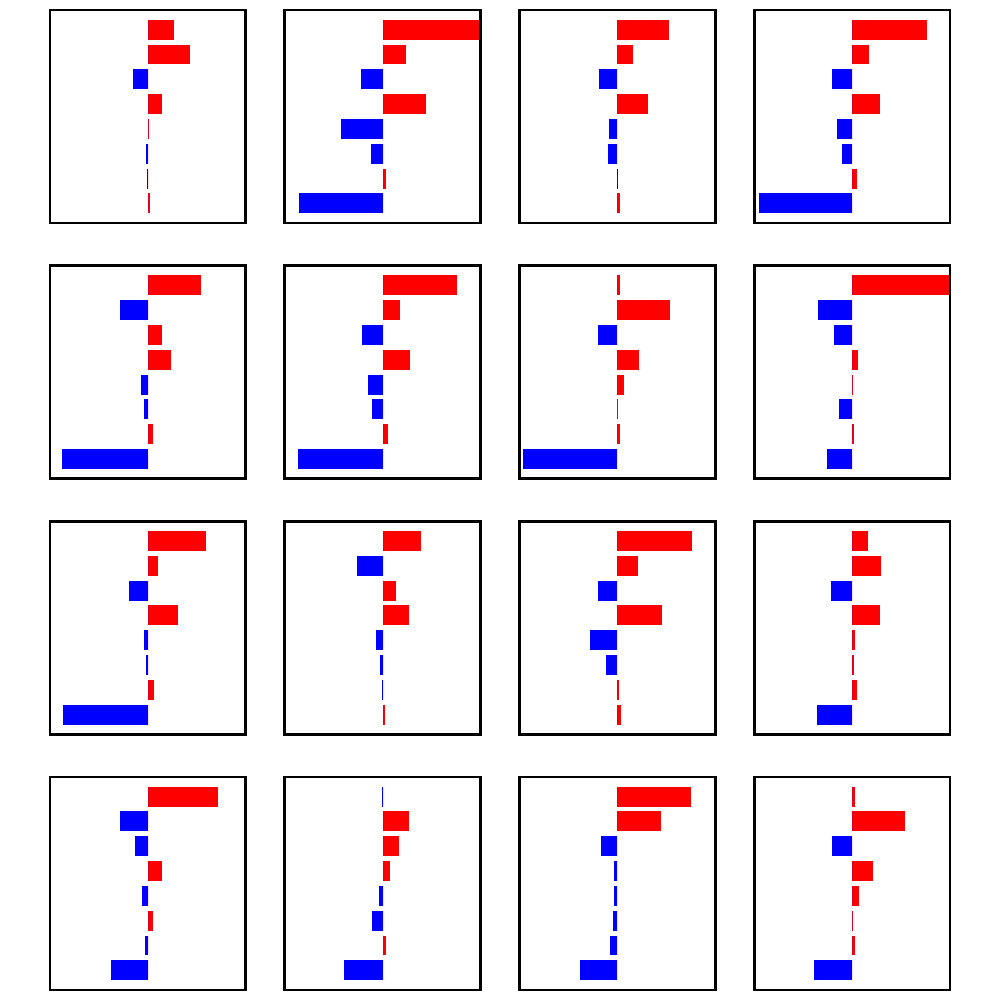}%
    \includegraphics[width=0.28\linewidth]{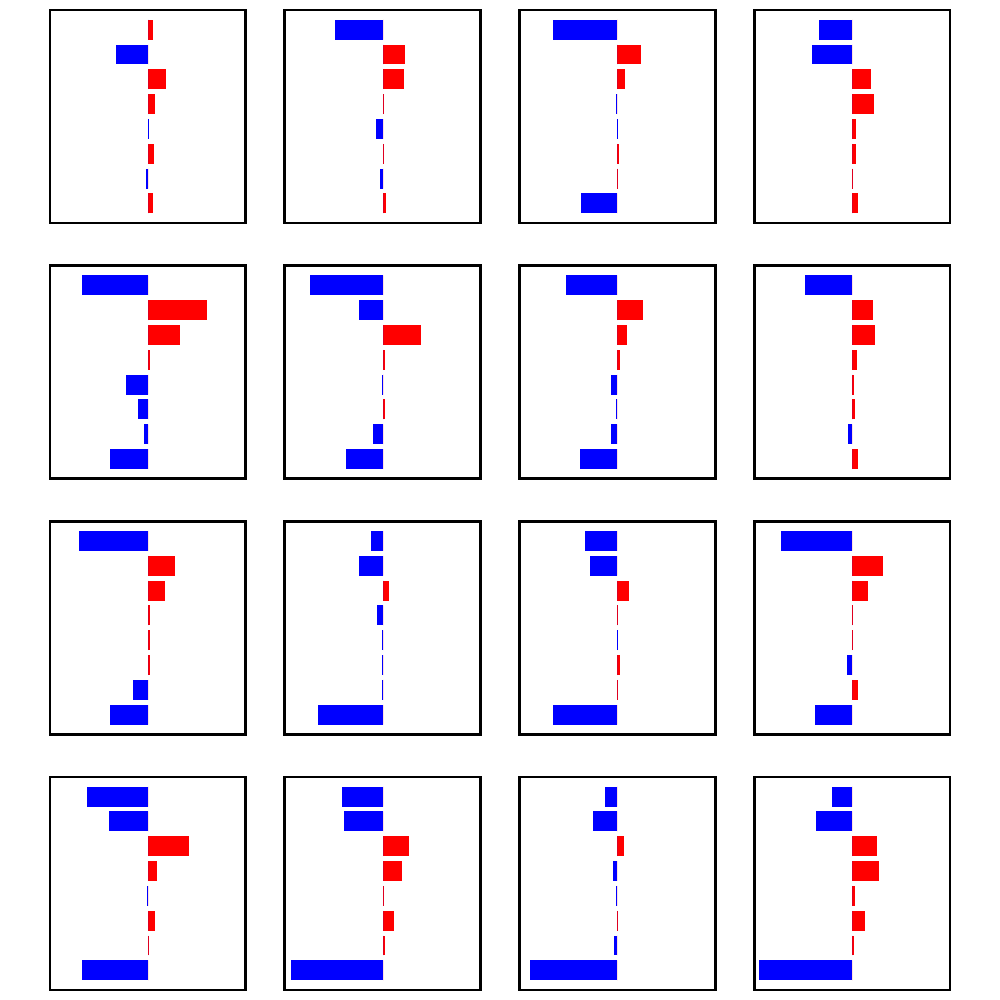}
    \\[2mm]
    \hfill
    \includegraphics[width=0.84\linewidth]{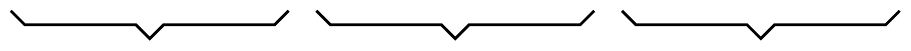}\\[2mm]
    \hfill
    \includegraphics[width=0.42\linewidth]{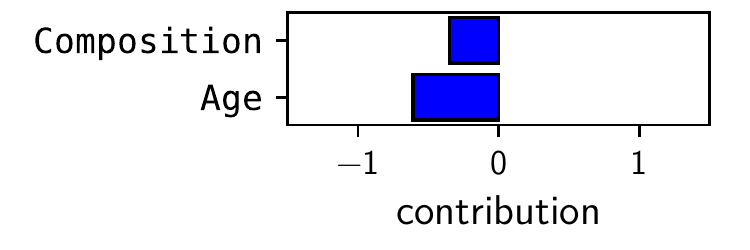}%
    \includegraphics[width=0.28\linewidth]{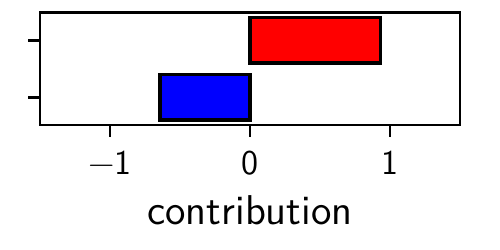}%
    \includegraphics[width=0.28\linewidth]{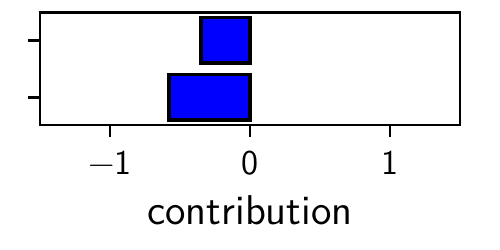}
    \caption{Pooled analysis. {\em Top:} Feature-wise contributions for the prediction on three clusters of the Concrete Compressive Strength Dataset \cite{Yeh1998}. {\em Bottom:} Coarse-grained explanations obtained by pooling contributions on data clusters and groups of features.}
    \label{fig:datasetwide}
\end{figure}

Practically, we may be not be interested in explaining every single data point in terms of every single input features. A more relevant information to the user would be the overall contribution of a subgroup of features $\mathcal{I}$ on a group of data points $\mathcal{G}$~(cf.\ \cite{LapCVPR16,montavon2018methods}). In particular the Integrated Gradient and LRP methods surveyed here produce explanations that satisfy the conservation property:
\begin{align*}
g(\x_1,\dots,\x_N) &\approx \sum_{n=1}^N \sum_{i=1}^d R_{i,n}
\intertext{and that can be converted to a coarse-grained explanation}
&\approx \sum_{\mathcal{G}} \sum_{\mathcal{I}} \underbrace{\sum_{n \in \mathcal{G}} \sum_{i \in \mathcal{I}} R_{i,n}}_{R_{\mathcal{I},\mathcal{G}}}
\end{align*}
that still satisfies the desired conservation property. As an illustration of the concept, we consider the `Concrete Compression Strength' example of Section \ref{section:towards}. Data points are grouped in three k-means clusters, and features are grouped in two sets: the singleton $\{\text{age}\}$, and the set of all remaining features describing concrete composition. The pooled analysis is illustrated in Fig.\ \ref{fig:datasetwide}.

This analysis gives further insight into our predictive model. We observe that most distinguishing factors, especially age, contribute negatively to strength. In other words, a `typical' age and composition is a recipe for strength whereas high/low values tend to be explanatory for weakness. Notably, one data cluster stands out by having composition features that are explanatory for strength.

\subsubsection{Spectral Relevance Analysis (SpRAy) \cite{lapuschkin2019unmasking}}
\label{section:spray}

While in Section \ref{section:pooling} we have reduced the dimensionality through pooling, other analyses are possible. For example, the SpRAy method \cite{lapuschkin2019unmasking} does not assume a fixed pooling structure (e.g.\ a partition of data points and a partition of features), and applies instead a clustering of explanations in order to identify protypical decision strategies. Algorithm \ref{algorithm:spray} outlines the three steps procedure used by SpRAy:

\begin{algorithm}
\caption{Spectral Relevance Analysis}
\label{algorithm:spray}
\begin{algorithmic}
\FOR{$n=1$ \textbf{to} $N$}
\STATE $\R^{(n)} \leftarrow \texttt{explain}(\x^{(n)},f)$
\STATE $\,\overline{\!\R}^{(n)} \leftarrow \texttt{normalize}(\R^{(n)})$
\ENDFOR
\STATE $\texttt{clustering}(\{\,\overline{\!\R}^{(1)},\dots,\,\overline{\!\R}^{(N)}\})$
\end{algorithmic}
\end{algorithm}

\begin{figure}
    \centering
    \includegraphics[width=.8\linewidth,clip=True,trim=0 15 0 15]{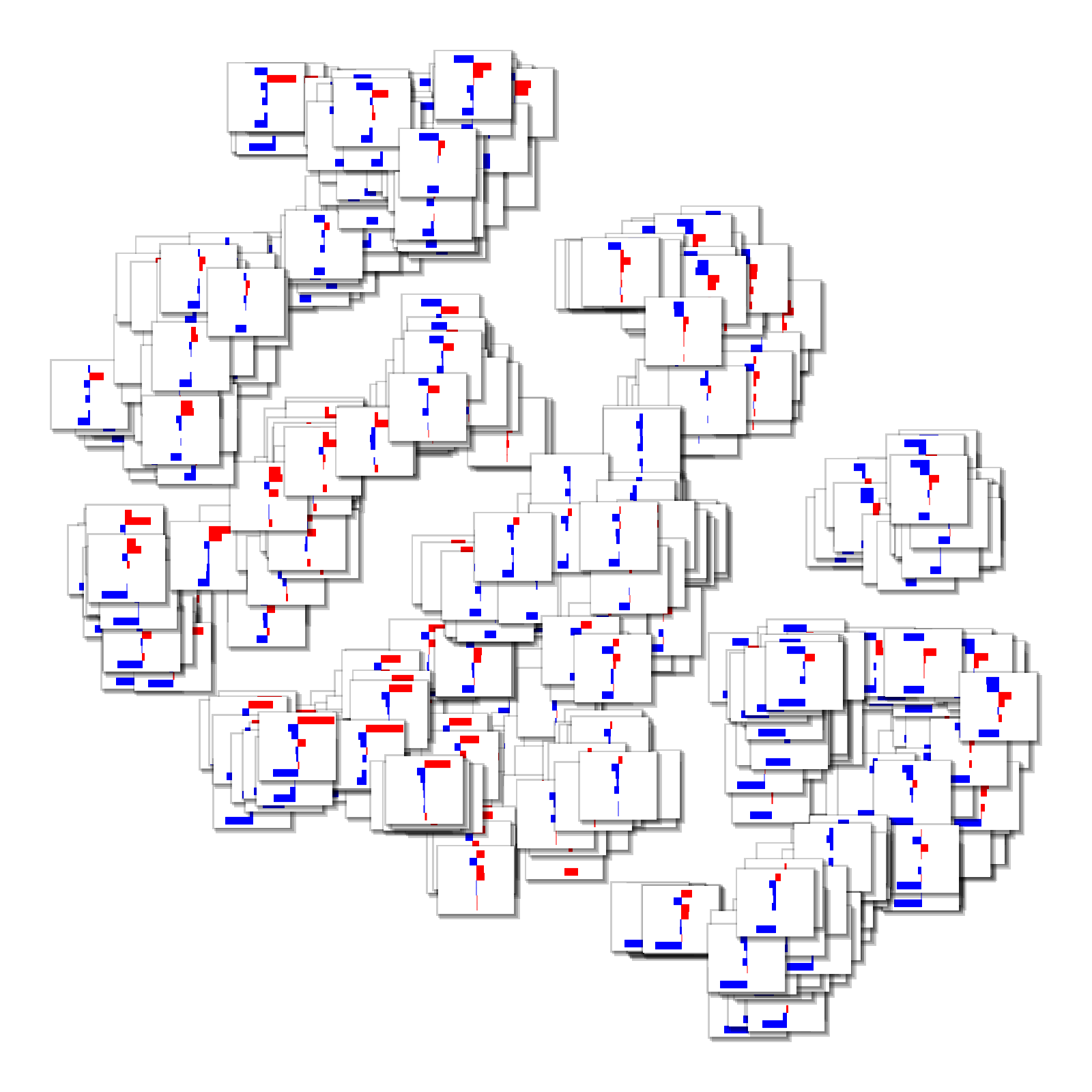}\vspace{-3mm}
    \caption{SpRAy analysis. Explanation of the predictions of the Concrete Compressive Strength Dataset \cite{Yeh1998}, displayed at coordinates corresponding to their t-SNE embedding. This analysis provides a visualization of the overall classifier's strategy.}
    \label{fig:spray-concrete}
\end{figure}

The method first produces an explanation for each data point. In principle, any explanation method can be used, e.g.\ occlusion, integrated gradients, or LRP. Explanations are then normalized (e.g.\ blurred and standardized) to become invariant to small pixel-wise or saliency variations. Finally, a clustering algorithm is applied to the normalized explanation, and examples with the same cluster index can be understood as being associated with some prototypical decision strategy, e.g.\ looking at the object, looking at the background, etc. Alternately, the clustering step can be replaced by a low-dimensional embedding step to produce a visual map of the overall decision structure of the ML model.
The SpRAy analysis is illustrated in Fig.\ \ref{fig:spray-concrete} on the same Concrete Compressive Strength Dataset \cite{Yeh1998} used before.

\medskip

Here, we observe, for example, that a typical decision strategy (bottom left) consists of explaining high compressive strength based on the first feature (high cement concentration), whereas another typical decision strategy (bottom right) consists of predicting low compressive strength based on the last feature (low concrete's age). This rich visual feedback can serve to validate the prediction strategy with expert knowledge to make sure that the model does not predict based on a dataset artefact. We will present in Section \ref{sec:systematic} a successful real-world use-case of the SpRAy analysis for inspecting a state-of-the-art model trained on a large image classification dataset.

Altogether, relevance pooling and SpRAy support a variety of dataset-wide analyses that are useful to explore and characterize the decision strategies of complex models trained on large datasets. 

%%%%%%%%%%%%%%%%%%%%%%%%%%%%%%%%%%%%%%%%%%%%%%%%%%%%%%%%%%%%%%%%%%%%%%%%%%%%%%%%%%%%
\section{Worked-Through Examples}

In this paper, we have motivated the use of explanation in the context of deep learning models and showcased some methods for obtaining explanations. Here, we aim to take a practical look for the user to assess when explanation is required, what are common issues with applying explanation techniques / setting their hyperparameters, and finally, how to make sure that the produced explanations deliver meaningful insights for the human.

%%%%%%%%%%%
\subsection{Example 1: Validating a Face Classifier}

In the first worked-through example we wish to train an accurate classifier for predicting a person's age from images of faces. We will show how to use explanation for this task, in particular, to verify that the model is not using ``wrong'' features for its decisions.

Let us use for this the Adience benchmark dataset~\cite{eidinger2014age} providing 26,580 images captured `in the wild' and labelled into eight ordinal groups of age ranges \{(0-2), (4-6), (8-13), (15-20), (25-32), (38-43), (48-53), (60+)\}.

Because the number of examples in this dataset is limited and likely not sufficient to extract good visual features, we adopt the common approach of starting with a generic pretrained classifier and fine-tune it on our task. We download a \mbox{VGG-16}~\cite{DBLP:journals/corr/SimonyanZ14a} neural network architecture pretrained on ImageNet~\cite{deng2009imagenet} obtainable from  \texttt{modelzoo.co}. First test results after training using Stochastic Gradient Descend (SGD)~\cite{lecun2012efficient} report reasonable performance, with \emph{exact} and \emph{1-off}~\cite{shan2010learning,eidinger2014age} prediction accuracy\footnote{Results have been averaged over the official pre-selected five-fold dataset split~\cite{eidinger2014age}.} of 56.5\% and 90.0\%, respectively. Here, the 1-off accuracy considers predictions of (up to) one age group away from the true label as correct.

\begin{figure}[ht]
    \centering
    \includegraphics[width=1.0\linewidth]{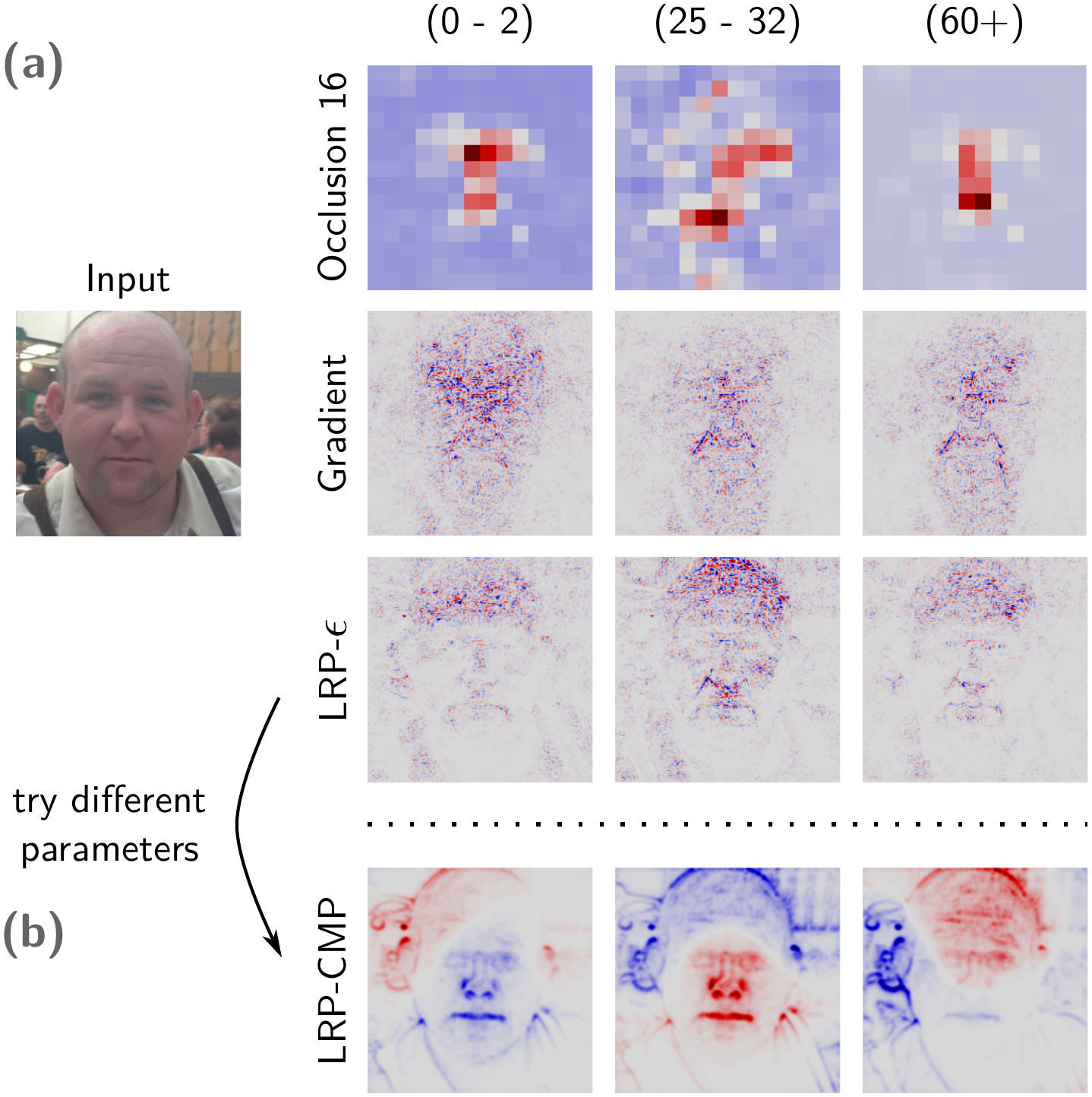}
    \caption{\emph{Top:} Explanations produced by different methods on the VGG-16 age model. Results are shown for the output neurons associated to age group labels \mbox{(0-2)}, \mbox{(25-32)} and \mbox{(60+)} respectively.
    \emph{Bottom:} Application of the layer-dependent LRP-CMP decomposition strategy.}
    \label{fig:walkthrough-faces-lrp}
\end{figure}

In order to understand the learned prediction strategies of our model and to verify that it uses meaningful features in the training data, we test different explanation methods, first, a simple occlusion and gradient analysis, and we also take an off-the-shelve explanation software,
the LRP Toolbox~\cite{LapJMLR16} for Caffe~\cite{jia2014caffe}, and choose the method LRP configured to perform `LRP-$\epsilon$' on all layers in a first attempt. Explanations are shown for a given image in Fig.~\ref{fig:walkthrough-faces-lrp} (a).

Some insight can be readily obtained from these explanations, e.g.\ the classifier has learned to attend the center of the image where the face is.
However, we also observe several limitations of the explanations: Here, the occlusion-based analysis produces a blue background (negatively relevant) instead of identifying the background as irrelevant. The next two explanations (gradient and LRP-$\epsilon$) are here overly complex with frequent local sign changes, making it hard to extract further insights, especially what are the features that contribute to different age groups. The reason or such poor performance is that we have either not considered a broad enough spectrum of explanation methods, or the parameters of these methods have not been set optimally.  This leads to our first recommendation:

\begin{center}
\fbox{\bf Try different parameters of the explanation techniques}
\end{center}

Specifically, we will now try an alternate LRP preset called `LRP-CMP' that applies a composite strategy \cite{LapICCVW17,DBLP:series/lncs/MontavonBLSM19,kohlbrenner2020towards} where different rules are applied at different layers. Explanations obtained with this new rule are given in Fig.~\ref{fig:walkthrough-faces-lrp}~\emph{(bottom)}. The new explanations highlight features in a much more interpretable way and we also start to better understand what speaks --- according to the model --- in favor of or against certain age groups.
For example, explanations amusingly reveal baldness as a feature corresponding to both age groups (0-2) and (60+).
In the shown sample, baldness contributes evidence for the classes (0-2) and (60+), while it speaks against the age group (25-32).
Relatedly, the expression of the man's chin and mouth area contradicts 
class (0-2) more than class (60+), but `looks like' it would belong to a person aged (25-32).

\medskip

\begin{figure}[ht]
    \centering
    \includegraphics[width=1.0\linewidth]{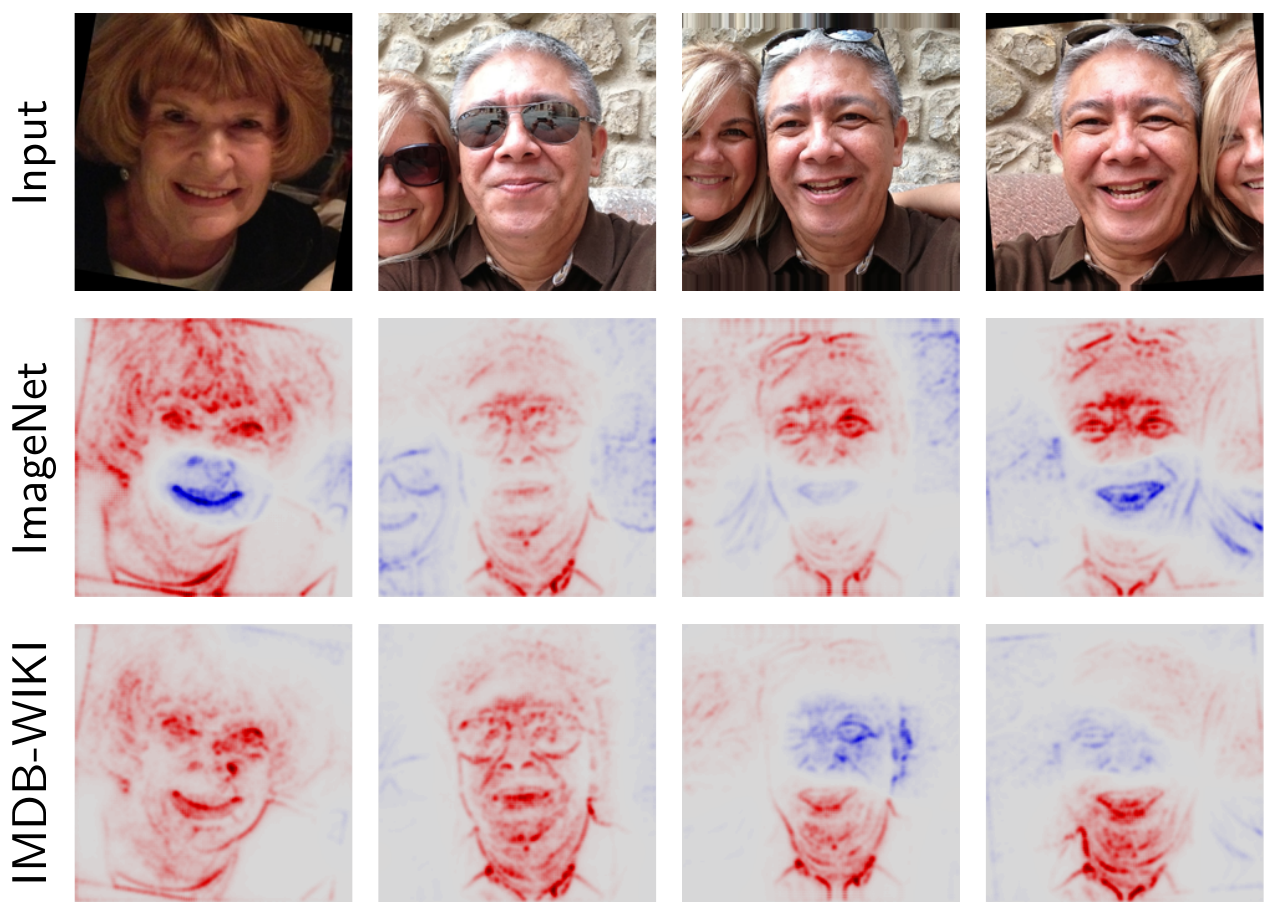}
    \caption{LRP heatmaps demonstrating the effects of ImageNet~\cite{deng2009imagenet} pretraining (\emph{middle}) compared to additional IMDB-WIKI~\cite{rothe2015dex} pretraining (\emph{bottom}).
    All heatmaps show the model decision w.r.t.\ age group (60+).
    }
    \label{fig:walkthrough-faces-cleverhans}
\end{figure}

Let's now move back to the initial question, namely how to verify that the model is using the right features for predicting. While the decision structure of the model was meaningful in Fig.\ \ref{fig:walkthrough-faces-lrp}, we would like to verify it is also the case for other test cases.
Fig.~\ref{fig:walkthrough-faces-cleverhans}~\emph{(top)} shows further examples from the Adience dataset; a woman labelled (60+) and three images of the same male labelled (48-53) with smiles of varying intensities.

We apply LRP with the same preset `LRP-CMP' on these images. LRP evidence for each image for the class (60+) is shown in Fig.\ \ref{fig:walkthrough-faces-cleverhans} (middle). Surprisingly, according to the model, broad smiling contradicts the prediction of belonging to the age group (60+). Smiling is however clearly a confounding factor, which reliably predicts age group only to the extent that no such case is present in the training data. This predicting strategy is related to the `Clever Hans'\footnote{`Clever Hans' was a famous horse at the beginning of the 20th century, which was believed by his trainer to be capable of performing arithmetic calculations. Subsequent analyses revealed that the horse was not performing arithmetic calculations but was detecting cues on the face of his trainer to produce the right answers. In machine learning, the term `Clever Hans' can be used to designate strategies that mimic the expected behavior but are based on unexpected correlations or artefacts in the data \cite{lapuschkin2019unmasking}.} effect \cite{lapuschkin2019unmasking} and we can therefore formulate our second recommendation:
\begin{center}
\fbox{\bf Unmask `Clever Hans' examples}
\end{center}
Note that instead of screening through all images manually, we can also use techniques such as SpRAy \cite{lapuschkin2019unmasking}, which perform such analysis semi-automatically for large datasets such as ImageNet (see also Section \ref{sec:systematic} for successful applications).

While for the examples showcased in Fig.\ \ref{fig:walkthrough-faces-cleverhans} other features may compensate for the `smiling' effect, --- here almost all features other than the smile also affect the decision towards this age group positively, --- this will cause errors for less clear-cut cases. This may explain why the accuracy of the ImageNet-based model is not very high, and can point at the fact that the test set accuracy may drop dramatically on new datasets, e.g.\ comprising more old people smiling.

Instead, we would like our model to be robust to a subject's mood when predicting his or her age. We thus need to find a way to prevent Clever Hans effects, e.g.\ prevent the model to associate smiling with age. One reason the model has learned that connection in the first place is the extreme population imbalance among the age groups of the Adience dataset; a problem which is shared with many other datasets of face images, e.g.\ \cite{LFWTech, rothe2015dex}.
We therefore add a second pre-training phase in between the ImageNet initialization and the \emph{actual} training based on the Adience data, by using the considerably larger IMDB-WIKI~\cite{rothe2015dex} dataset.
The IMDB-WIKI dataset consists of 523,051 images from 20,284 celebrities (460,723 images from the Internet Movie Data Base (Imdb) and 62,328 images from Wikipedia) at different ages, labelled with 101 labels (0-100 years, one label per year).
The IMDB-WIKI dataset also suffers from highly imbalanced label populations. However, we follow~\cite{rothe2015dex} and re-normalize the age distribution by under-sampling the more frequent classes until approximately 260,000 samples are selected overall. Furthermore, we assume that since the \mbox{IMDB-WIKI} dataset is composed of photos of public figures (taken at publicized events) the ratio of expressed smiles in higher age groups will be more frequent than in the Adience dataset, which has been captured `in the wild'. A comparison of performance on the Adience benchmark of the original model (pretrained on ImageNet only) and the improved model is given in the table below. 
%\begin{table}
%    \small
    \begin{center}
    \small
    %\caption{Generalization performance of the VGG-16 with different pretraining.}
    \begin{tabular}{lccc}\toprule
         & accuracy & 1-off\\\midrule
        ImageNet pretrained & 56.5 & 90.0 \\
        IMDB-WIKI pretrained & 63.0 & 96.0 \\
        \bottomrule
    \end{tabular}
    \label{table:face}
    \end{center}
%\end{table}
We observe that the additional and more domain-specific IMDB-WIKI pretraining step has improved the generalization performance of the VGG-16 model. Furthermore, we will see that it also prevented the model from associating smiling exclusively with younger age groups.
Fig.~\ref{fig:walkthrough-faces-cleverhans}~\emph{(bottom)} shows LRP heatmaps for all four examples and age label (60+).
For the woman, the model has shifted its attention from the hair and clothes to the face region and neck, and no longer considers the smile as contradictory to the class.
A similar effect can be observed for the samples showing the male person.
The model's age prediction capabilities can no longer be fooled by just smiling into the camera.
However, by introducing the IMDB-WIKI pretraining step, we have apparently replaced the smile-related Clever Hans strategy with another one, related to the presence of glasses in images of males in higher age groups. This leads to our third recommendation:
\begin{center}
\fbox{\bf Iteratively validate and improve the model}
\end{center}
We can do so until the model solely relies for its predictions on meaningful face features. For that, choosing a better pretraining may not be sufficient, and other more advanced interventions may be required.

%%%%%%%%%%%
\subsection{Example 2: Identifying Male and Female Speech Features}
After demonstrating how explanations can be used to unmask Clever Hans strategies, or more generally validate a classifier, we will now discuss another use case, where explanations are this time applied not to get a better model, but to gain new (scientific) insights.
In this worked-through example, we will show that explanations can be used to identify male and female features in speech.

Before going into the analysis, let us first introduce the data and the model used for the speaker's male vs.\ female classification task. 
As training data we use the recently recorded \mbox{AudioMNIST}~\cite{becker2018interpreting} dataset, comprised of 30000  audio recordings of spoken digits from 60 different speakers, with 50 repetitions per digit and speaker, in a 48kHz sampling frequency.
Next to annotations for spoken digit (0-9) and sex of speaker (48 male, 12 female), the dataset provides labels for speaker age, accent and origin.
We begin by training a deep neural network model on the raw waveform data,
which is first downsampled to 8kHz, 
and randomly padded with zeroes before and after the recorded signal to obtain a
8000 dimensional input vector per sample.
A CNN architecture comprised of six 1d-convolution layers interleaved with max-pooling layers and topped of with three fully connected layers~\cite{becker2018interpreting} and ReLU activation units after each weighted layer is prepared for optimization.
In order to prevent the model from overfitting on the more frequent population of samples labelled as `male', we (randomly) select 12 speakers from both classes.
The model is then trained and evaluated in a 4-fold cross-validation setting, in which the 24 speakers are grouped into four sets of 3 male speakers and 3 female speakers.
Each of the four splits thus contains 1000 waveform features.
Two folds are used for training, while one of the remaining data splits are reserved for validation and testing.
The model reaches an average test set accuracy ($\pm$~standard deviation) of $91.74\%~\pm~8.60\%$ across all splits.

\begin{figure}[ht]
    \centering
    \includegraphics[width=\linewidth]{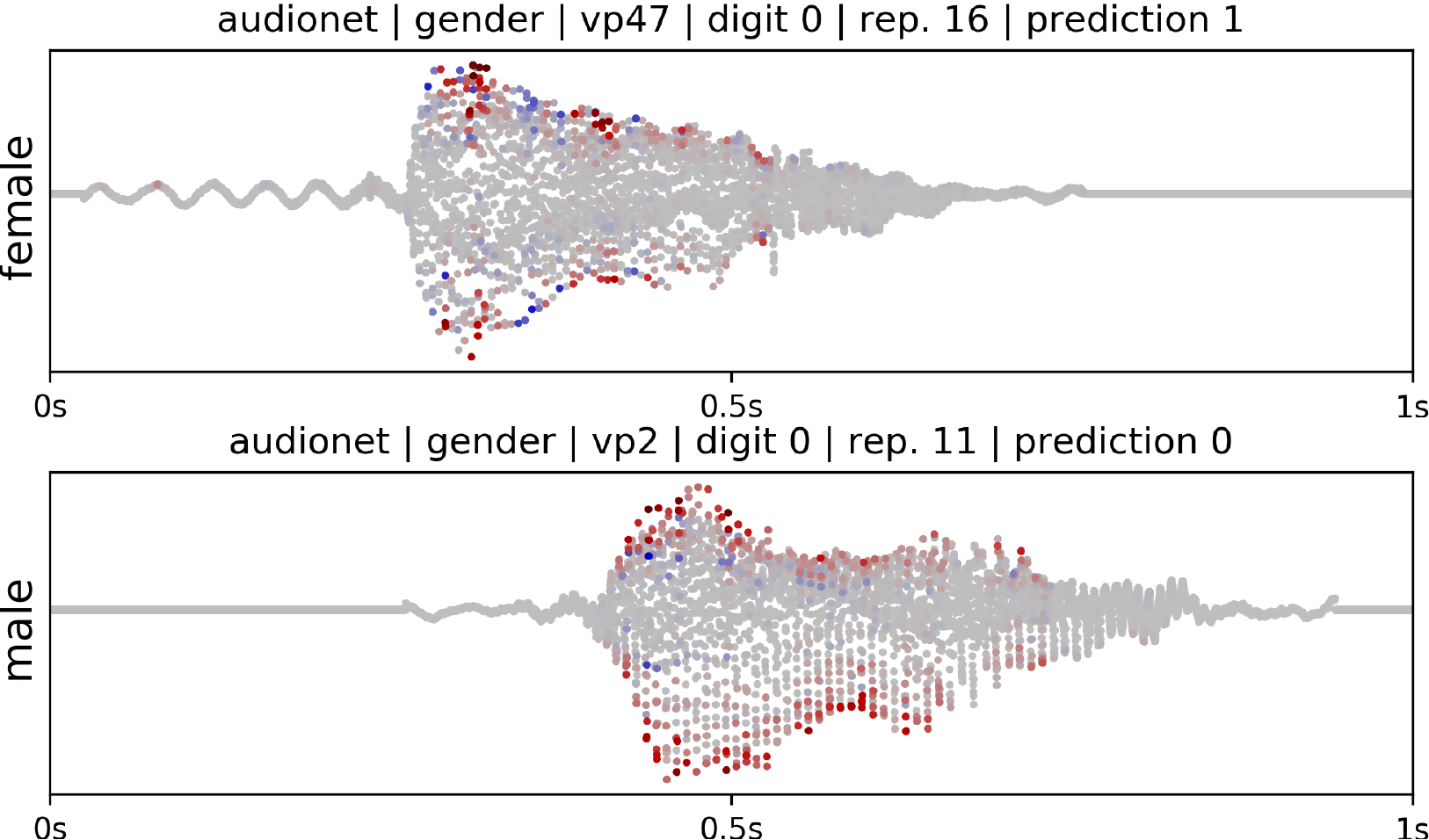} 
    \caption{Explanations based on waveform representation of speech data.
    Correct prediction of a female (\emph{top}) and male (\emph{bottom}) subject.
    The waveform data is visualized as a scatter plot of 8000 discrete measurements, color coded according to relevance attribution for the true class label.}
    \label{fig:walkthrough-audio-waveform-correct}
\end{figure}

With the goal of understanding the data better by explaining the model,
we consider two examples predicted by the network to be male and female and apply LRP to visualize those predictions. Here, the waveform is represented as a scatter plot where each time step is color-coded by its relevance. Results are shown in Fig.~\ref{fig:walkthrough-audio-waveform-correct}.

The explanations reveal that the model predominantly uses the \emph{outer hull} of the waveform signal for decision making.
For a human observer, however, these explanations are difficult to interpret due to the limited accessibility of the data representation in the first place (see Fig.~\ref{fig:walkthrough-audio-waveform-correct}). 
Although the model performs reasonably well on waveform data, it is hard to obtain a deeper understanding beyond the network's modus operandi based on relevance maps, due to the limitations imposed by the data representation itself. We therefore opt to change the data representation for improved interpretability.

\begin{center}
\fbox{\bf Make your input features interpretable}
\end{center}
More precisely, we exchange the raw waveform representation of the data with a corresponding \mbox{228 $\times$ 230} \mbox{(time $\times$ frequency)} shaped spectrogram representation by applying a short-time Fourier transform (time segment length of 455 samples, with 420 samples overlap), cropped to a \mbox{227 $\times$ 227} matrix by discarding the highest frequency bin and the last three time segments.
Consequently we also exchange the neural network architecture and use an AlexNet~\cite{DBLP:conf/nips/KrizhevskySH12} model, which is able to process the transformed input data using 2d-convolution operators.

\begin{figure}[ht]
        \centering
        \includegraphics[trim=0 300 0 0, clip,width=\linewidth]{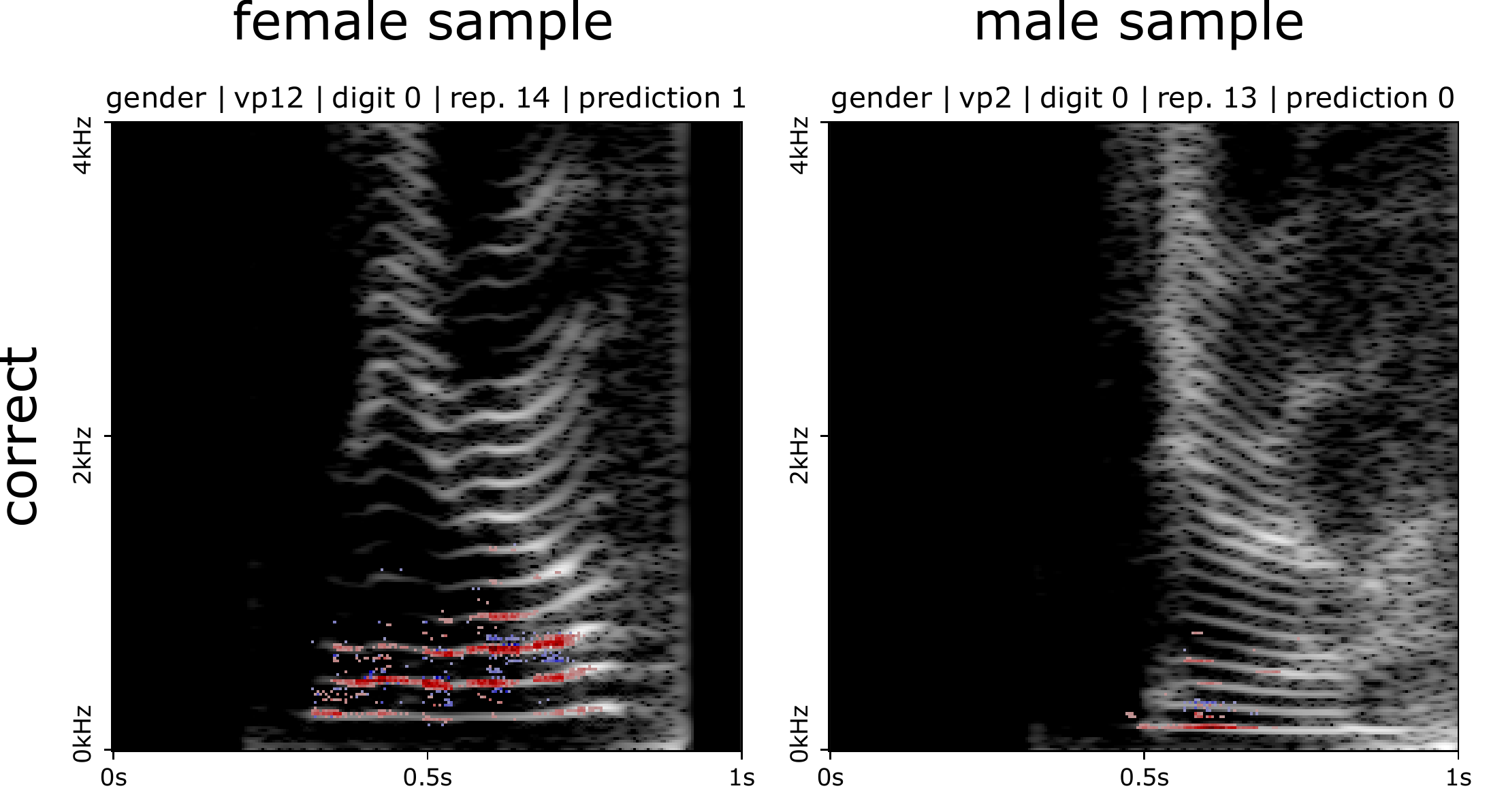}
        \caption{\emph{Left:} Spectrogram representation of digit `zero' spoken by female speaker `vp12'.
        \emph{Right:} Spectrogram representation of digit `zero' spoken by male speaker `vp2'. 
        Relevance maps are shown w.r.t.\ to the samples' true classes.
        }
        \label{fig:walkthrough-audio-spectrograms}
    \end{figure}

Fig.~\ref{fig:walkthrough-audio-spectrograms} visualizes two input spectrograms, with corresponding relevance maps (only relevance values with more than 10\% relative amplitude) drawn on top. 

Heatmap visualizations based on spectrogram input data are more informative than those for waveform data and reveal that the model has learned to distinguish between male and female speakers based on the lowest fundamental frequencies (male speakers, Fig.~\ref{fig:walkthrough-audio-spectrograms}~(\emph{right})),
and immediate harmonics (female speakers, Fig.~\ref{fig:walkthrough-audio-spectrograms}~(\emph{left})) shown in the spectrogram.
Many incorrectly classified samples with ground truth label `male' show large gaps between frequency bands often occurring in samples from female speakers.
Note that these insights are consistent with the literature \cite{traunmuller1995frequency}.

\begin{center}
\fbox{\bf Gain insights by explaining predictions}
\end{center}

As a noteworthy side effect, the increase in interpretability from switching from a waveform data representation to spectrogram data representation does \emph{not} come at a price of model performance.
On the contrary, model performance is even increased slightly from $91.74\%$ to $95.87\%$.

\smallskip
 
For applications where a change of input representation into images is not possible, e.g.\ because the training data are not available or because the human-interpretable image representation does not allow for training an accurate model, we can try to increase interpretability by different means. For our speech example, that could consist of segmenting the time series into meaningful phases, as done in ECG analysis, and compute relevance w.r.t.\ these phases rather than the individual time points. Furthermore, the data and the explanations can be projected in different domains (e.g.\ similar to the TCAV approach \cite{DBLP:conf/icml/KimWGCWVS18}). In our speech example, projecting the time series and the explanations on a Fourier basis would for example allow for understanding frequencies that really matter for the decision. All this ``post-precessing'' of the explanation can help to understand the results, even though the input features are hardly interpretable.

%%%%%%%%%%%
\subsection{Example 3: Identifying Interacting Mortality Risk Factors}
\label{sec:shapexample}
The third worked-through example demonstrates possible limitations of first-order explanation methods and the need for a detailed (higher-order) analysis of explanation results, especially when aiming to draw causal conclusions on the basis of the explanations.
The example is adapted from the work of Lundberg et al.\ \cite{lundberg2020local}. It concerns a medical application, in particular, a statistical model used for predicting mortality called `cox proportional hazards model' \cite{Cox1972}, fitted on data from NHANES I with follow-up mortality data from the NHANES I Epidemiologic Follow-up Study \cite{cox1992plan}. The mortality dataset consists of 14,407 individuals and 79 features and the authors use a tree-based model\footnote{In \cite{lundberg2020local} the authors also propose a polynomial time algorithm for computing the Shapley value explanations on tree-based models.} and a Shapley additive explanation (SHAP) method. 

\begin{figure}[ht]
    \centering
    \includegraphics[width=1\linewidth]{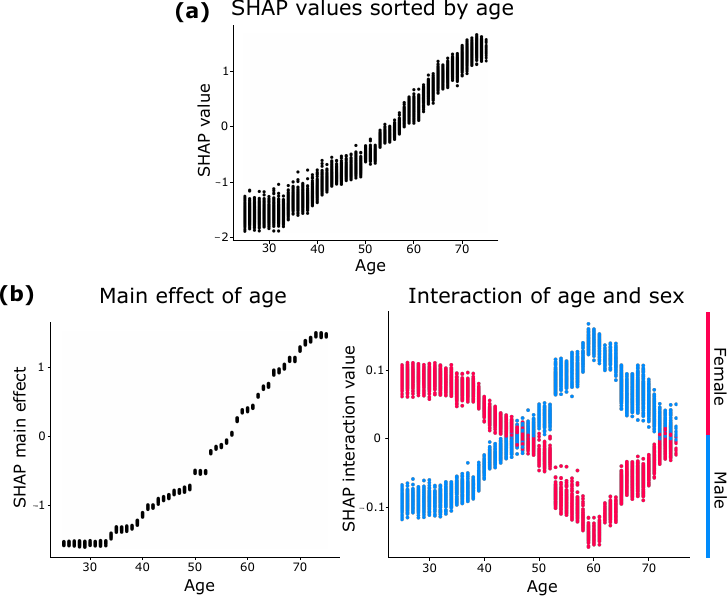}
    \caption{Identifying mortality risk factors from explanations. {\bf (a)} First-order SHAP values show that high age correlates with high mortality risk. The impact of other interacting factors on the age contribution to risk remain hidden. {\bf (b)} The SHAP interaction matrix can be decomposed into the main effects (diagonal elements) and the interaction effects (off-diagonal elements). Here the interaction effect shows that the risk factor ``age'' is largely modulated by the sex of the participant. (Figure is adapted from \cite{lundberg2020local}.)}
    \label{fig:shap}
\end{figure}

Fig.~\ref{fig:shap} (a) shows the computed SHAP values for one of the considered features, namely the age of the participants. Consistent with common medical knowledge, the SHAP values indicate a higher mortality risk for people with higher age.
The vertical dispersion at a single age value results from interaction effects in the model, which are hidden when only considering the first-order explanations. In this dataset the sex of the participant modulates the risk associated with the factor ``age''. This interesting relation can be derived from the SHAP interaction values.

Fig.~\ref{fig:shap} (b) shows the main effect of age (left plot) and the interaction effect between age and sex (right plot) for the risk of mortality. These values constitute the diagonal and off-diagonal elements of the SHAP interaction matrix, respectively. One can see that age is on its own a very strong risk factor, however, depending on the sex of the person this risk factor is modulated differently over different age groups. While at a young age the mortality risk is slightly higher for women, it reverses and is much higher for men at age 60, and finally reaches the same level for the high age group. This example therefore shows that by only considering first-level explanations, we may miss important details and arrive at incomplete conclusions. Therefore, we recommend whenever possible to look at these interactions.

\begin{center}
\fbox{\bf Look at interactions to get deeper insights}
\end{center}

We note that a general limitation of higher-order explanations is that they need more data to be extracted with statistical significance. This is not a problem when considering models taking few input features and trained on large datasets, but this can become more challenging for high-dimensional data, and in that case, only first-order effects can be captured robustly.
%Unfortunately, in many situations we have to live with this incompleteness, because higher-order explanations are in most cases infeasible to compute (Tree-based models are an exception).
Furthermore, even with interaction effects, performing causal inference, e.g.\ deriving mortality risk factors from epidemiological data, remains difficult due to unobserved effects such as collider or measurement bias \cite{hernan2009invited}.

%%%%%%%%%%%%%%%%%%%%%%%%%%%%%%%%%%%%%%%%%%%%%%%%%%%%%%%%%%%%%%%%%%%%%%%%%%%%%%%%%%%%
\section{Successful Uses of Explanation Techniques}
\label{section:applications}

Interpretation methods can be applied for a variety of purposes. Some works have aimed to understand the model's prediction strategies, e.g.\ in order to validate the model \cite{lapuschkin2019unmasking}. Others visualize the learned representations and try to make the model itself more interpretable \cite{Hong2019}. Finally, other works have sought to use explanations to learn about the data, e.g.\ by visualizing interesting input-prediction patterns extracted by a deep neural network model in scientific applications \cite{10.3389/fnins.2019.01321}. Technically, explanation methods have been applied to a broad range of models ranging from simple bag-of-words-type classifiers or logistic regression \cite{BachPLOS15, caruana2015intelligible} to feed-forward or recurrent deep neural networks \cite{BachPLOS15,shrikumar2016not,ArrWASSA17, DBLP:series/lncs/ArrasAWMGMHS19}, and more recently also to unsupervised learning models \cite{Kauffmann19, Kauffmann20}. At the same time these methods were able to handle different types of data, including images \cite{BachPLOS15}, speech \cite{becker2018interpreting}, text \cite{ArrPLOS17,ding2017visualizing}, and structured data such as molecules \cite{schutt2017quantum,Schnake2020Graphs} or genetic sequences~\cite{vidovic2015opening}.

Some of the first successes in interpreting deep neural networks have occurred in the context of image classification, where deep convolutional networks have also demonstrated very high predictive performance \cite{DBLP:conf/nips/KrizhevskySH12, DBLP:conf/cvpr/HeZRS16}. 
Explanation methods have for the first time allowed to open these ``black boxes'' and obtain insights into what the models have actually learned and how they arrive at their predictions. For instance, the works \cite{DBLP:journals/corr/SimonyanVZ13, DBLP:conf/nips/NguyenDYBC16}---also known in this context as ``deep dreams''---highlighted surprising effects when analyzing the inner behavior of deep image classification models by synthesizing meaningful preferred stimuli. They report that the preferred stimuli for the class `dumbbell' would indeed contain a visual rendering of a dumbbell, but the latter would systematically come with an arm attached to it \cite{mordvintsev15}, demonstrating that the output neurons do not only fire for the object of interest but also for correlated features. 

Another surprising finding was reported in \cite{LapCVPR16}. Here, interpretability---more precisely the ability to determine which pixels are being used for prediction---helped to reveal that the best performing ML model in a prestigious international competition, namely the PASCAL visual object classification (VOC) challenge, was actually relying partly on artefacts. The high performance of the model on the class ``horse'' could indeed be attributed to detecting a copyright tag present in the bottom left corner of many horse images of the dataset\footnote{The presence of these artifacts in the benchmark dataset had gone unnoticed for almost a decade.}, rather than detecting the actual horse in the image. Other effects of similar type have been reported for other classes and datasets in many other works, e.g.\ in \cite{ribeiro2016should} models were shown to distinguish between the class ``Husky'' and ``Wolf'' solely based on the presence or absence of snow in the background. %\todo{not sure if the husky vs. wolf is the best example from related work, because they explicitly hand-selected examples of each class so this was not a true dataset. Maybe replace by paper ` ``A Simple Method to Determine if a Music
%Information Retrieval System is a ``Horse'' ', where they refer to the Clever Hans effect.}

These discoveries have been made rather accidentally by researchers carefully analysing suspicious explanations. It is clear that such laborious manual inspection of heatmaps does not scale to big datasets with {\em millions} of examples. Therefore, systematic approaches to the interpretation of ML models have recently gained increased attention.

%%%%%%%%%%%
\subsection{From Explanations to Understanding Large ML Models}
\label{sec:systematic}
This section describes two examples of a systematic analysis of a large number of heatmaps. In the first case, the goal of the analysis is to systematically find data artefacts picked up by the model (e.g.\ copyright tags in horse images), whereas the second analysis aims to carefully investigate the learning process of a deep model, in particular the emergence of novel prediction strategies during training.

\begin{figure}[ht]
    \centering
    \includegraphics[width=1.0\linewidth]{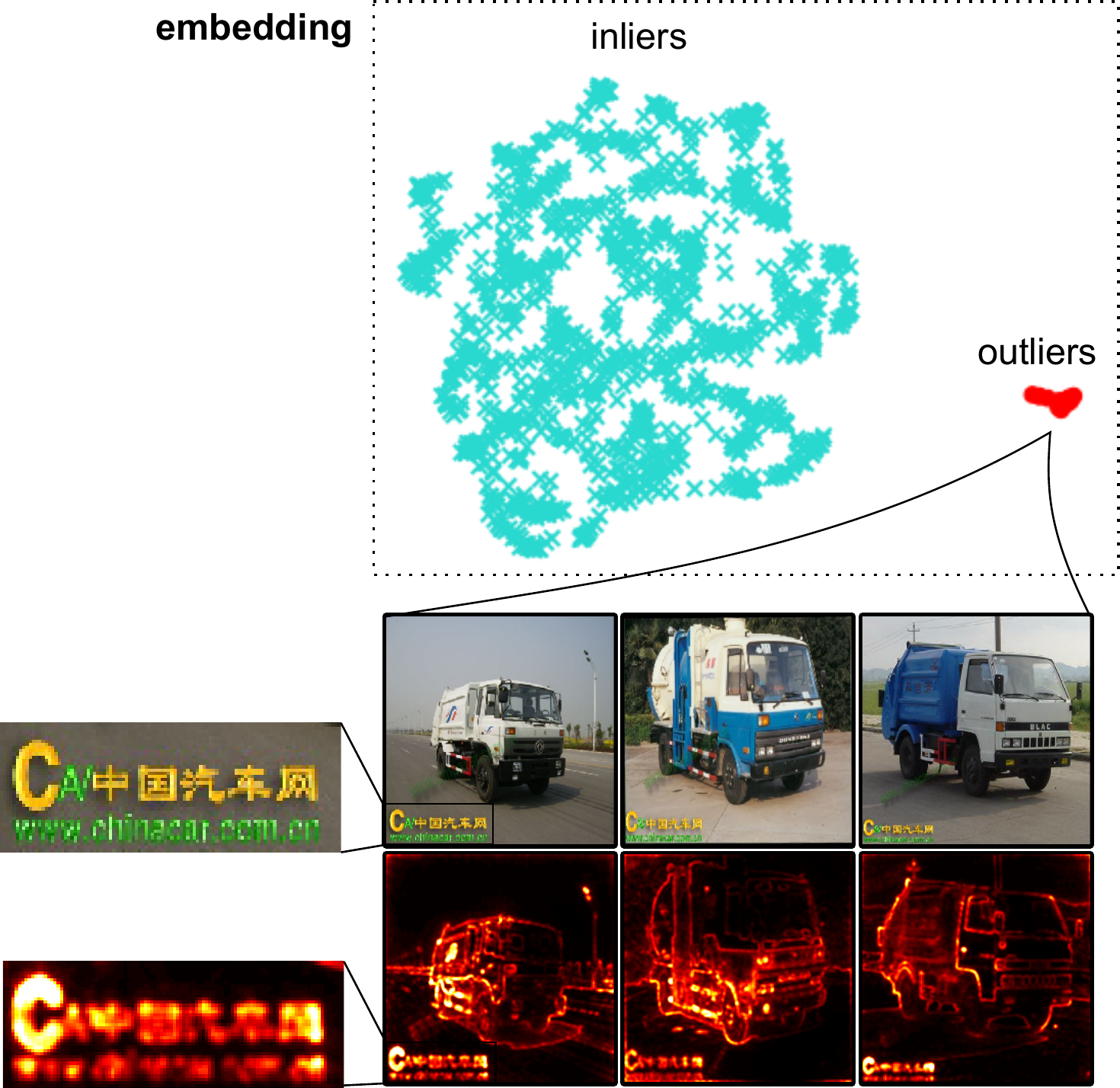}
    \caption{SpRAy analysis of the predictions of a pretrained VGG-16 model on images of the class `garbage truck'. {\it Top}: Low-dimensional embedding of the explained decisions for the class `garbage truck'. Points highlighted in red are outliers. {\it Bottom}: Images and corresponding decisions for some of the points highlighted in red.}
    \label{fig:spray}
\end{figure}

The process of systematically extracting data artefacts was automated by a method called Spectral Relevance Analysis (SpRAy) \cite{lapuschkin2019unmasking}, where after computing LRP-type explanations on a whole dataset (cf.\ Section \ref{section:datasetwide}), a cluster-based analysis was applied on the collection of produced explanations to extract \emph{prototypical} decision strategies. The SpRAy analysis would for example reveal for some shallow Fisher Vector model trained on Pascal VOC\,2007 dataset that images of the label `horse' would be predicted as such using a finite number of prototypical decision strategies ranging from detecting the horse itself to detecting weakly related features such as horse racing poles, or clear artefacts such as copyright tags~\cite{LapCVPR16}. The analysis was later on applied to the decisions of a state-of-the-art VGG-16 deep neural network classifier trained on ImageNet, and here again, interesting insight about the decision structure could be identified \cite{anders2019analyzing}. Certain predictions, e.g.\ for the class `garbage truck', could be found by SpRAy to rely on some watermark in the bottom-left corner of the image (see Fig.\ \ref{fig:spray}). This watermark which is only present in specific images would thus be used by the model as a confounding factor (or artefact) to artificially improve prediction accuracy on this benchmark\footnote{Or in the case of \cite{anders2019analyzing} \emph{deteriorate} model performance, as the identified confounding feature is exclusive to the training data.}.

 Such strategy employed by the ML classifier can be referred to as `Clever Hans' \cite{lapuschkin2019unmasking}. For machine learning models having implemented a Clever Hans strategy, an overconfident assessment of the true model accuracy would be produced by solely relying on the accuracy metric without an inspection of the model's decision structure. The model would have likely performed erratically once it is applied in a real-world setting, where, e.g.\ the copyright tag is decoupled from the concept of a horse or garbage truck respectively. Here, the ability to explain the decision-making of the model and to automatically analyze these explanations on a very large dataset, was therefore a key ingredient to more robustly assess the model's strength and weakness and potentially improving it.

\begin{figure}[ht]
    \centering
    \includegraphics[width=\linewidth]{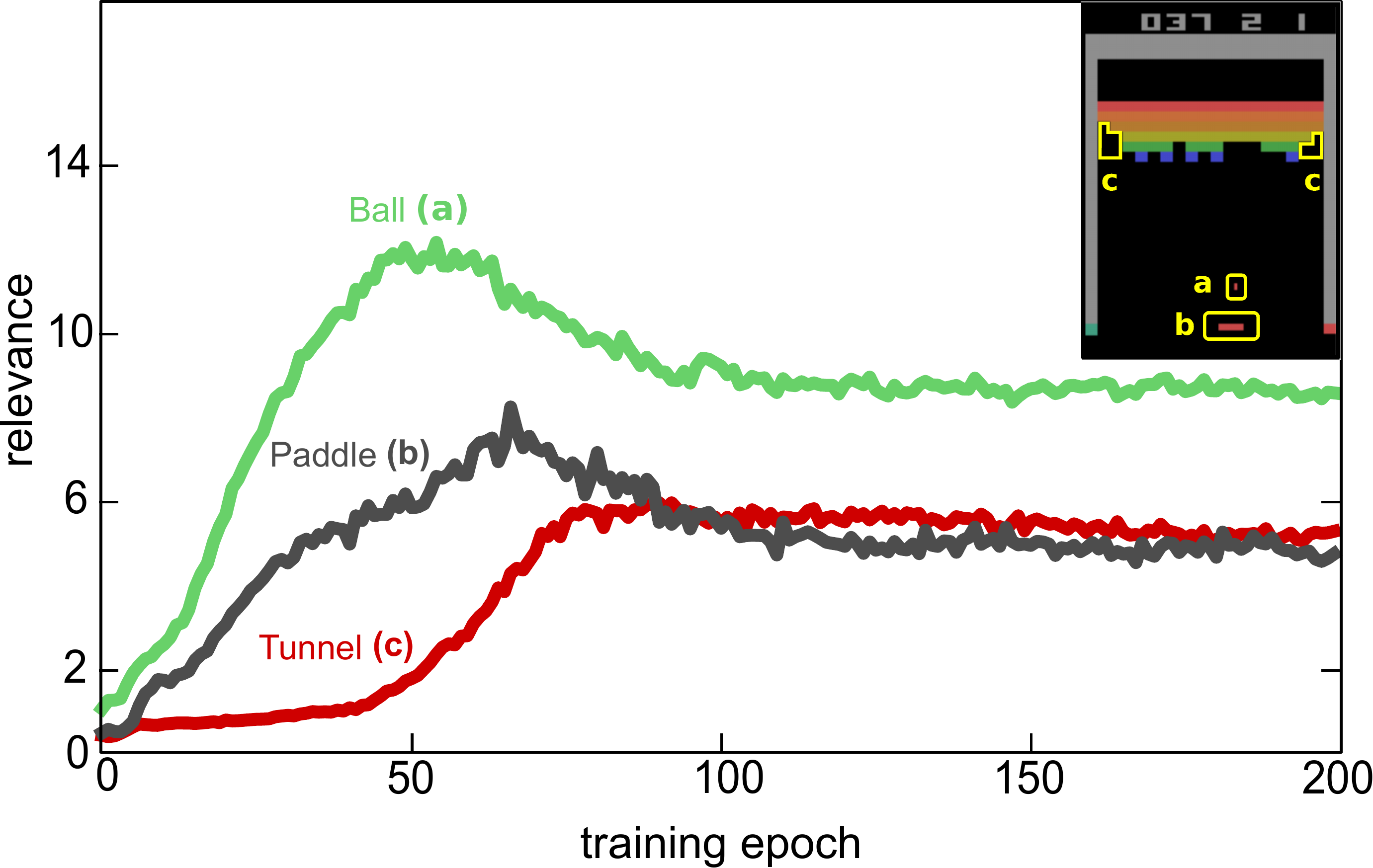}
    \caption{Analysis of the learning process of a deep model playing Atari Breakout. The curves show the development of the relative relevance of different game objects (ball, paddle, tunnel) averaged over six runs.}
    \label{fig:breakout}
\end{figure}

Another example of a systematic interpretation of ML models can be found in the context of reinforcement learning, in particular board and video games. Here large amounts of data can be easily generated (simulated games) and used to carefully analyze the strategies of a ML model and how these strategies emerge during training.
On games such as the arcade game Atari Breakout, the computer player would progressively learn strategies commonly employed by human players such as `tunnel-digging' \cite{DBLP:journals/nature/MnihKSRVBGRFOPB15,DBLP:conf/icml/ZahavyBM16}.  The work of \cite{lapuschkin2019unmasking} analyzes the emergence of this advanced `tunnel-digging' technique using explanations. First, LRP-type pixel-wise explanations of the player's decision were produced at various time steps and training stages. The produced collection of explanations were then pooled (cf.\ Section \ref{section:pooling}) on bounding boxes representing some key visual elements of the game, specifically, the ball, the paddle, and the tunnel. Pooled quantities could then be easily and quantitatively monitored over the different stages of training. The analysis is shown in Fig.\ \ref{fig:breakout}.

We observe that the neural network model would first learn to play conventionally by keeping track of the ball and the paddle, and only at a later stage of the training process would learn to focus on the tunnel area, allowing the ball to go past the wall and bounce repeatedly in the top area of the screen. This analysis highlights in a way that is easy interpretable to the human the multi-stage nature of learning, in particular, how the learning machine progressively develops increasingly sophisticated game playing strategies. Overall, this summarized information on the decision structure of the model and on the evolution of the learning process could prove to be crucial in learning improved models on purposely consolidated datasets. They could also prove useful for characterizing the different stages of learning and developing more efficient training procedures.

%%%%%%%%%%%%
\subsection{From Explanations to Novel Scientific Insights}
In the last subsection we demonstrated the use of explanation techniques for systematically analysing models and verifying that they have learned valid and meaningful prediction strategies. Once verified to not be Clever Hans predictors, non-linear models offer a lot of potential for the sciences to detect new interesting patterns in the data, which may lead to an improved understanding of the underlying natural structures and processes --- the primary goal of scientists. So far this was not possible, because non-linear models were actually considered to be ``black boxes'', i.e.\ scientists had to resort to the use of linear models (see e.g.\ \cite{DBLP:journals/neuroimage/HaufeMGDHBB14, DBLP:journals/bmcbi/MaSH07}), even if this came at the expense of predictivity. Only recently, the technologies have become available to extract scientific insights from complex nonlinear models \cite{DBLP:journals/access/RoscherBDG20,DBLP:series/lncs/SchuttGTM19,McGovern2019}. In the following we will present a selection of problems where ML explanation techniques bring the full potential of non-linear methods to scientific disciplines.

%%%%%%%%%%% Image
Let us start with the discussion of scientific problems, which concern images. These problems could directly benefit from the advances made in general image recognition in the last years.

\begin{figure*}[t]
    \centering
    \includegraphics[width=\linewidth]{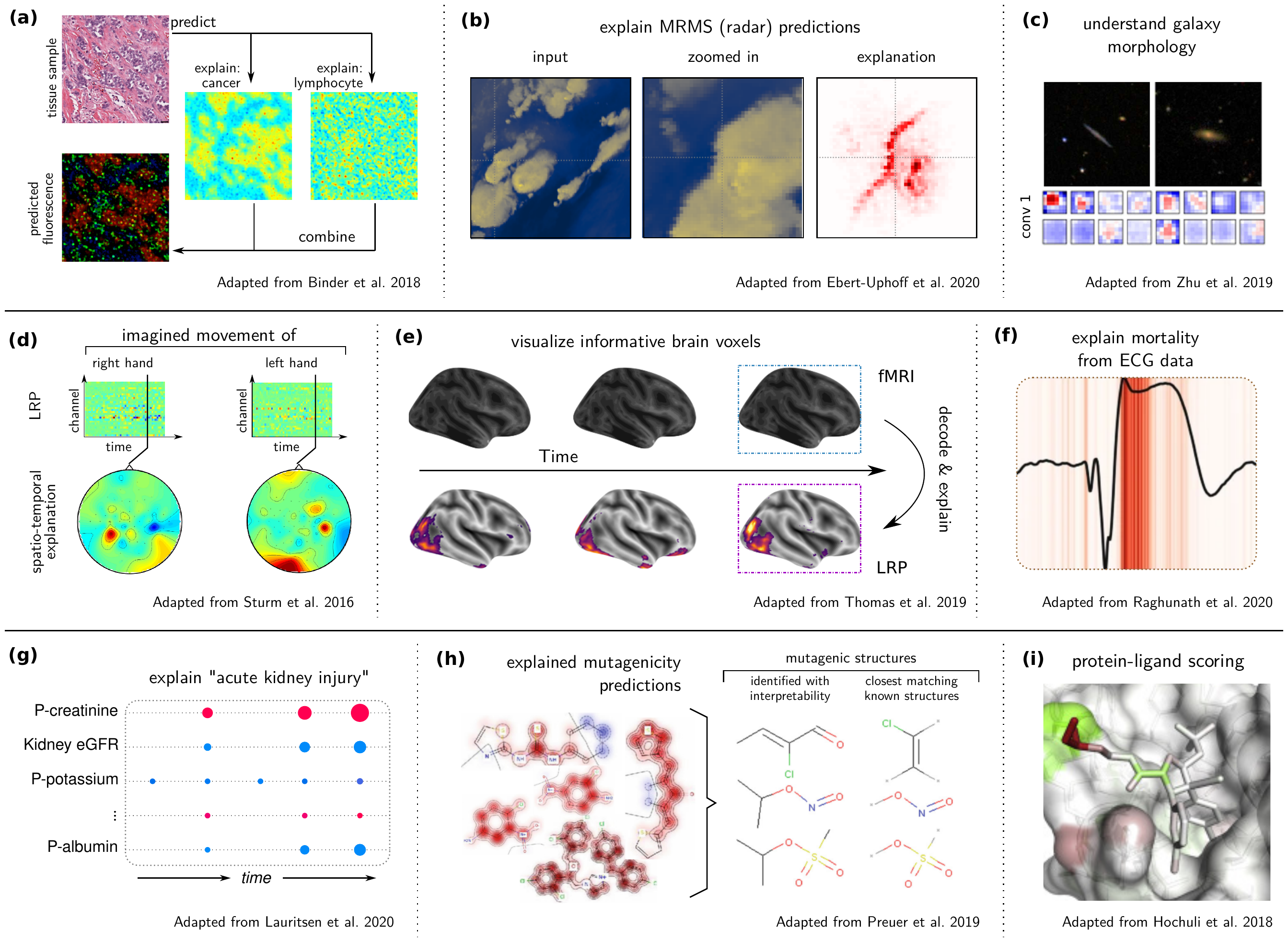}
    \caption{Different applications of explanation techniques in the sciences. \textbf{(a)} LRP heatmaps merged into a computationally predicted fluorescence image. Here, red identifies cancer, green shows lymphocytes and blue is stroma. Adapted from \cite{binder2018towards}. \textbf{(b)} Prediction of MRMS radar signal from  satellite image and pixel-wise explanation. Adapted from \cite{EbertUphoff2020}. \textbf{(c)} Learned filter weights from the first convolutional layer of a deep neural network trained for galaxy morphology classification. Adapted from \cite{zhu2019galaxy}. \textbf{(d)} 
     A whole-brain fMRI volume is decoded using a DNN. The decoding decision is explained voxel-wise to localize brain areas corresponding to the predicted cognitive state. Adapted from \cite{10.3389/fnins.2019.01321}. \textbf{(e)} Example of LRP relevance maps for a single EEG trial of an imagined movement (each class). The matrices indicate the relevance of each time point (abscissa) and EEG channel (ordinate). Below the matrix the relevance information for two single time points is plotted as a scalp topography. Adapted from \cite{StuJNM16}. \textbf{(f)} Prediction of mortality from ECG time series data. Red areas indicate the time steps that most strongly explain the prediction. Adapted from \cite{Raghunath2020}. \textbf{(g)} Some medical condition predicted from electronic health records (EHR) time series, and explained in terms of input features. Blue/red indicate low/high value and circle size indicates feature relevance. Adapted from  \cite{Lauritsen2020}. \textbf{(h)} Graph convolutional neural prediction of the molecule's mutagenicity, and attribution of individual atoms. Interpretability feedback reveals that the model has correctly identified molecular substructures known to interact with (human) DNA. Adapted from \cite{Preuer2019}. \textbf{(i)} The predicted atom score describing protein-ligand interaction is explained with CLRP (green corresponding to a more favorable score). Adapted from \cite{hochuli2018visualizing}.}
    \label{fig:sciences}
\end{figure*}

An important application area is in medicine \cite{holzinger2019causability}. Fig.~\ref{fig:sciences} (a) shows one such application: the task of predicting tissue type from histopathology imagery. The work of \cite{binder2018towards} proposes a bag-of-words model for the prediction task with invariances to rotation, shift and scale of the input data. For the verification of the prediction results, the LRP technique is applied to this model so that heatmaps are produced, offering per-pixel scores which indicate the presence of tumorous structures. Furthermore, LRP heatmaps computed for different target cell types can be combined for obtaining computationally predicted fluorescence images. The explanations are histopathologically meaningful and may potentially give interesting information about which tissue components are most indicative of cancer. Further analyses such as the identification, localization and counting of cells, i.e.\ lymphocytes, can be performed on these explanations (see \cite{klauschen2018scoring}). Recently deeper models have also been used for predicting and explaining tissue type \cite{hagele2019resolving}. In addition to visual explanations, \cite{zhang2019pathologist} also generate a free-text pathology report to clarify the decision of the classifier. Further applications of DNN explanation techniques for medical images include the detection of lesions in diabetic retinopathy data \cite{quellec2017deep}, pixel-wise analysis of microscopy images from global image annotations \cite{KrausBF16}, the validation of predictions in dermatology \cite{young2019deep}, and analysis of x-ray images \cite{gale2018producing}. The latter work produces in addition to the image-level explanations, descriptive sentences to further improve the interpretability of the ML decision for a doctor.

%Various other works apply interpretable machine learning to image-based medical applications \cite{holzinger2019causability}. For instance, \cite{KrausBF16} use deep multiple instance learning to classify and segment microscopy images using only whole image level annotations. The work of \cite{gale2018producing} introduces a model-agnostic interpretation method for the analysis of x-ray images, which not only visualizes the elements that have contributed to each decisions, but also produces descriptive sentences to clarify the decision of the classifier. The combined explanations are well adopted by doctors and are shown to be more informative than the visualisations or generated text alone.

Methods for explaining DNNs have also been applied outside of medical imaging, for example, in earth sciences. In \cite{amt-2020-420,EbertUphoff2020}, the authors predict various meteorological values such as convection, or MRMS radar, from satellite imagery, and the latter prediction are then attributed to the individual pixels. An example is shown in Fig.\ \ref{fig:sciences} (b), where the MRMS output is supported by two visual pattern, the edge of the cloud, and a brighter spot representing a denser region of the cloud. Fig.\ \ref{fig:sciences} (c) shows a machine learning based analysis of galaxy morphologies \cite{zhu2019galaxy}. The latter work aims to classify galaxy morphologies into five classes (completely round smooth, in-between smooth, cigar-shaped smooth, edge-on and spiral) using a convolutional neural networks, and the convolution filters as well as activation patterns are analyzed to understand what visual features are associated to these different classes (see Fig.\ \ref{fig:sciences} (f)). Lastly, the authors of \cite{ghosal2018explainable} also rely on training and explaining DNN image classifiers to get novel insights, this time in the area of plant stress classification.

%%%%%%%%%%% Time Series
Explanation methods for deep neural networks have also demonstrated their potential beyond the image domain, e.g.\ on scientific problems concerning time series data. For instance, the work of \cite{StuJNM16} presents one of the first uses of DNNs explanation techniques in cognitive neurosciences, specifically in brain computer interfacing \cite{dornhege2007toward} where linear methods are still the most widely used filtering methods \cite{blankertz2008optimizing,DBLP:journals/neuroimage/HaufeMGDHBB14}. 
The results in \cite{StuJNM16} show that deep models achieve similar decoding performances\footnote{Deep models usually require larger amounts of training data to have an advantage over linear techniques.} and learn neurophysiologically plausible patterns (see Fig.~\ref{fig:sciences} (b)), namely focus on the contralateral sensorimotor cortex -- an area where the event-related desynchronization occurs during motor imagery. However, in contrast to the patterns computed with conventional approaches \cite{blankertz2008optimizing,DBLP:journals/neuroimage/HaufeMGDHBB14}, which only allow to visualize the aggregated information (average activity) per class, the explanations computed with LRP are available for every single input of the deep learning classifier, i.e.\ for every time point of individual trials (see Fig.~\ref{fig:sciences} (d)). This increased resolution (knowing which sources are relevant at each time point, instead of just having the average patterns) sheds further light onto cognitive processes in the brain.

Another application of DNN explanation techniques in cognitive neuroscience is presented in \cite{10.3389/fnins.2019.01321}, where a deep model is learned to predict whole-brain fMRI data. The method, termed DeepLight, outperforms well-established local or linear decoding methods such as the generalized linear model and searchlight (see \cite{10.3389/fnins.2019.01321}). An adaptation of LRP maintains interpretability and substantiates that the model's predictions are based on physiologically appropriate brain areas for the classified cognitive states. The approach is depicted in Fig.~\ref{fig:sciences} (e) which visualizes the exemplar voxels that are used by the deep model to accurately decode the state from the fMRI signal. These voxels of high relevance have been shown to correspond very well to the active areas described in the fMRI literature (see \cite{10.3389/fnins.2019.01321}). Note that also here the deep model not only gives an advantage in terms of performance (i.e.\ better decoding accuracy) compared to the local or linear baseline methods, but its explanations are provided for every single input, i.e.\ for every fMRI volume over time. This increased resolution allows to study the spatio-temporal dynamics of the fMRI signal and its impact on decoding, something which is not possible with classical decoding methods\footnote{In classical fMRI analyses, p-values indicate the relevance of brain voxels. However, these p-values are usually obtained on a subject or group-level, not for single trials or single time points.}.

Many other studies use explanation methods to analyze time series signals in the sciences. For example, \cite{Raghunath2020} builds a model of mortality from ECG data, and applies DNN explanation techniques to identify for individual subjects' patterns in the ECG time series that support the prediction (cf.\ Fig.\ \ref{fig:sciences} (f) for one such explanation.) In another work \cite{Lauritsen2020}, the input time series is formed by electronic health records (EHR) sampled at a much lower rate, from which a DNN model can decide among a set of medical conditions. The medical condition can then be attributed to the different medical measurements. An example of explanation is given in Fig.\ \ref{fig:sciences} (g). The produced explanation provides valuable additional information to decide in favor or against a particular treatment. Reference \cite{horst2019explaining} introduces explanation techniques to the domain of human gait recognition and show that non-linear learning models are not only the better predictors but that they can at the same time learn physiologically meaningful features for subject prediction which align with expected features used by linear models. Another work \cite{Kratzert2019} applies Long Short-Term Memory (LSTM) networks to the field of hydrology to predict the river discharge from meteorological observations. The authors apply the integrated gradients technique to analyze the internals of the network and obtain insights which are consistent with our understanding of the hydrological system.

%%%%%%%%%%% Molecules
Structured data such as molecules or gene sequences are another very important domain for scientific investigations. Therefore, non-linear ML coupled with explanation techniques have also attracted attention in scientific communities working with this type of data. One successful example of the use of DNN explanation methods in this domain has been reported in \cite{Preuer2019}. The authors train a deep model to predict molecular properties and bioactivities and apply the Integrated Gradients method to analyze what the model has learned and extract interesting insights (see Fig.~\ref{fig:sciences} (h)). For instance, they find that single neurons play the role of pharmacophore detectors and demonstrate that the model uses pharmacophore-like features to reach its conclusions, which are consistent with pharmacologist literature. Another work \cite{hochuli2018visualizing} (see Fig.~\ref{fig:sciences} (i)) applies an extended version of LRP called CLRP to visualize how CNNs interpret individual protein-ligand complexes in molecular modeling. Also here the trained model learns meaningful features and has the ability to provide new insights into the mechanisms underlying protein-ligand interactions. Yet another work \cite{yinchong2018explaining} applies LSTM predictors together with LRP for transparent therapy prediction on patients suffering from metastatic breast cancer. Clinical experts verify that the features used for prediction, as revealed via LRP, largely agree with established clinical guidelines and knowledge. The work by \cite{kelley2018sequential} uses a gradient-based explanation technique to understand the activity prediction across chromosomes, whereas \cite{chong2018mouse} uses the LRP explanation technique for understanding automated decisions on behavioral biometrics.  Recently, also the physics community started to use machine learning explanations for the task of energy prediction. The work of \cite{schutt2017quantum,DBLP:series/lncs/SchuttGTM19} showed that accurate predictions are possible and obtained also physical meaningful insights from the model. Other works \cite{Liu655639} showed that explanations in gene analysis lead to interpretable patterns consistent with literature knowledge.

%%%%%%%%%%%%%%%%%%%%%%%%%%%%%%%%%%%%%%%%%%%%%%%%%%%%%%%%%%%%%%%%%%%%%%%%%%%%%%%%%%%%
\section{Challenges and Outlook}
\label{section:challenges}
While recent years have seen astonishing conceptual and technical progress in XAI, it is important to carefully discuss the current limits and the challenges that will need to be addressed by researchers to further establish the field and increase the usefulness of XAI systems.

Foundational theoretical work in XAI has so far been limited. As discussed above in Section \ref{section:theory}, Shapley values \cite{Shapley,DBLP:conf/nips/LundbergL17}, and (deep) Taylor decomposition \cite{DBLP:journals/pr/MontavonLBSM17} have been proposed as principled frameworks for formalizing the task of explanation. Other frameworks such as rate distortion theory \cite{macdonald2019rate} have also been proposed. Numerous theoretical questions however remain: For example, it remains unclear how to weigh the model and the data distribution into the explanation, in particular, whether an explanation should be based on {\em any} features the model locally reacts to, or only those that are expressed locally. Related to this question is that of causality, i.e.\ assuming a causal link between two input variables, it has not been answered yet whether the two variables, or only the source variable, must constitute the explanation. A deeper formalization and theoretical understanding of XAI will be instrumental for shedding light into these important questions.

Another central question in XAI is that of optimality of an explanation. So far, there is no well-agreed understanding of what should be an optimal explanation. Also, ground-truth explanations cannot be collected by humans as this would presuppose they are able to make sense of the complex ML model they would like to explain in the first place. Methods such as `pixel-flipping' \cite{samek2016evaluating} assess explanation quality indirectly by testing how flipping relevant pixels affects the output score. The `axiomatic approach' \cite{SundararajanTY17,DBLP:series/lncs/Montavon19} does not have this indirect step, however, axioms are usually too generic to evaluate an explanation comprehensively. Approaches relying on ground truth information (e.g.\ \cite{OsmArXiv20}) derived from a synthetic dataset offer a direct and objective way to evaluate and compare explanations, however, it needs to be demonstrated that the synthetic problem is representative of the typically more complex real-world problem, where the ground truth explanations are usually not available. Self-explainable models (e.g.\ \cite{Rudin2019}) provide another potential avenue towards optimal explanations, by forcing the model to be built in a way that explanations can be unambiguously extracted, thereby bypassing some of the technical challenges of post-hoc explanations. Such self-explainable approach comes however at the possible expense of prediction accuracy or runtime efficiency. Hence, the question of producing optimal explanations is still largely an open question. Lastly, accuracy is only one factor in a more general assessment of the overall practical value of an explanation, which further includes human factors such as understandability, manageability, and overall utility of the XAI system \cite{Swartout1993,ribeiro2016should,Naranyan2018}. Application-driven evaluations account for these additional factors, however, they are also hard to implement in practice \cite{doshi2017towards}.

Further challenges arise when applying XAI on problems where different actors (e.g.\ the explainer and the explainee) have conflicting interests. Recent work has shown that an `adversary' can modify the ML model in an imperceptible fashion so that the prediction behavior remains intact but the explanation of those predictions changes drastically \cite{heo2019fooling}. Relatedly, even when the model remains unchanged, inputs could be perturbed imperceptibly to produce arbitrary explanations \cite{DBLP:conf/nips/DombrowskiAAAMK19}.
An important challenge will be to provide provable guarantees on the robustness of explanations to various types of external distortion.
%It  will in particular be interesting to also relate this aspect to techniques for uncertainty quantification or also algorithms for assessing the relevant structural parts in learning models \cite{JMLR:v9:braun08a, JMLR:v12:montavon11a}.

Interpretability may also find itself at odds with the constant quest for higher predicting accuracy. The model development and the explanation of an already trained model are often treated as two independent processes, i.e.\ the model developers aim to build the best possible model and the post-hoc XAI community provides the tools to interpret it.
%Some authors (e.g.\ \cite{Rudin2019}) argue against this approach and instead suggest to make the model interpretable by design. While such an intrinsic interpretability is certainly desirable in many situations, it remains still unclear whether such an approach generalizes competitively (i.e.\ how much performance we loose by such a strong restriction).
Because highly predictive models are becoming increasingly complex both in terms of number of parameters and structure of the model, XAI software must keep up with this increasing complexity \cite{DBLP:journals/jmlr/AlberLSHSMSMDK19}, and at the same time, the human must also deal with explanations of increasingly subtler predictions \cite{baehrens2010explain}. Same challenges also occur for self-explainable models (cf.\ Section \ref{section:others}), where the incorporated interpretability structures, would have to be continuously refined and extended to cope with the increased complexity of data and tasks, and the intrinsic limits of the human receiving the explanations.

When designing new XAI-driven applications, adopting a holistic view that sets the right tradeoffs and delivers the optimal amount of information and range of action to the multiple and potentially conflicting actors, will constitute an important practical challenge.

Another question of utmost importance, especially, in safety critical domains, is whether we can fully trust the model after having explained some predictions. Here, we need to distinguish between model interpretation and model certification: While it is helpful to explain models for available input data, e.g.\ to potentially detect some erroneous decision strategies, certification would require to verify the model for {\em all possible} inputs, not only those included in the data. Furthermore, it must be remembered that explanations returned to the user are summaries of a potentially complex decision process, i.e.\ there may be different decision strategies, the wrong ones and the correct ones, mapping to the same explanation. Lastly, explanations are subject to their own biases and approximations, and they can be manipulated by an adversary to loose their informative content. Therefore, in order to ultimately establish a truly safe and trustworthy model, further steps are needed, potentially including the use of formal verification methods \cite{berkenkamp2017safe,
DBLP:conf/cav/KatzBDJK17}.

Moreover, it may be worthwhile to explore new forms of explanations that are optimally suited to their user. Such explanations could for example leverage the user's prior knowledge or personal preferences. Novel approaches from knowledge engineering, cognitive sciences, and human-computer interfaces, will need to contribute. Also, while heatmaps provide a first intuition to users, they may not take advantage of the complex abstract reasoning capabilities of humans and can be very difficult to interpret for certain data modalities (e.g.\ time series data). As we have discussed
%it may be worth to convert the data to a different representation, e.g.\ from waveforms to spectrograms in order to make the input features more human-interpretable, or to project the explanation in a basis that is more interpretable.
%In other cases, e.g.\ in ECG classification, it may be helpful to segment the time-series into (ECG) phases and provide explanations on this more coarse, but for the cardiologist very familiar, level. Here obtaining interaction values may be very informative, e.g.\ the importance of the relative length of a specific ECG phase (e.g.\ PR or QT interval) or the relative height of a peak over the baseline activity level (R wave).
in the second worked-through example, explaining using a different modality (or by projecting the data on a more interpretable basis) can lead to an explanation that is more useful to the user. An important challenge will be to develop systematic ways of transferring explanations from the original input domain to a more interpretable target domain.

Finally, we see as a key future challenge the design of XAI techniques that can automatically extract meaningful collective variables and explain in their terms. Collective variables are central in many area of physics. In solid state physics, they have led to groundbreaking advances, defining quasiparticles such as phonons, plasmons, polarons, magnons, exitons \cite{Kittel}, etc.
%It is conceivable that ML model trained to model physical systems also builds its own set of collective variables. 
Ideally, collective variables in this sense would be inferred from a learning model by e.g.\ automatically binding explanation variables in meaningful abstract ways. In the neurosciences, von der Malsburg has coined the concept of `binding' for neural strategies that allow sets of variables (neurons) to synchronize collectively by learning \cite{Malsburg}. While steps have already been taken in XAI to identify interacting variables \cite{DBLP:conf/iclr/TsangC018, DBLP:conf/ecai/CuiMK20, Eberle2020Similarity, Schnake2020Graphs,lundberg2020local}, it will be necessary for a broader usage to generalize the approach, e.g.\ to extract `mathematical formulas' that first build the needed collective variables, and use them to explain, concisely but deeply, the ML predictions.

%%%%%%%%%%%%%%%%%%%%%%%%%%%%%%%%%%%%%%%%%%%%%%%%%%%%%%%%%%%%%%%%%%%%%%%%%%%%%%%%%%%%
\section{Conclusion}
\label{section:conclusion}

Complex nonlinear ML models such as neural networks or kernel machines have become game changers in the sciences and industry. Fast progress in the field of explainable AI, has made virtually any of these complex models, supervised or unsupervised, interpretable to the user. Consequently, we no longer need to give up predictivity in favor of interpretability, and we can take full advantage of strong nonlinear machine learning in practical applications.

In this review we have made the attempt to provide a systematic path to bring XAI to the attention of an interested readership. This included an introduction to the technical foundations of XAI, a presentation of practical algorithms such as Occlusion, Integrated Gradients and LRP, concrete examples illustrating how to use explanation techniques in practice, and a discussion of successful applications. We would like to stress that the techniques introduced in this paper can be readily and broadly applied to the workhorses of supervised and unsupervised learning, e.g.\ clustering, anomaly detection, kernel machines, deep networks, as well as state-of-the-art pretrained convolutional networks and LSTMs.

XAI techniques not only shed light into the inner workings of non-linear learning machines, explaining why they arrive at their successful predictions; they also help to discover biases and quality issues in large data corpora with millions of examples \cite{anders2019analyzing}. This is an increasingly relevant direction since modern machine learning relies more and more on reference datasets and reference pretrained models. Furthermore, initial steps have been taken to use XAI beyond validation to arrive at better and more predictive models e.g.\ \cite{DBLP:conf/ijcai/RossHD17,anders2018understanding,arjona2019rudder,anders2019analyzing}.

We would like to stress the importance of XAI, notably in safety critical operations such as medical assistance or diagnosis, where the highest level of transparency is required in order to avoid fatal outcomes.

Finally as a versatile tool in the sciences, XAI has been allowing to gain novel insights (e.g.\ \cite{schutt2017quantum,binder2018towards, hochuli2018visualizing, 10.3389/fnins.2019.01321, escalante2018explainable, DBLP:series/lncs/11700, park2020estimating}) ultimately contributing to further our scientific knowledge.

While XAI has seen an almost exponential rise in interest (and progress) with communities forming and many workshops emerging, there is a wealth of open problems and challenges with ample opportunities to contribute (see Section \ref{section:challenges}). Concluding, we firmly believe that XAI will in the future become an indispensable practical ingredient to obtain improved, transparent, safe, fair and unbiased learning models.

%%%%%%%%%%%%%%%%%%%%%%%%%%%%%%%%%%%%%%%%%%%%%%%%%%%%%%%%%%%%%%%%%%%%%%%%%%%%%%%%%%%%
\section*{Acknowledgment}
 This work was supported by the Institute of Information \& Communications Technology Planning \& Evaluation (IITP) grants funded by the Korea Government (No. 2017-0-00451, Development of BCI based Brain and Cognitive Computing Technology for Recognizing User's Intentions using Deep Learning and No. 2019-0-00079,  Artificial Intelligence Graduate School Program, Korea University), by the German Ministry for Education and Research (BMBF) under Grants 01IS14013A-E, 01GQ1115, 01GQ0850, 01IS18025A and 01IS18037A; and by the German Research Foundation (DFG) under Grant Math+, EXC 2046/1, Project ID 390685689. Correspondence to WS, GM, KRM.

\ifCLASSOPTIONcaptionsoff
  \newpage
\fi

\appendices

\section{Implementing Smooth Integrated Gradients}
\label{appendix:igs}

In this appendix, we present the algorithm combining SmoothGrad \cite{DBLP:journals/corr/SmilkovTKVW17} and Integrated Gradients \cite{SundararajanTY17}, which we use in Section \ref{section:comparison} in our comparison of explanation methods. Its implementation is shown in Algorithm~\ref{alg:ig-smooth}.

\begin{algorithm}[ht]
\caption{Integrated Gradients with Smoothing}
\label{alg:ig-smooth}
\begin{algorithmic}
\STATE $\R = \boldsymbol{0}$
\FOR{$s=1 \dots S$}
\STATE $\widetilde{\x} \sim \mathcal{N}(0,\sigma I)$
\FOR{$t=1 \dots T$}
\STATE $\R = \R + (\x - \widetilde{\x}) \odot \nabla f(\widetilde{\x} + \frac{t-0.5}{T} \cdot (\x - \widetilde{\x}))$
\ENDFOR
\ENDFOR
\RETURN $\frac{1}{TS} \cdot \R$
\end{algorithmic}
\end{algorithm}

The procedure consists of a simple nested loop of $S$ smoothing and $T$ integration steps, where each integration starts at some random location near the origin. Here, we note that these locations are not strict root points. However, in the context of image data, random noise does not significantly change evidence in favor or against a particular class. Thus, the explanation remains approximately complete.

\section{Implementing Layer-wise Relevance Propagation}
\label{appendix:lrp}

In this appendix, we outline two possible implementations of LRP \cite{BachPLOS15,DBLP:series/lncs/MontavonBLSM19}. A first one that is intuitive and based on looping forward and backward over the multiple layers of the neural network. This procedure can be applied to simple sequential structures such as VGG-16 \cite{DBLP:journals/corr/SimonyanZ14a}. The second approach we present is based on `forward hooks' and serves to extend the LRP method to more complex architectures such as ResNet \cite{DBLP:conf/cvpr/HeZRS16}.

\subsection{Standard Sequential Implementation}

The standard implementation is based on the forward-backward procedure outlined in Algorithm \ref{algorithm:lrp}. We focus here on the \texttt{relprop} function of this procedure, which is called at each layer to propagate relevance to the layer below. We give an implementation for the LRP-$0/\epsilon/\gamma$ rules \cite{BachPLOS15,DBLP:series/lncs/MontavonBLSM19} and one for the $z^\mathcal{B}$-rule \cite{DBLP:journals/pr/MontavonLBSM17}. The first three rules can be seen as special cases of the more general rule
$$
R_j = \sum_k \frac{a_j \rho(w_{jk})}{\epsilon + \sum_{0,j} a_j \rho(w_{jk})} R_k
$$
where $\rho(w_{jk}) = w_{jk} + \gamma w_{jk}^+$. This propagation rule can be computed in four steps.

\begin{algorithm}[ht]
\caption{LRP-$0/\epsilon/\gamma$}
\label{algorithm:lrp0eg}
\begin{algorithmic}
\STATE $\boldsymbol{z} = \epsilon + f_l^{\rho}(\ba^{(l-1)})$ \hfill \text{(Step 1)}
\STATE $\boldsymbol{s} = \R^{(l)} \oslash \boldsymbol{z}$ \hfill \text{(Step 2)}
\STATE $\boldsymbol{c} = \nabla \langle \boldsymbol{z}, [\boldsymbol{s}]_\text{cst.} \rangle$ \hfill \text{(Step 3)}
\STATE $\R^{(l-1)} = \ba^{(l-1)} \odot \boldsymbol{c}$ \hfill \text{(Step 4)}
\RETURN $\R^{(l-1)}$
\end{algorithmic}
\end{algorithm}

The first step applies $f_l^{\rho}$, a forward evaluation of a copy of the layer whose parameters have gone through some function $\rho$, and also adds a small positive term $\epsilon$. The second step applies an element-wise division (denoted by "$\oslash$''). The third step is conveniently expressed as a gradient of some dot product $\langle \boldsymbol{z},[\boldsymbol{s}]_\text{cst.} \rangle$ w.r.t.\ the input activations. The notation $[\cdot ]_\text{cst.}$ indicates that the term has been detached from the gradient computation and is therefore treated as a constant. In PyTorch, for example, this can be achieved by calling \texttt{().data}. The \texttt{relprop} function implemented by Algorithm \ref{algorithm:lrp0eg} is applicable for most linear and convolution layers of a deep rectifier network. For the pixel-layer, we use instead the $z^\mathcal{B}$-rule \cite{DBLP:journals/pr/MontavonLBSM17,DBLP:series/lncs/MontavonBLSM19}:
$$
R_i = \sum_j \frac{x_i w_{ij} - l_i w_{ij}^+ - h_i w_{ij}^-}{\sum_i x_i w_{ij} - l_i w_{ij}^+ - h_i w_{ij}^-} R_j
$$
where $l_i$ and $h_i$ are the lowest/highest possible pixel values of $x_i$, and where $w_{ij}^- = \min(0,w_{ij})$. The corresponding implementation is shown in Algorithm \ref{algorithm:lrpzb} and again consists of four steps:

\begin{algorithm}[ht]
\caption{$z^\mathcal{B}$-rule}
\label{algorithm:lrpzb}
\begin{algorithmic}
\STATE $\boldsymbol{z} = f_1(\x) - f_1^+(\boldsymbol{l}) - f_1^- (\boldsymbol{h})$ \hfill \text{(Step 1)}
\STATE $\boldsymbol{s} = \R^{(1)} \oslash \boldsymbol{z}$ \hfill \text{(Step 2)}
\STATE $\boldsymbol{c} = \nabla_{\x,\boldsymbol{l},\boldsymbol{h}} \langle \boldsymbol{z}, [\boldsymbol{s}]_\text{cst.} \rangle$ \hfill \text{(Step 3)}
\STATE $\R^{(0)} = \x \odot \boldsymbol{c}_1
+\boldsymbol{l} \odot \boldsymbol{c}_2
+\boldsymbol{h} \odot \boldsymbol{c}_3
$ \hfill \text{(Step 4)}
\RETURN $\R^{(0)}$
\end{algorithmic}
\end{algorithm}

The functions $f_1^+$ and $f_1^-$ are forward passes on copies of the first layer whose parameters have been processed by the functions $\max(0,\cdot)$ and $\min(0,\cdot)$ respectively.

\subsection{Forward-Hook Implementation}

When the architecture has non-sequential components (e.g.\ ResNet \cite{DBLP:conf/cvpr/HeZRS16}), it is more convenient to reuse the graph traversing procedures readily implemented by the model's existing forward pass and the automatically generated gradient propagation pass. To achieve this, we can implement `forward hooks' at each linear and convolution layers. In this case, we leverage the `smooth gradient' view of LRP (cf.\ Eq.~\eqref{eq:modgradient}) and modify the implementation of the forward pass in a way that it keeps the forward pass functionally equivalent but modifies the local gradient computation. This is achieved by strategically detaching terms from the gradient in a way that calling the gradient becomes equivalent to computing Eq.\ \eqref{eq:modgradient} at each layer. Once the forward functions have been redefined at each layer, the explanation can be computed globally by calling the gradient of the whole function as shown in Algorithm \ref{algorithm:fhook}. (Note that unlike the original function $f(\x)$ the new function that includes the hooks receives three arguments as input: the data point $\x$, and the bounds $\boldsymbol{l}$ and $\boldsymbol{h}$ used by the first layer.)

\begin{algorithm}
\caption{LRP implementation based on forward hooks}
\label{algorithm:fhook}

\smallskip

{\em Forward hook for intermediate layers (LRP-$0/\epsilon/\gamma$)}

\begin{algorithmic}
\STATE $\boldsymbol{z} = \epsilon + f_l^\rho(\ba^{(l-1)})$
\RETURN $\displaystyle \boldsymbol{z} \odot [f_l(\ba^{(l-1)}) \oslash \boldsymbol{z} ]_\text{cst.}$
\end{algorithmic}

\dotfill

\smallskip
{\em Forward hook for the first layer ($z^\mathcal{B}$-rule)}

\begin{algorithmic}
\STATE $\boldsymbol{z} = f_1(\x) - f_1^+(\boldsymbol{l}) - f_1^- (\boldsymbol{h})$
\RETURN $\displaystyle \boldsymbol{z} \odot [f_1(\x) \oslash \boldsymbol{z} ]_\text{cst.}$
\end{algorithmic}

\dotfill

\smallskip
{\em Global LRP computation}

\begin{algorithmic}
\STATE $y = f(\x,\boldsymbol{l},\boldsymbol{h})$
\STATE $\boldsymbol{c}_1,\boldsymbol{c}_2,\boldsymbol{c}_3 = \widehat{\nabla} y$
\STATE $\R = \x \odot \boldsymbol{c}_1 + \boldsymbol{l} \odot \boldsymbol{c}_2 + \boldsymbol{h} \odot \boldsymbol{c}_3$
\RETURN $\R$

\smallskip
\end{algorithmic}
\end{algorithm}

The forward-hook implementation produces exactly the same output as the original function $f(\x)$, but its `gradient', which we denote by $\widehat{\nabla}$ is no longer the same due to the detached terms. As a result, calling the gradient of this function, and recombining it with the input yields the same desired LRP explanation as one would get with the standard LRP implementation, but has now gained applicability to a broader set of neural network architectures.

\section{Explanation Software}
\label{appendix:software}

The attention to interpretability in machine learning has grown frantically throughout the past decade
alongside research on, and the development of computationally efficient deep learning frameworks.
This attention in turn caused a strong demand for accessible and efficient software solutions for out-of-the-box applicability of XAI.
In this section we briefly highlight a collection of software toolboxes released in recent years,
providing convenient access to a plethora of methods of XAI and supporting various computational backends.
A summarizing overview over the presented software solutions is given in Table~\ref{tab:xai-software},
alongside a glossary of methods with respective abbreviations used throughout our review in Table~\ref{tab:xai-glossary}. 

\begin{table*}[!ht]
\centering
\caption{Interpretability software packages by time of release}
\label{tab:xai-software}
\resizebox{\linewidth}{!}{
\begin{tabular}{l|l|l|l|c|c}
    \toprule
    Software Package                                             & Release         & Available from \texttt{https://github.com/} $\dots$                                                   & Compute Backend             & GPU Support                     & Methods   \\
    \midrule
    LRP Toolbox \cite{LapJMLR16}                              & 2016            & \texttt{sebastian-lapuschkin/lrp\_toolbox}     & Caffe                       & \no                             & DCN, DTD, GB, LRP, SA \\
                          &                                               &                                                                                   & numpy/cupy                  & \yes                            & LRP, SA \\
                          &                                               &                                                                                & Matlab                      & \no                             & LRP, SA \\
    \\ % blank line to better separate software packages
    DeepExplain \cite{ancona2018towards}                      & 2017            & \texttt{marcoancona/DeepExplain}               & Keras+TensorFlow            & \yes                            & DLR, IG, LRP-$\epsilon$\\
                          &                                                             &                                                                   &                             &                                 & SA, AS, OCC  \\
    \\ % blank line to better separate software packages
    iNNvestigate \cite{DBLP:journals/jmlr/AlberLSHSMSMDK19}    & 2019            & \texttt{albermax/innvestigate}                 & Keras+TensorFlow            & \yes                            & DCN, DL, DTD, GB,  \\
                          &                                               &                                                                                   &                             &                                 & IG, LRP, PA, PN, SA, SG  \\
                          &                                               &                                                                                   &                             &                                 & Perturbation Analysis  \\
    \\ % blank line to better separate software packages
    TorchRay \cite{fong2019understanding}                  & 2019            & \texttt{facebookresearch/TorchRay}             &  PyTorch                    & \yes                            & DCN, EB, EP \\
                          &                                               &                                                                                   &                             &                                 & GC, GB, RISE, SA, \\
                          &                                               &                                                                                &                             &                                 & various benchmarks\\
    \\ % blank line to better separate software packages
    Captum  \cite{captum2019github}                       & 2019 (beta)     & \texttt{pytorch/captum}                        &  PyTorch                    & \yes                            & DCN, DLR, DLSHAP,  \\
                          &                                               &                 &                                                                                               &                                 & GB, GC, GGC, GSHAP, IG, II,  LC, \\
                          &                                               &                 &                                                                                               &                                 & NC, NGB, NIG, SA, SG, SG-SQ, TC, VG \\
    \bottomrule
\end{tabular}
}
\end{table*}

\begin{table*}[!ht]
    \centering
    \caption{Glossary of interpretability methods with abbreviations referenced throughout our review} 
    \label{tab:xai-glossary}
    \begin{minipage}{0.5\textwidth}
    \centering
    \begin{tabular}{ll|l}
        \toprule
        Method & & Abbrv. \\
        \midrule
         Anchors     & \cite{ribeiro2018anchors} & ANCH \\
         ApproShapley (Shapley Value Sampling)  & \cite{castro2009polynomial} & AS \\
         Class Activation Mapping & \cite{zhou2016learning} & CAM \\
         \scalebox{0.9}[1.0]{Contextual Prediction Difference Analysis} &  \cite{gu2019contextual} & CPDA \\
         DeconvNet & \cite{zeiler2014visualizing} & DCN \\
         DeepLIFT & \cite{shrikumar2017learning} & DL \\
         DeepLIFT (Rescale) & \cite{shrikumar2017learning} & DLR \\
         DeepLIFT SHAP & \cite{DBLP:conf/nips/LundbergL17} & DLSHAP \\
         Deep Taylor Decomposition & \cite{DBLP:journals/pr/MontavonLBSM17} & DTD \\
         ExcitationBackprop & \cite{zhang2018neural} & EB \\
         ExtremalPerturbation &\cite{fong2019understanding} & EP \\
	GNNExplainer & \cite{DBLP:conf/nips/YingBYZL19} & GNNEXP \\
	GNN-LRP & \cite{Schnake2020Graphs} & GLRP \\
         GradCAM & \cite{selvaraju2017visual} & GC \\
         Gradient SHAP & \cite{DBLP:conf/nips/LundbergL17} & GSHAP \\
         Gradient$\,\times\,$Input & \cite{shrikumar2017learning} &   GI \\ 
         GuidedBackprop & \cite{springenberg2015striving} & GB \\
         Guided GradCam & \cite{selvaraju2017visual} & GGC \\
         Integrated Gradients & \cite{SundararajanTY17} & IG \\
         Internal Influence & \cite{leino2018influence} & II \\
	Kernel SHAP & \cite{DBLP:conf/nips/LundbergL17} & KSHAP \\
         LayerConductance & \cite{shrikumar2018computationally} & LC \\
	Local Rule-based Explanations & \cite{guidotti2018local} &  LORE \\
         \bottomrule
    \end{tabular}
    \end{minipage}%
    \begin{minipage}{0.5\textwidth}
    \centering
    \begin{tabular}{ll|l}
        \toprule
        Method & & Abbrv. \\
        \midrule
        Layer-wise Relevance Propagation (full) & \cite{BachPLOS15} & LRP\\
        LRP (composite strategy) & \cite{LapICCVW17,DBLP:series/lncs/MontavonBLSM19} & LRP-CMP \\
         LRP (specific variants) & \cite{BachPLOS15, DBLP:series/lncs/MontavonBLSM19} & LRP-$\ast$\\
         \scalebox{0.85}[1.0]{Local Interpretable Model-agnostic Explanations} & \cite{ribeiro2016should} & LIME \\
         Meaningful Perturbation & \cite{fong2017interpretable} & MP \\
         NeuronConductance & \cite{dhamere2018conductance} & NC \\
         NeuronGuidedBackprop & \cite{springenberg2015striving} & NGB \\
         NeuronIntegratedGradients & \cite{shrikumar2018computationally} & NIG \\
         Occlusion Analysis & \cite{zeiler2014visualizing} & OCC \\
         PatternAttribution & \cite{kindermans2017learning} & PA \\
         PatternNet & \cite{kindermans2017learning} & PN \\
         Prediction Difference Analysis & \cite{zintgraf2017visualizing} & PDA \\
         \scalebox{0.9}[1.0]{Randomized Input Sampling for Explanation} & \cite{petsiuk2018rise} & RISE \\
         Saliency Analysis / Gradient & \cite{baehrens2010explain, DBLP:journals/corr/SimonyanVZ13} & SA \\
         SHapley Additive exPlanations & \cite{DBLP:conf/nips/LundbergL17} & SHAP \\ 
         SHAP Interaction Index & \cite{lundberg2020local} & SHAPIDX \\
         SmoothGrad & \cite{DBLP:journals/corr/SmilkovTKVW17} & SG \\
         SmoothGrad$^2$ & \cite{hooker2019benchmark} & SG-SQ \\
         Spectral Relevance Analysis & \cite{lapuschkin2019unmasking} & SpRAy \\
         TreeExplainer & \cite{lundberg2020local} & TEXP \\
         VarGrad & \cite{adebayo2018local} & VG \\
         Testing with Concept Activation Vectors & \cite{DBLP:conf/icml/KimWGCWVS18} & TCAV \\
         TotalConductance & \cite{dhamere2018conductance} & TC \\
         \bottomrule
    \end{tabular}
    \end{minipage}
\end{table*}

One of the earlier and comprehensive XAI software packages is the LRP Toolbox~\cite{LapJMLR16},
providing presently up to date
%and complete
implementations of
%-- as the name suggests -- 
LRP for the --- until very recently --- popular Caffe deep learning framework~\cite{jia2014caffe},
as well as Matlab and Python via custom neural network interfaces.
While support for Caffe is restricted to the C++ programming language and thus to CPU hardware, it provides functionality implementing DCN, GB, DTD, and SA and can be built and used as a stand-alone executable binary for predictors based on the Caffe neural network format.
The sub-packages available for Matlab and Python provide out-of-the-box support for LRP and SA, while being
easily extensible via custom neural network modules written with clarity and the methods' intelligibility in mind.
The \texttt{cupy}~\cite{cupylearningsys2017} backend constitutes an alternative to the CPU-bound \texttt{numpy}~\cite{oliphant2006guide} package, providing optional support for modern GPU hardware from NVIDIA.

Both the DeepExplain~\cite{ancona2018towards} and iNNvestigate~\cite{DBLP:journals/jmlr/AlberLSHSMSMDK19} toolboxes built on top of the popular Keras~\cite{chollet2015keras} package for Python with TensorFlow backend for explaining Deep Neural Network models, and thus provide support for both CPU and GPU hardware and convenient access for users of Keras models.
While the more recent iNNvestigate Toolbox implements a superset of the modified backpropagation methods available in DeepExplain, the latter also offers functionalty for perturbation-based attribution methods, i.e.\ the Occlusion method~\cite{zeiler2014visualizing} and Shapley Value Resampling~\cite{castro2009polynomial}.
For explaining a model's prediction DeepExplain allows for an ad-hoc selection of the explanation method via pythonic context managers.
The iNNvestigate package on the other hand operates by attaching and automatically configuring (several) modified backward graphs called ``analyzers'' to a model of interest --- one per XAI method to compute attributions with.

A present trend in the machine learning community is a migration to the PyTorch framework with its eager execution paradigm, away from other backends.
Both the TorchRay~\cite{fong2019understanding} and Captum~\cite{captum2019github} packages for Python and PyTorch enable the use of interpretability methods for neural network models defined in context of PyTorch's high level neural network description modules.
Captum can be understood as a rich selection of XAI methods based on modified backprop and is part of the PyTorch project itself.
While not as extensive as Captum, the TorchRay package offers a series  benchmarks for XAI alongside its selection of (benchmarked) interpretability methods.

\bibliographystyle{abbrv}
\bibliography{bibliography}

\begin{thebibliography}{100}

\bibitem{adebayo2018local}
J.~Adebayo, J.~Gilmer, I.~J. Goodfellow, and B.~Kim.
\newblock Local explanation methods for deep neural networks lack sensitivity
  to parameter values.
\newblock In {\em International Conference on Learning Representations}, 2018.

\bibitem{Agarwal_2020_ACCV}
C.~Agarwal and A.~Nguyen.
\newblock Explaining image classifiers by removing input features using
  generative models.
\newblock In {\em Proceedings of the Asian Conference on Computer Vision},
  2020.

\bibitem{DBLP:journals/jmlr/AlberLSHSMSMDK19}
M.~Alber, S.~Lapuschkin, P.~Seegerer, M.~H{\"{a}}gele, K.~T. Sch{\"{u}}tt,
  G.~Montavon, W.~Samek, K.-R. M{\"{u}}ller, S.~D{\"{a}}hne, and P.-J.
  Kindermans.
\newblock i{NN}vestigate neural networks!
\newblock {\em Journal of Machine Learning Research}, 20:93:1--93:8, 2019.

\bibitem{ancona2018towards}
M.~Ancona, E.~Ceolini, C.~\"Oztireli, and M.~Gross.
\newblock Towards better understanding of gradient-based attribution methods
  for deep neural networks.
\newblock In {\em International Conference of Learning Representations (ICLR)},
  2018.

\bibitem{pmlr-v97-ancona19a}
M.~Ancona, C.~Oztireli, and M.~Gross.
\newblock Explaining deep neural networks with a polynomial time algorithm for
  shapley value approximation.
\newblock In {\em Proceedings of the 36th International Conference on Machine
  Learning}, volume~97, pages 272--281, 2019.

\bibitem{anders2018understanding}
C.~J. Anders, G.~Montavon, W.~Samek, and K.-R. M{\"{u}}ller.
\newblock Understanding patch-based learning of video data by explaining
  predictions.
\newblock In {\em Explainable {AI:} Interpreting, Explaining and Visualizing
  Deep Learning}, pages 297--309. Springer, 2019.

\bibitem{anders2019analyzing}
C.~J. Anders, L.~Weber, D.~Neumann, W.~Samek, K.-R. M{\"{u}}ller, and
  S.~Lapuschkin.
\newblock Finding and removing {C}lever {H}ans: Using explanation methods to
  debug and improve deep models.
\newblock {\em arXiv preprint arXiv:1912.11425}, 2019.

\bibitem{arjona2019rudder}
J.~A. Arjona-Medina, M.~Gillhofer, M.~Widrich, T.~Unterthiner, J.~Brandstetter,
  and S.~Hochreiter.
\newblock Rudder: Return decomposition for delayed rewards.
\newblock In {\em Advances in Neural Information Processing Systems}, pages
  13544--13555, 2019.

\bibitem{DBLP:series/lncs/ArrasAWMGMHS19}
L.~Arras, J.~A. Arjona{-}Medina, M.~Widrich, G.~Montavon, M.~Gillhofer, K.-R.
  M{\"{u}}ller, S.~Hochreiter, and W.~Samek.
\newblock Explaining and interpreting lstms.
\newblock In {\em Explainable {AI:} Interpreting, Explaining and Visualizing
  Deep Learning}, volume 11700 of {\em Lecture Notes in Computer Science},
  pages 211--238. Springer, 2019.

\bibitem{ArrPLOS17}
L.~Arras, F.~Horn, G.~Montavon, K.-R. M{\"u}ller, and W.~Samek.
\newblock "{W}hat is relevant in a text document?": An interpretable machine
  learning approach.
\newblock {\em PLoS ONE}, 12(8):e0181142, 2017.

\bibitem{ArrWASSA17}
L.~Arras, G.~Montavon, K.-R. M{\"u}ller, and W.~Samek.
\newblock Explaining recurrent neural network predictions in sentiment
  analysis.
\newblock In {\em Proceedings of the EMNLP'17 Workshop on Computational
  Approaches to Subjectivity, Sentiment \& Social Media Analysis (WASSA)},
  pages 159--168, 2017.

\bibitem{OsmArXiv20}
L.~Arras, A.~Osman, and W.~Samek.
\newblock Ground truth evaluation of neural network explanations with
  clevr-xai.
\newblock {\em arXiv:2003.07258}, 2020.

\bibitem{BachPLOS15}
S.~Bach, A.~Binder, G.~Montavon, F.~Klauschen, K.-R. M{\"u}ller, and W.~Samek.
\newblock On pixel-wise explanations for non-linear classifier decisions by
  layer-wise relevance propagation.
\newblock {\em PLoS ONE}, 10(7):e0130140, 2015.

\bibitem{baehrens2010explain}
D.~Baehrens, T.~Schroeter, S.~Harmeling, M.~Kawanabe, K.~Hansen, and K.-R.
  M{\"{u}}ller.
\newblock How to explain individual classification decisions.
\newblock {\em Journal of Machine Learning Research}, 11:1803--1831, 2010.

\bibitem{DBLP:journals/corr/BahdanauCB14}
D.~Bahdanau, K.~Cho, and Y.~Bengio.
\newblock Neural machine translation by jointly learning to align and
  translate.
\newblock In {\em 3rd International Conference on Learning Representations},
  2015.

\bibitem{DBLP:conf/icml/BalduzziFLLMM17}
D.~Balduzzi, M.~Frean, L.~Leary, J.~P. Lewis, K.~W. Ma, and B.~McWilliams.
\newblock The shattered gradients problem: If resnets are the answer, then what
  is the question?
\newblock In {\em Proceedings of the 34th International Conference on Machine
  Learning}, volume~70 of {\em Proceedings of Machine Learning Research}, pages
  342--350. {PMLR}, 2017.

\bibitem{bau2019visualizing}
D.~Bau, J.~Zhu, H.~Strobelt, B.~Zhou, J.~B. Tenenbaum, W.~T. Freeman, and
  A.~Torralba.
\newblock {GAN} dissection: Visualizing and understanding generative
  adversarial networks.
\newblock In {\em 7th International Conference on Learning Representations},
  2019.

\bibitem{Bau2020}
D.~Bau, J.-Y. Zhu, H.~Strobelt, A.~Lapedriza, B.~Zhou, and A.~Torralba.
\newblock Understanding the role of individual units in a deep neural network.
\newblock {\em Proceedings of the National Academy of Sciences},
  117(48):30071--30078, 2020.

\bibitem{Bazen2013}
S.~Bazen and X.~Joutard.
\newblock The {T}aylor decomposition: {A} unified generalization of the
  {O}axaca method to nonlinear models.
\newblock Working papers, HAL, 2013.

\bibitem{becker2018interpreting}
S.~Becker, M.~Ackermann, S.~Lapuschkin, K.-R. M{\"{u}}ller, and W.~Samek.
\newblock Interpreting and explaining deep neural networks for classification
  of audio signals.
\newblock {\em arXiv preprint arXiv:1807.03418}, 2018.

\bibitem{berkenkamp2017safe}
F.~Berkenkamp, M.~Turchetta, A.~Schoellig, and A.~Krause.
\newblock Safe model-based reinforcement learning with stability guarantees.
\newblock In {\em Advances in Neural Information Processing Systems}, pages
  908--918, 2017.

\bibitem{binder2018towards}
A.~Binder, M.~Bockmayr, M.~H{\"{a}}gele, S.~Wienert, D.~Heim, K.~Hellweg,
  A.~Stenzinger, L.~Parlow, J.~Budczies, B.~Goeppert, D.~Treue, M.~Kotani,
  M.~Ishii, M.~Dietel, A.~Hocke, C.~Denkert, K.-R. M{\"{u}}ller, and
  F.~Klauschen.
\newblock Morphological and molecular breast cancer profiling through
  explainable machine learning.
\newblock {\em Nature Machine Intelligence}, 2021.

\bibitem{bishop1996neuralnets}
C.~M. Bishop.
\newblock {\em Neural Networks for Pattern Recognition}.
\newblock Oxford University Press, Inc., USA, 1996.

\bibitem{blankertz2008optimizing}
B.~Blankertz, R.~Tomioka, S.~Lemm, M.~Kawanabe, and K.-R. M{\"u}ller.
\newblock Optimizing spatial filters for robust eeg single-trial analysis.
\newblock {\em IEEE Signal Processing Magazine}, 25(1):41--56, 2008.

\bibitem{DBLP:conf/iclr/BrendelB19}
W.~Brendel and M.~Bethge.
\newblock Approximating cnns with bag-of-local-features models works
  surprisingly well on imagenet.
\newblock In {\em 7th International Conference on Learning Representations},
  2019.

\bibitem{carlini2017towards}
N.~Carlini and D.~Wagner.
\newblock Towards evaluating the robustness of neural networks.
\newblock In {\em 2017 IEEE Symposium on Security and Privacy (SP)}, pages
  39--57. IEEE, 2017.

\bibitem{DBLP:conf/eccv/CaronBJD18}
M.~Caron, P.~Bojanowski, A.~Joulin, and M.~Douze.
\newblock Deep clustering for unsupervised learning of visual features.
\newblock In {\em 15th European Conference on Computer Vision}, volume 11218 of
  {\em Lecture Notes in Computer Science}, pages 139--156. Springer, 2018.

\bibitem{caruana2015intelligible}
R.~Caruana, Y.~Lou, J.~Gehrke, P.~Koch, M.~Sturm, and N.~Elhadad.
\newblock Intelligible models for healthcare: Predicting pneumonia risk and
  hospital 30-day readmission.
\newblock In {\em Proceedings of the 21th ACM SIGKDD International Conference
  on Knowledge Discovery and Data Mining}, pages 1721--1730, 2015.

\bibitem{castro2009polynomial}
J.~Castro, D.~G{\'o}mez, and J.~Tejada.
\newblock Polynomial calculation of the shapley value based on sampling.
\newblock {\em Computers \& Operations Research}, 36(5):1726--1730, 2009.

\bibitem{chollet2015keras}
F.~Chollet et~al.
\newblock Keras.
\newblock \texttt{https://keras.io}, 2015.

\bibitem{chong2018mouse}
P.~Chong, Y.~X.~M. Tan, J.~Guarnizo, Y.~Elovici, and A.~Binder.
\newblock Mouse authentication without the temporal aspect -- what does a
  2d-cnn learn?
\newblock In {\em 2018 IEEE Security and Privacy Workshops (SPW)}, pages
  15--21. IEEE, 2018.

\bibitem{cox1992plan}
C.~S. Cox.
\newblock {\em Plan and operation of the NHANES I Epidemiologic Followup Study,
  1987}.
\newblock Number~27. US Department of Health and Human Services, Public Health
  Service, Centers for Disease Control, National Center for Health Statistics,
  1992.

\bibitem{Cox1972}
D.~R. Cox.
\newblock Regression models and life-tables.
\newblock {\em Journal of the Royal Statistical Society: Series B
  (Methodological)}, 34(2):187--202, 1972.

\bibitem{DBLP:conf/ecai/CuiMK20}
T.~Cui, P.~Marttinen, and S.~Kaski.
\newblock Learning global pairwise interactions with bayesian neural networks.
\newblock In {\em 24th European Conference on Artificial Intelligence}, volume
  325 of {\em Frontiers in Artificial Intelligence and Applications}, pages
  1087--1094. {IOS} Press, 2020.

\bibitem{deng2009imagenet}
J.~Deng, W.~Dong, R.~Socher, L.-J. Li, K.~Li, and L.~Fei-Fei.
\newblock Imagenet: A large-scale hierarchical image database.
\newblock In {\em Proceedings of the IEEE Conference on Computer Vision and
  Pattern Recognition (CVPR)}, pages 248--255, 2009.

\bibitem{dhamere2018conductance}
K.~Dhamdhere, M.~Sundararajan, and Q.~Yan.
\newblock How important is a neuron?
\newblock {\em arXiv preprint arXiv:1805.12233}, 2018.

\bibitem{DBLP:conf/kdd/DhillonGK04}
I.~S. Dhillon, Y.~Guan, and B.~Kulis.
\newblock Kernel k-means: spectral clustering and normalized cuts.
\newblock In {\em Proceedings of the Tenth {ACM} {SIGKDD} International
  Conference on Knowledge Discovery and Data Mining}, pages 551--556. {ACM},
  2004.

\bibitem{ding2017visualizing}
Y.~Ding, Y.~Liu, H.~Luan, and M.~Sun.
\newblock Visualizing and understanding neural machine translation.
\newblock In {\em Proceedings of the 55th Annual Meeting of the Association for
  Computational Linguistics}, pages 1150--1159, 2017.

\bibitem{DBLP:conf/nips/DombrowskiAAAMK19}
A.~Dombrowski, M.~Alber, C.~J. Anders, M.~Ackermann, K.-R. M{\"{u}}ller, and
  P.~Kessel.
\newblock Explanations can be manipulated and geometry is to blame.
\newblock In {\em Advances in Neural Information Processing Systems}, pages
  13567--13578, 2019.

\bibitem{dornhege2007toward}
G.~Dornhege, J.~d.~R. Mill{\'a}n, T.~Hinterberger, D.~McFarland, K.-R.
  M{\"u}ller, et~al.
\newblock {\em Toward Brain-Computer Interfacing}, volume~63.
\newblock MIT press Cambridge, MA, 2007.

\bibitem{doshi2017towards}
F.~Doshi{-}Velez and B.~Kim.
\newblock A roadmap for a rigorous science of interpretability.
\newblock {\em arXiv preprint arXiv:1702.08608}, 2017.

\bibitem{Eberle2020Similarity}
O.~{Eberle}, J.~{Büttner}, F.~{Kräutli}, K.-R. {Müller}, M.~{Valleriani},
  and G.~{Montavon}.
\newblock Building and interpreting deep similarity models.
\newblock {\em IEEE Transactions on Pattern Analysis and Machine Intelligence},
  pages 1--1, 2020.

\bibitem{EbertUphoff2020}
I.~Ebert-Uphoff and K.~Hilburn.
\newblock Evaluation, tuning, and interpretation of neural networks for working
  with images in meteorological applications.
\newblock {\em Bulletin of the American Meteorological Society},
  101(12):E2149--E2170, 2020.

\bibitem{eidinger2014age}
E.~Eidinger, R.~Enbar, and T.~Hassner.
\newblock Age and gender estimation of unfiltered faces.
\newblock {\em IEEE Transactions on Information Forensics and Security},
  9(12):2170--2179, 2014.

\bibitem{escalante2018explainable}
H.~J. Escalante, S.~Escalera, I.~Guyon, X.~Bar{\'o},
  Y.~G{\"u}{\c{c}}l{\"u}t{\"u}rk, U.~G{\"u}{\c{c}}l{\"u}, and M.~van Gerven.
\newblock {\em Explainable and Interpretable Models in Computer Vision and
  Machine Learning}.
\newblock Springer, 2018.

\bibitem{fong2019understanding}
R.~Fong, M.~Patrick, and A.~Vedaldi.
\newblock Understanding deep networks via extremal perturbations and smooth
  masks.
\newblock In {\em Proceedings of the IEEE International Conference on Computer
  Vision}, pages 2950--2958, 2019.

\bibitem{fong2017interpretable}
R.~C. Fong and A.~Vedaldi.
\newblock Interpretable explanations of black boxes by meaningful perturbation.
\newblock In {\em {IEEE} International Conference on Computer Vision ({ICCV})},
  pages 3449--3457, 2017.

\bibitem{gale2018producing}
W.~Gale, L.~Oakden{-}Rayner, G.~Carneiro, L.~J. Palmer, and A.~P. Bradley.
\newblock Producing radiologist-quality reports for interpretable deep
  learning.
\newblock In {\em 16th {IEEE} International Symposium on Biomedical Imaging},
  pages 1275--1279, 2019.

\bibitem{DBLP:journals/csur/GamaZBPB14}
J.~Gama, I.~Zliobaite, A.~Bifet, M.~Pechenizkiy, and A.~Bouchachia.
\newblock A survey on concept drift adaptation.
\newblock {\em {ACM} Computing Surveys}, 46(4):44:1--44:37, 2014.

\bibitem{ghosal2018explainable}
S.~Ghosal, D.~Blystone, A.~K. Singh, B.~Ganapathysubramanian, A.~Singh, and
  S.~Sarkar.
\newblock An explainable deep machine vision framework for plant stress
  phenotyping.
\newblock {\em Proceedings of the National Academy of Sciences},
  115(18):4613--4618, 2018.

\bibitem{DBLP:journals/jmlr/GlorotBB11}
X.~Glorot, A.~Bordes, and Y.~Bengio.
\newblock Deep sparse rectifier neural networks.
\newblock In {\em Proceedings of the Fourteenth International Conference on
  Artificial Intelligence and Statistics}, volume~15 of {\em {JMLR}
  Proceedings}, pages 315--323. JMLR.org, 2011.

\bibitem{goodfellow2016deep}
I.~Goodfellow, Y.~Bengio, and A.~Courville.
\newblock {\em Deep learning}.
\newblock MIT press, 2016.

\bibitem{DBLP:journals/corr/GoodfellowSS14}
I.~J. Goodfellow, J.~Shlens, and C.~Szegedy.
\newblock Explaining and harnessing adversarial examples.
\newblock In {\em 3rd International Conference on Learning Representations},
  2015.

\bibitem{goodman2016european}
B.~Goodman and S.~R. Flaxman.
\newblock European union regulations on algorithmic decision-making and a
  ``right to explanation''.
\newblock {\em {AI} Magazine}, 38(3):50--57, 2017.

\bibitem{grabisch1999axiomatic}
M.~Grabisch and M.~Roubens.
\newblock An axiomatic approach to the concept of interaction among players in
  cooperative games.
\newblock {\em International Journal of Game Theory}, 28(4):547--565, 1999.

\bibitem{gu2019contextual}
J.~Gu and V.~Tresp.
\newblock Contextual prediction difference analysis.
\newblock {\em arXiv preprint arXiv:1910.09086}, 2019.

\bibitem{DBLP:conf/accv/GuYT18}
J.~Gu, Y.~Yang, and V.~Tresp.
\newblock Understanding individual decisions of cnns via contrastive
  backpropagation.
\newblock In {\em 14th Asian Conference on Computer Vision}, volume 11363 of
  {\em Lecture Notes in Computer Science}, pages 119--134. Springer, 2018.

\bibitem{guidotti2018local}
R.~Guidotti, A.~Monreale, S.~Ruggieri, D.~Pedreschi, F.~Turini, and
  F.~Giannotti.
\newblock Local rule-based explanations of black box decision systems.
\newblock {\em arXiv preprint arXiv:1805.10820}, 2018.

\bibitem{DBLP:journals/csur/GuidottiMRTGP19}
R.~Guidotti, A.~Monreale, S.~Ruggieri, F.~Turini, F.~Giannotti, and
  D.~Pedreschi.
\newblock A survey of methods for explaining black box models.
\newblock {\em {ACM} Computing Surveys}, 51(5):93:1--93:42, 2019.

\bibitem{DBLP:conf/cvpr/GuptaJFSA18}
A.~Gupta, J.~Johnson, L.~Fei{-}Fei, S.~Savarese, and A.~Alahi.
\newblock Social {GAN:} socially acceptable trajectories with generative
  adversarial networks.
\newblock In {\em {IEEE} Conference on Computer Vision and Pattern
  Recognition}, pages 2255--2264, 2018.

\bibitem{hagele2019resolving}
M.~H\"{a}gele, P.~Seegerer, S.~Lapuschkin, M.~Bockmayr, W.~Samek, F.~Klauschen,
  K.-R. M\"{u}ller, and A.~Binder.
\newblock Resolving challenges in deep learning-based analyses of
  histopathological images using explanation methods.
\newblock {\em Scientific Reports}, 10(1):6423, 2020.

\bibitem{hansen2011visual}
K.~Hansen, D.~Baehrens, T.~Schroeter, M.~Rupp, and K.-R. M{\"u}ller.
\newblock Visual interpretation of kernel-based prediction models.
\newblock {\em Molecular Informatics}, 30(9):817--826, 2011.

\bibitem{DBLP:journals/neuroimage/HaufeMGDHBB14}
S.~Haufe, F.~C. Meinecke, K.~G{\"{o}}rgen, S.~D{\"{a}}hne, J.~Haynes,
  B.~Blankertz, and F.~Bie{\ss}mann.
\newblock On the interpretation of weight vectors of linear models in
  multivariate neuroimaging.
\newblock {\em NeuroImage}, 87:96--110, 2014.

\bibitem{hawkins1974}
D.~M. Hawkins.
\newblock The detection of errors in multivariate data using principal
  components.
\newblock {\em Journal of the American Statistical Association},
  69(346):340--344, 1974.

\bibitem{DBLP:conf/cvpr/HeZRS16}
K.~He, X.~Zhang, S.~Ren, and J.~Sun.
\newblock Deep residual learning for image recognition.
\newblock In {\em {IEEE} Conference on Computer Vision and Pattern
  Recognition}, pages 770--778, 2016.

\bibitem{DBLP:conf/www/HeLZNHC17}
X.~He, L.~Liao, H.~Zhang, L.~Nie, X.~Hu, and T.~Chua.
\newblock Neural collaborative filtering.
\newblock In {\em Proceedings of the 26th International Conference on World
  Wide Web}, pages 173--182. {ACM}, 2017.

\bibitem{DBLP:conf/eccv/HendricksARDSD16}
L.~A. Hendricks, Z.~Akata, M.~Rohrbach, J.~Donahue, B.~Schiele, and T.~Darrell.
\newblock Generating visual explanations.
\newblock In {\em 14th European Conference on Computer Vision}, volume 9908 of
  {\em Lecture Notes in Computer Science}, pages 3--19. Springer, 2016.

\bibitem{heo2019fooling}
J.~Heo, S.~Joo, and T.~Moon.
\newblock Fooling neural network interpretations via adversarial model
  manipulation.
\newblock In {\em Advances in Neural Information Processing Systems 32}, pages
  2921--2932, 2019.

\bibitem{hernan2009invited}
M.~A. Hern{\'a}n and S.~R. Cole.
\newblock Invited commentary: causal diagrams and measurement bias.
\newblock {\em American journal of epidemiology}, 170(8):959--962, 2009.

\bibitem{hochreiter1997long}
S.~Hochreiter and J.~Schmidhuber.
\newblock Long short-term memory.
\newblock {\em Neural Computation}, 9(8):1735--1780, 1997.

\bibitem{hochuli2018visualizing}
J.~Hochuli, A.~Helbling, T.~Skaist, M.~Ragoza, and D.~R. Koes.
\newblock Visualizing convolutional neural network protein-ligand scoring.
\newblock {\em Journal of Molecular Graphics and Modelling}, 2018.

\bibitem{hoffmann2007}
H.~Hoffmann.
\newblock Kernel {PCA} for novelty detection.
\newblock {\em Pattern Recognition}, 40(3):863--874, 2007.

\bibitem{holzinger2018machine}
A.~Holzinger.
\newblock From machine learning to explainable ai.
\newblock In {\em 2018 World Symposium on Digital Intelligence for Systems and
  Machines (DISA)}, pages 55--66, 2018.

\bibitem{holzinger2019causability}
A.~Holzinger, G.~Langs, H.~Denk, K.~Zatloukal, and H.~M{\"u}ller.
\newblock Causability and explainability of artificial intelligence in
  medicine.
\newblock {\em Wiley Interdisciplinary Reviews: Data Mining and Knowledge
  Discovery}, 9(4):e1312, 2019.

\bibitem{Hong2019}
S.~Hong, D.~Yang, J.~Choi, and H.~Lee.
\newblock Interpretable text-to-image synthesis with hierarchical semantic
  layout generation.
\newblock In {\em Explainable {AI:} Interpreting, Explaining and Visualizing
  Deep Learning}, volume 11700 of {\em Lecture Notes in Computer Science},
  pages 77--95. Springer, 2019.

\bibitem{hooker2019benchmark}
S.~Hooker, D.~Erhan, P.-J. Kindermans, and B.~Kim.
\newblock A benchmark for interpretability methods in deep neural networks.
\newblock In {\em Advances in Neural Information Processing Systems}, pages
  9734--9745, 2019.

\bibitem{horst2019explaining}
F.~Horst, S.~Lapuschkin, W.~Samek, K.-R. M{\"u}ller, and W.~I. Sch{\"o}llhorn.
\newblock Explaining the unique nature of individual gait patterns with deep
  learning.
\newblock {\em Scientific Reports}, 9(2391), 2019.

\bibitem{LFWTech}
G.~B. Huang, M.~Ramesh, T.~Berg, and E.~Learned-Miller.
\newblock Labeled faces in the wild: A database for studying face recognition
  in unconstrained environments.
\newblock Technical Report 07-49, University of Massachusetts, Amherst, 2007.

\bibitem{DBLP:journals/corr/IandolaMAHDK16}
F.~N. Iandola, M.~W. Moskewicz, K.~Ashraf, S.~Han, W.~J. Dally, and K.~Keutzer.
\newblock Squeezenet: Alexnet-level accuracy with 50x fewer parameters and
  {\textless}1mb model size.
\newblock {\em arXiv preprint arXiv:1602.07360}, 2016.

\bibitem{DBLP:conf/iccvw/IwanaKU19}
B.~K. Iwana, R.~Kuroki, and S.~Uchida.
\newblock Explaining convolutional neural networks using softmax gradient
  layer-wise relevance propagation.
\newblock In {\em {IEEE} International Conference on Computer Vision
  Workshops}, pages 4176--4185. {IEEE}, 2019.

\bibitem{Janizek2020IntegratedHessians}
J.~D. Janizek, P.~Sturmfels, and S.~Lee.
\newblock Explaining explanations: Axiomatic feature interactions for deep
  networks.
\newblock {\em arXiv preprint arXiv:2002.04138}, 2020.

\bibitem{Jarrahi2018}
M.~H. Jarrahi.
\newblock Artificial intelligence and the future of work: Human-{AI} symbiosis
  in organizational decision making.
\newblock {\em Business Horizons}, 61(4):577--586, 2018.

\bibitem{jia2014caffe}
Y.~Jia, E.~Shelhamer, J.~Donahue, S.~Karayev, J.~Long, R.~Girshick,
  S.~Guadarrama, and T.~Darrell.
\newblock Caffe: Convolutional architecture for fast feature embedding.
\newblock In {\em Proceedings of the 22nd ACM international conference on
  Multimedia}, pages 675--678, 2014.

\bibitem{DBLP:conf/cav/KatzBDJK17}
G.~Katz, C.~W. Barrett, D.~L. Dill, K.~Julian, and M.~J. Kochenderfer.
\newblock Reluplex: An efficient {SMT} solver for verifying deep neural
  networks.
\newblock In {\em {CAV} {(1)}}, volume 10426 of {\em Lecture Notes in Computer
  Science}, pages 97--117. Springer, 2017.

\bibitem{Kauffmann19}
J.~Kauffmann, M.~Esders, G.~Montavon, W.~Samek, and K.-R. M{\"{u}}ller.
\newblock From clustering to cluster explanations via neural networks.
\newblock {\em arXiv preprint arXiv:1906.07633}, 2019.

\bibitem{Kauffmann20}
J.~Kauffmann, K.-R. M\"{u}ller, and G.~Montavon.
\newblock Towards explaining anomalies: A deep {T}aylor decomposition of
  one-class models.
\newblock {\em Pattern Recognition}, 101:107198, 2020.

\bibitem{kelley2018sequential}
D.~R. Kelley, Y.~Reshef, M.~Bileschi, D.~Belanger, C.~Y. McLean, and J.~Snoek.
\newblock Sequential regulatory activity prediction across chromosomes with
  convolutional neural networks.
\newblock {\em Genome Research}, 28(5):739--750, 2018.

\bibitem{Khan2001}
J.~Khan, J.~S. Wei, M.~Ringn{\'{e}}r, L.~H. Saal, M.~Ladanyi, F.~Westermann,
  F.~Berthold, M.~Schwab, C.~R. Antonescu, C.~Peterson, and P.~S. Meltzer.
\newblock Classification and diagnostic prediction of cancers using gene
  expression profiling and artificial neural networks.
\newblock {\em Nature Medicine}, 7(6):673--679, 2001.

\bibitem{DBLP:conf/icml/KimWGCWVS18}
B.~Kim, M.~Wattenberg, J.~Gilmer, C.~J. Cai, J.~Wexler, F.~B. Vi{\'{e}}gas, and
  R.~Sayres.
\newblock Interpretability beyond feature attribution: Quantitative testing
  with concept activation vectors {(TCAV)}.
\newblock In {\em Proceedings of the 35th International Conference on Machine
  Learning}, volume~80 of {\em Proceedings of Machine Learning Research}, pages
  2673--2682. {PMLR}, 2018.

\bibitem{kindermans2017learning}
P.-J. Kindermans, K.~T. Sch{\"{u}}tt, M.~Alber, K.-R. M{\"{u}}ller, D.~Erhan,
  B.~Kim, and S.~D{\"{a}}hne.
\newblock Learning how to explain neural networks: {PatternNet} and
  {PatternAttribution}.
\newblock In {\em 6th International Conference on Learning Representations},
  2018.

\bibitem{DBLP:conf/iclr/KipfW17}
T.~N. Kipf and M.~Welling.
\newblock Semi-supervised classification with graph convolutional networks.
\newblock In {\em 5th International Conference on Learning Representations},
  2017.

\bibitem{Kittel}
C.~Kittel.
\newblock {\em Introduction to Solid State Physics}, volume~8.
\newblock Wiley New York, 2004.

\bibitem{klauschen2018scoring}
F.~Klauschen, K.-R. M{\"u}ller, A.~Binder, M.~Bockmayr, M.~H{\"a}gele,
  P.~Seegerer, S.~Wienert, G.~Pruneri, S.~de~Maria, S.~Badve, et~al.
\newblock Scoring of tumor-infiltrating lymphocytes: From visual estimation to
  machine learning.
\newblock {\em Seminars in Cancer Biology}, 52:151--157, 2018.

\bibitem{kohlbrenner2020towards}
M.~Kohlbrenner, A.~Bauer, S.~Nakajima, A.~Binder, W.~Samek, and S.~Lapuschkin.
\newblock Towards best practice in explaining neural network decisions with
  {LRP}.
\newblock In {\em 2020 International Joint Conference on Neural Networks},
  pages 1--7. {IEEE}, 2020.

\bibitem{captum2019github}
N.~Kokhlikyan, V.~Miglani, M.~Martin, E.~Wang, J.~Reynolds, A.~Melnikov,
  N.~Lunova, and O.~Reblitz-Richardson.
\newblock Pytorch {Captum}.
\newblock \texttt{https://github.com/pytorch/captum}, 2019.

\bibitem{Kourou2015}
K.~Kourou, T.~P. Exarchos, K.~P. Exarchos, M.~V. Karamouzis, and D.~I.
  Fotiadis.
\newblock Machine learning applications in cancer prognosis and prediction.
\newblock {\em Computational and Structural Biotechnology Journal}, 13:8--17,
  2015.

\bibitem{Kratzert2019}
F.~Kratzert, M.~Herrnegger, D.~Klotz, S.~Hochreiter, and G.~Klambauer.
\newblock Neuralhydrology - interpreting lstms in hydrology.
\newblock In {\em Explainable {AI:} Interpreting, Explaining and Visualizing
  Deep Learning}, volume 11700 of {\em Lecture Notes in Computer Science},
  pages 347--362. Springer, 2019.

\bibitem{KrausBF16}
O.~Z. Kraus, L.~J. Ba, and B.~J. Frey.
\newblock Classifying and segmenting microscopy images with deep multiple
  instance learning.
\newblock {\em Bioinformatics}, 32(12):52--59, 2016.

\bibitem{DBLP:conf/nips/KrizhevskySH12}
A.~Krizhevsky, I.~Sutskever, and G.~E. Hinton.
\newblock Imagenet classification with deep convolutional neural networks.
\newblock In {\em {NIPS}}, pages 1106--1114, 2012.

\bibitem{LandeckerTBMKB13}
W.~Landecker, M.~D. Thomure, L.~M.~A. Bettencourt, M.~Mitchell, G.~T. Kenyon,
  and S.~P. Brumby.
\newblock Interpreting individual classifications of hierarchical networks.
\newblock In {\em {IEEE} Symposium on Computational Intelligence and Data
  Mining}, pages 32--38, 2013.

\bibitem{LapCVPR16}
S.~Lapuschkin, A.~Binder, G.~Montavon, K.-R. M{\"u}ller, and W.~Samek.
\newblock Analyzing classifiers: Fisher vectors and deep neural networks.
\newblock In {\em Proceedings of the IEEE Conference on Computer Vision and
  Pattern Recognition (CVPR)}, pages 2912--2920, 2016.

\bibitem{LapJMLR16}
S.~Lapuschkin, A.~Binder, G.~Montavon, K.-R. M{\"u}ller, and W.~Samek.
\newblock The layer-wise relevance propagation toolbox for artificial neural
  networks.
\newblock {\em Journal of Machine Learning Research}, 17(114):1--5, 2016.

\bibitem{LapICCVW17}
S.~Lapuschkin, A.~Binder, K.-R. M\"uller, and W.~Samek.
\newblock Understanding and comparing deep neural networks for age and gender
  classification.
\newblock In {\em Proceedings of the IEEE International Conference on Computer
  Vision Workshops (ICCVW)}, pages 1629--38, 2017.

\bibitem{lapuschkin2019unmasking}
S.~Lapuschkin, S.~W{\"a}ldchen, A.~Binder, G.~Montavon, W.~Samek, and K.-R.
  M{\"u}ller.
\newblock Unmasking {C}lever {H}ans predictors and assessing what machines
  really learn.
\newblock {\em Nature Communications}, 10(1):1096, 2019.

\bibitem{DBLP:conf/nips/LarochelleH10}
H.~Larochelle and G.~E. Hinton.
\newblock Learning to combine foveal glimpses with a third-order {B}oltzmann
  machine.
\newblock In {\em Advances in Neural Information Processing Systems 23}, pages
  1243--1251, 2010.

\bibitem{Lauritsen2020}
S.~M. Lauritsen, M.~Kristensen, M.~V. Olsen, M.~S. Larsen, K.~M. Lauritsen,
  M.~J. J{\o}rgensen, J.~Lange, and B.~Thiesson.
\newblock Explainable artificial intelligence model to predict acute critical
  illness from electronic health records.
\newblock {\em Nature Communications}, 11(1), 2020.

\bibitem{lecun2015deep}
Y.~LeCun, Y.~Bengio, and G.~Hinton.
\newblock Deep learning.
\newblock {\em Nature}, 521(7553):436, 2015.

\bibitem{lecun2012efficient}
Y.~A. LeCun, L.~Bottou, G.~B. Orr, and K.-R. M{\"u}ller.
\newblock Efficient backprop.
\newblock In {\em Neural Networks: Tricks of the Trade}, pages 9--48. Springer,
  2012.

\bibitem{amt-2020-420}
Y.~Lee, C.~D. Kummerow, and I.~Ebert-Uphoff.
\newblock Applying machine learning methods to detect convection using goes-16
  abi data.
\newblock {\em Atmospheric Measurement Techniques Discussions}, 2020:1--28,
  2020.

\bibitem{leino2018influence}
K.~Leino, S.~Sen, A.~Datta, M.~Fredrikson, and L.~Li.
\newblock Influence-directed explanations for deep convolutional networks.
\newblock In {\em IEEE International Test Conference}, pages 1--8. IEEE, 2018.

\bibitem{lin2014network}
M.~Lin, Q.~Chen, and S.~Yan.
\newblock Network in network.
\newblock In {\em International Conference of Learning Representations (ICLR)},
  2014.

\bibitem{lipton2018mythos}
Z.~C. Lipton.
\newblock The mythos of model interpretability.
\newblock {\em {ACM} Queue}, 16(3):30, 2018.

\bibitem{DBLP:conf/acl/LiuYW19}
H.~Liu, Q.~Yin, and W.~Y. Wang.
\newblock Towards explainable {NLP:} {A} generative explanation framework for
  text classification.
\newblock In {\em Proceedings of the 57th Conference of the Association for
  Computational Linguistics}, pages 5570--5581. Association for Computational
  Linguistics, 2019.

\bibitem{Liu655639}
Y.~Liu, K.~Barr, and J.~Reinitz.
\newblock Fully interpretable deep learning model of transcriptional control.
\newblock {\em Bioinformatics}, 36(Supplement\_1):i499--i507, 2020.

\bibitem{lundberg2020local}
S.~M. Lundberg, G.~Erion, H.~Chen, A.~DeGrave, J.~M. Prutkin, B.~Nair, R.~Katz,
  J.~Himmelfarb, N.~Bansal, and S.-I. Lee.
\newblock From local explanations to global understanding with explainable ai
  for trees.
\newblock {\em Nature Machine Intelligence}, 2(1):2522--5839, 2020.

\bibitem{DBLP:conf/nips/LundbergL17}
S.~M. Lundberg and S.~Lee.
\newblock A unified approach to interpreting model predictions.
\newblock In {\em Advances in Neural Information Processing Systems 30}, pages
  4765--4774, 2017.

\bibitem{DBLP:journals/bmcbi/MaSH07}
S.~Ma, X.~Song, and J.~Huang.
\newblock Supervised group lasso with applications to microarray data analysis.
\newblock {\em {BMC} Bioinformatics}, 8, 2007.

\bibitem{macdonald2019rate}
J.~MacDonald, S.~W{\"{a}}ldchen, S.~Hauch, and G.~Kutyniok.
\newblock A rate-distortion framework for explaining neural network decisions.
\newblock {\em arXiv preprint arXiv:1905.11092}, 2019.

\bibitem{macqueen1967}
J.~MacQueen.
\newblock Some methods for classification and analysis of multivariate
  observations.
\newblock In {\em Proceedings of the Fifth Berkeley Symposium on Mathematical
  Statistics and Probability, Volume 1: Statistics}, pages 281--297, Berkeley,
  Calif., 1967. University of California Press.

\bibitem{matsui2001np}
Y.~Matsui and T.~Matsui.
\newblock Np-completeness for calculating power indices of weighted majority
  games.
\newblock {\em Theoretical Computer Science}, 263(1-2):305--310, 2001.

\bibitem{McGovern2019}
A.~McGovern, R.~Lagerquist, D.~J. Gagne, G.~E. Jergensen, K.~L. Elmore, C.~R.
  Homeyer, and T.~Smith.
\newblock Making the black box more transparent: Understanding the physical
  implications of machine learning.
\newblock {\em Bulletin of the American Meteorological Society},
  100(11):2175--2199, 2019.

\bibitem{DBLP:journals/pami/Memisevic13}
R.~Memisevic.
\newblock Learning to relate images.
\newblock {\em {IEEE} Transactions on Pattern Analysis and Machine
  Intelligence}, 35(8):1829--1846, 2013.

\bibitem{DBLP:journals/ai/Miller19}
T.~Miller.
\newblock Explanation in artificial intelligence: Insights from the social
  sciences.
\newblock {\em Artificial Intelligence}, 267:1--38, 2019.

\bibitem{DBLP:journals/nature/MnihKSRVBGRFOPB15}
V.~Mnih, K.~Kavukcuoglu, D.~Silver, A.~A. Rusu, J.~Veness, M.~G. Bellemare,
  A.~Graves, M.~A. Riedmiller, A.~Fidjeland, G.~Ostrovski, S.~Petersen,
  C.~Beattie, A.~Sadik, I.~Antonoglou, H.~King, D.~Kumaran, D.~Wierstra,
  S.~Legg, and D.~Hassabis.
\newblock Human-level control through deep reinforcement learning.
\newblock {\em Nature}, 518(7540):529--533, 2015.

\bibitem{DBLP:series/lncs/Montavon19}
G.~Montavon.
\newblock Gradient-based vs. propagation-based explanations: An axiomatic
  comparison.
\newblock In {\em Explainable {AI:} Interpreting, Explaining and Visualizing
  Deep Learning}, volume 11700 of {\em Lecture Notes in Computer Science},
  pages 253--265. Springer, 2019.

\bibitem{DBLP:series/lncs/MontavonBLSM19}
G.~Montavon, A.~Binder, S.~Lapuschkin, W.~Samek, and K.-R. M{\"{u}}ller.
\newblock Layer-wise relevance propagation: An overview.
\newblock In {\em Explainable {AI:} Interpreting, Explaining and Visualizing
  Deep Learning}, volume 11700 of {\em Lecture Notes in Computer Science},
  pages 193--209. Springer, 2019.

\bibitem{DBLP:journals/pr/MontavonLBSM17}
G.~Montavon, S.~Lapuschkin, A.~Binder, W.~Samek, and K.-R. M{\"{u}}ller.
\newblock Explaining nonlinear classification decisions with deep {T}aylor
  decomposition.
\newblock {\em Pattern Recognition}, 65:211--222, 2017.

\bibitem{montavon2018methods}
G.~Montavon, W.~Samek, and K.-R. M\"uller.
\newblock Methods for interpreting and understanding deep neural networks.
\newblock {\em Digital Signal Processing}, 73:1--15, 2018.

\bibitem{DBLP:conf/nips/MontufarPCB14}
G.~F. Mont{\'{u}}far, R.~Pascanu, K.~Cho, and Y.~Bengio.
\newblock On the number of linear regions of deep neural networks.
\newblock In {\em Advances in Neural Information Processing Systems 27}, pages
  2924--2932, 2014.

\bibitem{morch1995visualization}
N.~Morch, U.~Kjems, L.~K. Hansen, C.~Svarer, I.~Law, B.~Lautrup, S.~Strother,
  and K.~Rehm.
\newblock Visualization of neural networks using saliency maps.
\newblock In {\em Proceedings of ICNN'95-International Conference on Neural
  Networks}, volume~4, pages 2085--2090, 1995.

\bibitem{mordvintsev15}
A.~Mordvintsev, C.~Olah, and M.~Tyka.
\newblock Inceptionism: Going deeper into neural networks.
\newblock {\em Google Research Blog}, 2015.

\bibitem{muller2001introduction}
K.-R. M{\"u}ller, S.~Mika, G.~R{\"a}tsch, K.~Tsuda, and B.~Sch{\"o}lkopf.
\newblock An introduction to kernel-based learning algorithms.
\newblock {\em IEEE Transactions on Neural Networks}, 12(2):181--201, 2001.

\bibitem{Naranyan2018}
M.~Narayanan, E.~Chen, J.~He, B.~Kim, S.~Gershman, and F.~Doshi{-}Velez.
\newblock How do humans understand explanations from machine learning systems?
  an evaluation of the human-interpretability of explanation.
\newblock {\em arXiv preprint arXiv:1802.00682}, 2018.

\bibitem{DBLP:conf/nips/NguyenDYBC16}
A.~M. Nguyen, A.~Dosovitskiy, J.~Yosinski, T.~Brox, and J.~Clune.
\newblock Synthesizing the preferred inputs for neurons in neural networks via
  deep generator networks.
\newblock In {\em Advances in Neural Information Processing Systems 29}, pages
  3387--3395, 2016.

\bibitem{DBLP:conf/cvpr/NguyenYC15}
A.~M. Nguyen, J.~Yosinski, and J.~Clune.
\newblock Deep neural networks are easily fooled: High confidence predictions
  for unrecognizable images.
\newblock In {\em {IEEE} Conference on Computer Vision and Pattern
  Recognition}, pages 427--436, 2015.

\bibitem{DBLP:journals/corr/NguyenYC16}
A.~M. Nguyen, J.~Yosinski, and J.~Clune.
\newblock Multifaceted feature visualization: Uncovering the different types of
  features learned by each neuron in deep neural networks.
\newblock {\em arXiv preprint arXiv:1602.03616}, 2016.

\bibitem{cupylearningsys2017}
R.~Okuta, Y.~Unno, D.~Nishino, S.~Hido, and C.~Loomis.
\newblock Cupy: A numpy-compatible library for nvidia gpu calculations.
\newblock In {\em Advances in Neural Information Processing Systems}, 2017.

\bibitem{oliphant2006guide}
T.~E. Oliphant.
\newblock {\em A guide to NumPy}, volume~1.
\newblock Trelgol Publishing USA, 2006.

\bibitem{DBLP:journals/tnn/PapadopoulosEM01}
G.~Papadopoulos, P.~J. Edwards, and A.~F. Murray.
\newblock Confidence estimation methods for neural networks: a practical
  comparison.
\newblock {\em {IEEE} Transactions on Neural Networks}, 12(6):1278--1287, 2001.

\bibitem{park2020estimating}
Y.~Park, B.~Kwon, J.~Heo, X.~Hu, Y.~Liu, and T.~Moon.
\newblock Estimating pm2. 5 concentration of the conterminous united states via
  interpretable convolutional neural networks.
\newblock {\em Environmental Pollution}, 256:113395, 2020.

\bibitem{parzen1962}
E.~Parzen.
\newblock On estimation of a probability density function and mode.
\newblock {\em The Annals of Mathematical Statistics}, 33(3):1065--1076, 1962.

\bibitem{petsiuk2018rise}
V.~Petsiuk, A.~Das, and K.~Saenko.
\newblock {RISE:} randomized input sampling for explanation of black-box
  models.
\newblock In {\em British Machine Vision Conference}, page 151, 2018.

\bibitem{PoulinESLGWFPMA06}
B.~Poulin, R.~Eisner, D.~Szafron, P.~Lu, R.~Greiner, D.~S. Wishart, A.~Fyshe,
  B.~Pearcy, C.~Macdonell, and J.~Anvik.
\newblock Visual explanation of evidence with additive classifiers.
\newblock In {\em Proceedings, the 21st National Conference on Artificial
  Intelligence and the 18th Innovative Applications of Artificial Intelligence
  Conference}, pages 1822--1829, 2006.

\bibitem{Preuer2019}
K.~Preuer, G.~Klambauer, F.~Rippmann, S.~Hochreiter, and T.~Unterthiner.
\newblock Interpretable deep learning in drug discovery.
\newblock In {\em Explainable {AI:} Interpreting, Explaining and Visualizing
  Deep Learning}, volume 11700 of {\em Lecture Notes in Computer Science},
  pages 331--345. Springer, 2019.

\bibitem{quellec2017deep}
G.~Quellec, K.~Charri{\`{e}}re, Y.~Boudi, B.~Cochener, and M.~Lamard.
\newblock Deep image mining for diabetic retinopathy screening.
\newblock {\em Medical Image Analysis}, 39:178--193, 2017.

\bibitem{Raghunath2020}
S.~Raghunath, A.~E.~U. Cerna, L.~Jing, D.~P. vanMaanen, J.~Stough, D.~N.
  Hartzel, J.~B. Leader, H.~L. Kirchner, M.~C. Stumpe, A.~Hafez, A.~Nemani,
  T.~Carbonati, K.~W. Johnson, K.~Young, C.~W. Good, J.~M. Pfeifer, A.~A.
  Patel, B.~P. Delisle, A.~Alsaid, D.~Beer, C.~M. Haggerty, and B.~K. Fornwalt.
\newblock Prediction of mortality from 12-lead electrocardiogram voltage data
  using a deep neural network.
\newblock {\em Nature Medicine}, 26(6):886--891, 2020.

\bibitem{ribeiro2016should}
M.~T. Ribeiro, S.~Singh, and C.~Guestrin.
\newblock "why should {I} trust you?": Explaining the predictions of any
  classifier.
\newblock In {\em Proceedings of the 22nd {ACM} {SIGKDD} International
  Conference on Knowledge Discovery and Data Mining}, pages 1135--1144, 2016.

\bibitem{ribeiro2018anchors}
M.~T. Ribeiro, S.~Singh, and C.~Guestrin.
\newblock Anchors: High-precision model-agnostic explanations.
\newblock In {\em Proceedings of the {AAAI} Conference on Artificial
  Intelligence}, volume~18, pages 1527--1535, 2018.

\bibitem{rieger18}
L.~Rieger, P.~Chormai, G.~Montavon, L.~Hansen, and K.-R. M{\"u}ller.
\newblock Structuring neural networks for more explainable predictions.
\newblock In {\em Explainable and Interpretable Models in Computer Vision and
  Machine Learning}, pages 115--131. Springer, 2019.

\bibitem{DBLP:journals/access/RoscherBDG20}
R.~Roscher, B.~Bohn, M.~F. Duarte, and J.~Garcke.
\newblock Explainable machine learning for scientific insights and discoveries.
\newblock {\em {IEEE} Access}, 8:42200--42216, 2020.

\bibitem{Rosenblatt1958}
F.~Rosenblatt.
\newblock The perceptron: A probabilistic model for information storage and
  organization in the brain.
\newblock {\em Psychological Review}, 65(6):386--408, 1958.

\bibitem{DBLP:conf/ijcai/RossHD17}
A.~S. Ross, M.~C. Hughes, and F.~Doshi{-}Velez.
\newblock Right for the right reasons: Training differentiable models by
  constraining their explanations.
\newblock In {\em Proceedings of the Twenty-Sixth International Joint
  Conference on Artificial Intelligence}, pages 2662--2670, 2017.

\bibitem{rothe2015dex}
R.~Rothe, R.~Timofte, and L.~Van~Gool.
\newblock Dex: Deep expectation of apparent age from a single image.
\newblock In {\em Proceedings of the IEEE International Conference on Computer
  Vision Workshops}, pages 10--15, 2015.

\bibitem{Rudin2019}
C.~Rudin.
\newblock Stop explaining black box machine learning models for high stakes
  decisions and use interpretable models instead.
\newblock {\em Nature Machine Intelligence}, 1(5):206--215, 2019.

\bibitem{ruff2020unifying}
L.~Ruff, J.~R. Kauffmann, R.~A. Vandermeulen, G.~Montavon, W.~Samek, M.~Kloft,
  T.~G. Dietterich, and K.-R. M{\"u}ller.
\newblock A unifying review of deep and shallow anomaly detection.
\newblock {\em Proceedings of the IEEE}, pages 1--40, 2021.

\bibitem{ruff2018deep}
L.~Ruff, R.~Vandermeulen, N.~Goernitz, L.~Deecke, S.~A. Siddiqui, A.~Binder,
  E.~M{\"u}ller, and M.~Kloft.
\newblock Deep one-class classification.
\newblock In {\em International Conference on Machine Learning}, pages
  4393--4402, 2018.

\bibitem{DBLP:journals/ijcv/RussakovskyDSKS15}
O.~Russakovsky, J.~Deng, H.~Su, J.~Krause, S.~Satheesh, S.~Ma, Z.~Huang,
  A.~Karpathy, A.~Khosla, M.~S. Bernstein, A.~C. Berg, and F.~Li.
\newblock Imagenet large scale visual recognition challenge.
\newblock {\em International Journal of Computer Vision}, 115(3):211--252,
  2015.

\bibitem{samek2016evaluating}
W.~Samek, A.~Binder, G.~Montavon, S.~Lapuschkin, and K.-R. M{\"u}ller.
\newblock Evaluating the visualization of what a deep neural network has
  learned.
\newblock {\em IEEE Transactions on Neural Networks and Learning Systems},
  28(11):2660--2673, 2016.

\bibitem{DBLP:series/lncs/11700}
W.~Samek, G.~Montavon, A.~Vedaldi, L.~K. Hansen, and K.-R. M{\"{u}}ller,
  editors.
\newblock {\em Explainable {AI:} Interpreting, Explaining and Visualizing Deep
  Learning}, volume 11700 of {\em Lecture Notes in Computer Science}.
\newblock Springer, 2019.

\bibitem{DBLP:journals/tnn/ScarselliGTHM09}
F.~Scarselli, M.~Gori, A.~C. Tsoi, M.~Hagenbuchner, and G.~Monfardini.
\newblock The graph neural network model.
\newblock {\em {IEEE} Transactions on Neural Networks}, 20(1):61--80, 2009.

\bibitem{schmidhuber2015deep}
J.~Schmidhuber.
\newblock Deep learning in neural networks: An overview.
\newblock {\em Neural Networks}, 61:85--117, 2015.

\bibitem{Schnake2020Graphs}
T.~Schnake, O.~Eberle, J.~Lederer, S.~Nakajima, K.~T. Sch{\"{u}}tt, K.-R.
  M{\"{u}}ller, and G.~Montavon.
\newblock Higher-order explanations of graph neural networks via relevant
  walks.
\newblock {\em arXiv preprint arXiv:2006.03589}, 2020.

\bibitem{scholkopf2002learning}
B.~Sch{\"{o}}lkopf and A.~J. Smola.
\newblock {\em Learning with Kernels: Support Vector Machines, Regularization,
  Optimization, and Beyond}.
\newblock Adaptive Computation and Machine Learning Series. {MIT} Press, 2002.

\bibitem{scholkopf1998nonlinear}
B.~Sch{\"o}lkopf, A.~J. Smola, and K.-R. M{\"u}ller.
\newblock Nonlinear component analysis as a kernel eigenvalue problem.
\newblock {\em Neural Computation}, 10(5):1299--1319, 1998.

\bibitem{schutt2017quantum}
K.~T. Sch{\"u}tt, F.~Arbabzadah, S.~Chmiela, K.~R. M{\"u}ller, and
  A.~Tkatchenko.
\newblock Quantum-chemical insights from deep tensor neural networks.
\newblock {\em Nature Communications}, 8:13890, 2017.

\bibitem{DBLP:series/lncs/SchuttGTM19}
K.~T. Sch{\"{u}}tt, M.~Gastegger, A.~Tkatchenko, and K.-R. M{\"{u}}ller.
\newblock Quantum-chemical insights from interpretable atomistic neural
  networks.
\newblock In {\em Explainable {AI:} Interpreting, Explaining and Visualizing
  Deep Learning}, volume 11700 of {\em Lecture Notes in Computer Science},
  pages 311--330. Springer, 2019.

\bibitem{selvaraju2017visual}
R.~R. Selvaraju, M.~Cogswell, A.~Das, R.~Vedantam, D.~Parikh, and D.~Batra.
\newblock Grad-{CAM}: Visual explanations from deep networks via gradient-based
  localization.
\newblock In {\em {IEEE} International Conference on Computer Vision ({ICCV})},
  pages 618--626, 2017.

\bibitem{shan2010learning}
C.~Shan.
\newblock Learning local features for age estimation on real-life faces.
\newblock In {\em Proceedings of the 1st ACM international workshop on
  Multimodal pervasive video analysis}, pages 23--28. ACM, 2010.

\bibitem{Shapley}
L.~S. Shapley.
\newblock 17. a value for n-person games.
\newblock In {\em Contributions to the Theory of Games ({AM}-28), Volume {II}}.
  Princeton University Press, 1953.

\bibitem{shrikumar2017learning}
A.~Shrikumar, P.~Greenside, and A.~Kundaje.
\newblock Learning important features through propagating activation
  differences.
\newblock In {\em Proceedings of the 34th International Conference on Machine
  Learning ({ICML})}, pages 3145--3153, 2017.

\bibitem{shrikumar2016not}
A.~Shrikumar, P.~Greenside, A.~Shcherbina, and A.~Kundaje.
\newblock Not just a black box: Learning important features through propagating
  activation differences.
\newblock {\em arXiv preprint arXiv:1605.01713}, 2016.

\bibitem{shrikumar2018computationally}
A.~Shrikumar, J.~Su, and A.~Kundaje.
\newblock Computationally efficient measures of internal neuron importance.
\newblock {\em arXiv preprint arXiv:1807.09946}, 2018.

\bibitem{DBLP:journals/pami/SimonRDD20}
M.~Simon, E.~Rodner, T.~Darrell, and J.~Denzler.
\newblock The whole is more than its parts? from explicit to implicit pose
  normalization.
\newblock {\em IEEE Transactions on Pattern Analysis and Machine Intelligence},
  42(3):749--763, 2020.

\bibitem{DBLP:journals/corr/SimonyanVZ13}
K.~Simonyan, A.~Vedaldi, and A.~Zisserman.
\newblock Deep inside convolutional networks: Visualising image classification
  models and saliency maps.
\newblock In {\em International Conference of Learning Representations
  Workshops}, 2014.

\bibitem{DBLP:journals/corr/SimonyanZ14a}
K.~Simonyan and A.~Zisserman.
\newblock Very deep convolutional networks for large-scale image recognition.
\newblock In {\em 3rd International Conference on Learning Representations},
  2015.

\bibitem{DBLP:journals/corr/SmilkovTKVW17}
D.~Smilkov, N.~Thorat, B.~Kim, F.~B. Vi{\'{e}}gas, and M.~Wattenberg.
\newblock Smoothgrad: removing noise by adding noise.
\newblock {\em arXiv preprint arXiv:1706.03825}, 2017.

\bibitem{Soneson2014}
C.~Soneson, S.~Gerster, and M.~Delorenzi.
\newblock Batch effect confounding leads to strong bias in performance
  estimates obtained by cross-validation.
\newblock {\em {PLoS} {ONE}}, 9(6):e100335, 2014.

\bibitem{springenberg2015striving}
J.~T. Springenberg, A.~Dosovitskiy, T.~Brox, and M.~A. Riedmiller.
\newblock Striving for simplicity: The all convolutional net.
\newblock In {\em International Conference of Learning Representations (ICLR)},
  2015.

\bibitem{DBLP:journals/jmlr/StrumbeljK10}
E.~Strumbelj and I.~Kononenko.
\newblock An efficient explanation of individual classifications using game
  theory.
\newblock {\em Journal of Machine Learning Research}, 11:1--18, 2010.

\bibitem{StuJNM16}
I.~Sturm, S.~Lapuschkin, W.~Samek, and K.-R. M{\"u}ller.
\newblock Interpretable deep neural networks for single-trial eeg
  classification.
\newblock {\em Journal of Neuroscience Methods}, 274:141--145, 2016.

\bibitem{Sturmfels2020}
P.~Sturmfels, S.~Lundberg, and S.-I. Lee.
\newblock Visualizing the impact of feature attribution baselines.
\newblock {\em Distill}, 5(1), Jan. 2020.

\bibitem{DBLP:conf/eccv/SuZCYCG18}
D.~Su, H.~Zhang, H.~Chen, J.~Yi, P.~Chen, and Y.~Gao.
\newblock Is robustness the cost of accuracy? - {A} comprehensive study on the
  robustness of 18 deep image classification models.
\newblock In {\em 15th European Conference on Computer Vision}, volume 11216 of
  {\em Lecture Notes in Computer Science}, pages 644--661. Springer, 2018.

\bibitem{SundararajanTY17}
M.~Sundararajan, A.~Taly, and Q.~Yan.
\newblock Axiomatic attribution for deep networks.
\newblock In {\em Proceedings of the 34th International Conference on Machine
  Learning}, pages 3319--3328, 2017.

\bibitem{Swartout1993}
W.~R. Swartout and J.~D. Moore.
\newblock Explanation in second generation expert systems.
\newblock In {\em Second Generation Expert Systems}, pages 543--585. Springer
  Berlin Heidelberg, 1993.

\bibitem{DBLP:journals/corr/SzegedyZSBEGF13}
C.~Szegedy, W.~Zaremba, I.~Sutskever, J.~Bruna, D.~Erhan, I.~J. Goodfellow, and
  R.~Fergus.
\newblock Intriguing properties of neural networks.
\newblock In {\em International Conference of Learning Representations (ICLR)},
  2014.

\bibitem{10.3389/fnins.2019.01321}
A.~W. Thomas, H.~R. Heekeren, K.-R. M\"uller, and W.~Samek.
\newblock Analyzing neuroimaging data through recurrent deep learning models.
\newblock {\em Frontiers in Neuroscience}, 13:1321, 2019.

\bibitem{traunmuller1995frequency}
H.~Traunm{\"u}ller and A.~Eriksson.
\newblock The frequency range of the voice fundamental in the speech of male
  and female adults.
\newblock {\em Unpublished manuscript}, 1995.

\bibitem{DBLP:conf/iclr/TsangC018}
M.~Tsang, D.~Cheng, and Y.~Liu.
\newblock Detecting statistical interactions from neural network weights.
\newblock In {\em 6th International Conference on Learning Representations},
  2018.

\bibitem{DBLP:conf/fat/UstunSL19}
B.~Ustun, A.~Spangher, and Y.~Liu.
\newblock Actionable recourse in linear classification.
\newblock In {\em Proceedings of the Conference on Fairness, Accountability,
  and Transparency}, pages 10--19. {ACM}, 2019.

\bibitem{vapnik95}
V.~Vapnik.
\newblock {\em The Nature of Statistical Learning Theory}.
\newblock Springer, 1995.

\bibitem{vidovic2015opening}
M.~M.-C. Vidovic, N.~G{\"o}rnitz, K.-R. M{\"u}ller, G.~R{\"a}tsch, and
  M.~Kloft.
\newblock Opening the black box: Revealing interpretable sequence motifs in
  kernel-based learning algorithms.
\newblock In {\em Joint European Conference on Machine Learning and Knowledge
  Discovery in Databases}, pages 137--153. Springer, 2015.

\bibitem{Malsburg}
C.~Von~der Malsburg.
\newblock Binding in models of perception and brain function.
\newblock {\em Current Opinion in Neurobiology}, 5(4):520--526, 1995.

\bibitem{warnecke2019don}
A.~Warnecke, D.~Arp, C.~Wressnegger, and K.~Rieck.
\newblock Don't paint it black: White-box explanations for deep learning in
  computer security.
\newblock {\em arXiv preprint arXiv:1906.02108}, 2019.

\bibitem{williams2006gaussian}
C.~K. Williams and C.~E. Rasmussen.
\newblock {\em Gaussian Processes for Machine Learning}.
\newblock MIT press Cambridge, MA, 2006.

\bibitem{DBLP:conf/icml/XuBKCCSZB15}
K.~Xu, J.~Ba, R.~Kiros, K.~Cho, A.~C. Courville, R.~Salakhutdinov, R.~S. Zemel,
  and Y.~Bengio.
\newblock Show, attend and tell: Neural image caption generation with visual
  attention.
\newblock In {\em Proceedings of the 32nd International Conference on Machine
  Learning}, volume~37 of {\em {JMLR} Workshop and Conference Proceedings},
  pages 2048--2057. JMLR.org, 2015.

\bibitem{yinchong2018explaining}
Y.~Yang, V.~Tresp, M.~Wunderle, and P.~A. Fasching.
\newblock Explaining therapy predictions with layer-wise relevance propagation
  in neural networks.
\newblock In {\em {IEEE} International Conference on Healthcare Informatics},
  pages 152--162, 2018.

\bibitem{Yeh1998}
I.-C. Yeh.
\newblock Modeling of strength of high-performance concrete using artificial
  neural networks.
\newblock {\em Cement and Concrete Research}, 28(12):1797--1808, 1998.

\bibitem{DBLP:conf/nips/YingBYZL19}
Z.~Ying, D.~Bourgeois, J.~You, M.~Zitnik, and J.~Leskovec.
\newblock Gnnexplainer: Generating explanations for graph neural networks.
\newblock In {\em Advances in Neural Information Processing Systems 32}, pages
  9240--9251, 2019.

\bibitem{young2019deep}
K.~Young, G.~Booth, B.~Simpson, R.~Dutton, and S.~Shrapnel.
\newblock Deep neural network or dermatologist?
\newblock In {\em Interpretability of Machine Intelligence in Medical Image
  Computing and Multimodal Learning for Clinical Decision Support}, pages
  48--55. Springer, 2019.

\bibitem{DBLP:conf/icml/ZahavyBM16}
T.~Zahavy, N.~Ben{-}Zrihem, and S.~Mannor.
\newblock Graying the black box: Understanding dqns.
\newblock In {\em Proceedings of the 33nd International Conference on Machine
  Learning}, volume~48 of {\em {JMLR} Workshop and Conference Proceedings},
  pages 1899--1908. JMLR.org, 2016.

\bibitem{zeiler2014visualizing}
M.~D. Zeiler and R.~Fergus.
\newblock Visualizing and understanding convolutional networks.
\newblock In {\em European Conference Computer Vision - {ECCV} 2014}, pages
  818--833, 2014.

\bibitem{zhang2018neural}
J.~Zhang, Z.~Bargal, Sarah Adeland~Lin, J.~Brandt, X.~Shen, and S.~Sclaroff.
\newblock Top-down neural attention by excitation backprop.
\newblock {\em International Journal of Computer Vision}, 126(10):1084--1102,
  2018.

\bibitem{Zhang2020ExGraph}
Q.~{Zhang}, X.~{Wang}, R.~{Cao}, Y.~N. {Wu}, F.~{Shi}, and S.-C. {Zhu}.
\newblock Extracting an explanatory graph to interpret a cnn.
\newblock {\em IEEE Transactions on Pattern Analysis and Machine Intelligence},
  pages 1--1, 2020.

\bibitem{zhang2019pathologist}
Z.~Zhang, P.~Chen, M.~McGough, F.~Xing, C.~Wang, M.~Bui, Y.~Xie, M.~Sapkota,
  L.~Cui, J.~Dhillon, et~al.
\newblock Pathologist-level interpretable whole-slide cancer diagnosis with
  deep learning.
\newblock {\em Nature Machine Intelligence}, 1(5):236--245, 2019.

\bibitem{zhou2018interpreting}
B.~Zhou, D.~Bau, A.~Oliva, and A.~Torralba.
\newblock Interpreting deep visual representations via network dissection.
\newblock {\em IEEE Transactions on Pattern Analysis and Machine Intelligence},
  41(9):2131--2145, 2018.

\bibitem{zhou2016learning}
B.~Zhou, A.~Khosla, {\`{A}}.~Lapedriza, A.~Oliva, and A.~Torralba.
\newblock Learning deep features for discriminative localization.
\newblock In {\em {IEEE} Conference on Computer Vision and Pattern Recognition
  ({CVPR})}, pages 2921--2929, 2016.

\bibitem{zhu2019galaxy}
X.-P. Zhu, J.-M. Dai, C.-J. Bian, Y.~Chen, S.~Chen, and C.~Hu.
\newblock Galaxy morphology classification with deep convolutional neural
  networks.
\newblock {\em Astrophysics and Space Science}, 364(4):55, 2019.

\bibitem{zintgraf2017visualizing}
L.~M. Zintgraf, T.~S. Cohen, T.~Adel, and M.~Welling.
\newblock Visualizing deep neural network decisions: Prediction difference
  analysis.
\newblock In {\em International Conference on Learning Representations (ICLR),
  2017}, 2017.

\end{thebibliography}

%\input{biographies.tex}

%%%%%%%%%%%%%%%%%%%%%%%%%%%%%%%%%%%%%%%%%%%%%%%%%%%%%%%%%%%%%%%%%%%%%%%%%%%%%%%%%%%%

\end{document}